\documentclass{article}
\usepackage[accepted]{icml2020}
\usepackage{times}
\usepackage{latexsym}
\usepackage{url}

\usepackage{microtype}
\usepackage{makecell}
\usepackage{amsmath}
\usepackage{algorithm}
\usepackage{algorithmic}
\usepackage{chngcntr}
\usepackage{graphicx}
\usepackage{float}
\floatstyle{plaintop}
\restylefloat{table}
\usepackage{caption}
\usepackage{subcaption}
\usepackage{scrextend}
\usepackage{hyperref}
\usepackage{xcolor}
\usepackage{xspace}
\usepackage[explicit]{titlesec}
\usepackage{sidecap}

\definecolor{amaranth}{rgb}{0.7, 0.17, 0.31}
\definecolor{airforceblue}{rgb}{0.30, 0.44, 0.56}
\definecolor{darkgreen}{rgb}{0.13, 0.45, 0.13}
\definecolor{bronze}{rgb}{0.8, 0.5, 0.2}

\graphicspath{ {./figs/} }

\newcommand\pegasusbase{$\text{PEGASUS}_{\text{BASE}}$\xspace}
\newcommand\transformerbase{$\text{Transformer}_{\text{BASE}}$\xspace}
\newcommand\pegasuslarge{$\text{PEGASUS}_{\text{LARGE}}$\xspace}

\newcommand\largemodelsize{568M\xspace}
\newcommand\basemodelsize{223M\xspace}
\newcommand\gnews{HugeNews\xspace}
\newcommand\pegasuscode{\url{https://github.com/google-research/pegasus}}
\DeclareMathOperator*{\argmax}{arg\,max}

\usepackage[linewidth=1pt]{mdframed}

\icmltitlerunning{PEGASUS: Pre-training with Extracted Gap-sentences for Abstractive Summarization}

\begin{document}

\twocolumn[
\icmltitle{PEGASUS: Pre-training with Extracted Gap-sentences for \\ Abstractive Summarization}

\icmlsetsymbol{equal}{*}
\begin{icmlauthorlist}
\icmlauthor{Jingqing Zhang}{equal,ic}
\icmlauthor{Yao Zhao}{equal,goo}
\icmlauthor{Mohammad Saleh}{goo}
\icmlauthor{Peter J. Liu}{goo}
\end{icmlauthorlist}

\icmlaffiliation{ic}{Data Science Institute, Imperial College London, London, UK}
\icmlaffiliation{goo}{Brain Team, Google Research, Mountain View, CA, USA}

\icmlcorrespondingauthor{Jingqing Zhang}{jingqing.zhang15@imperial.ac.uk}
\icmlcorrespondingauthor{Yao Zhao}{yaozhaoyz@google.com}
\icmlcorrespondingauthor{Mohammad Saleh}{msaleh@google.com}
\icmlcorrespondingauthor{Peter J. Liu}{peterjliu@google.com}

\icmlkeywords{Text Summarization, Natural Language Processing, Deep Learning, Pre-training,
Self-supervised Learning}
\vskip 0.3in
]


\printAffiliationsAndNotice{\icmlEqualContribution} 

\begin{abstract}
Recent work pre-training Transformers with self-supervised objectives on large text corpora has shown great success when fine-tuned on downstream NLP tasks including text summarization.
However, pre-training objectives tailored for abstractive text summarization have not been explored. Furthermore there is a lack of systematic evaluation across diverse domains. 
In this work, we propose pre-training large Transformer-based encoder-decoder models on massive text corpora with a new self-supervised objective.
In PEGASUS, important sentences are removed/masked from an input document and
are generated together as one output sequence from the remaining sentences, similar to an
extractive summary.
We evaluated our best PEGASUS model on 12 downstream summarization tasks spanning news, science, stories, instructions, emails, patents, and legislative bills. Experiments demonstrate it achieves state-of-the-art performance on all 12 downstream datasets measured by ROUGE scores.
Our model 
also shows surprising performance on low-resource summarization, surpassing previous  state-of-the-art results on 6 datasets with only 1000 examples. 
 Finally we validated our results using human evaluation and show that our model
 summaries achieve  human performance on multiple datasets.

\end{abstract}


\begin{figure}[t]
\centering
\includegraphics[width=0.45\textwidth, trim={5cm 5cm 10cm 5cm},clip]{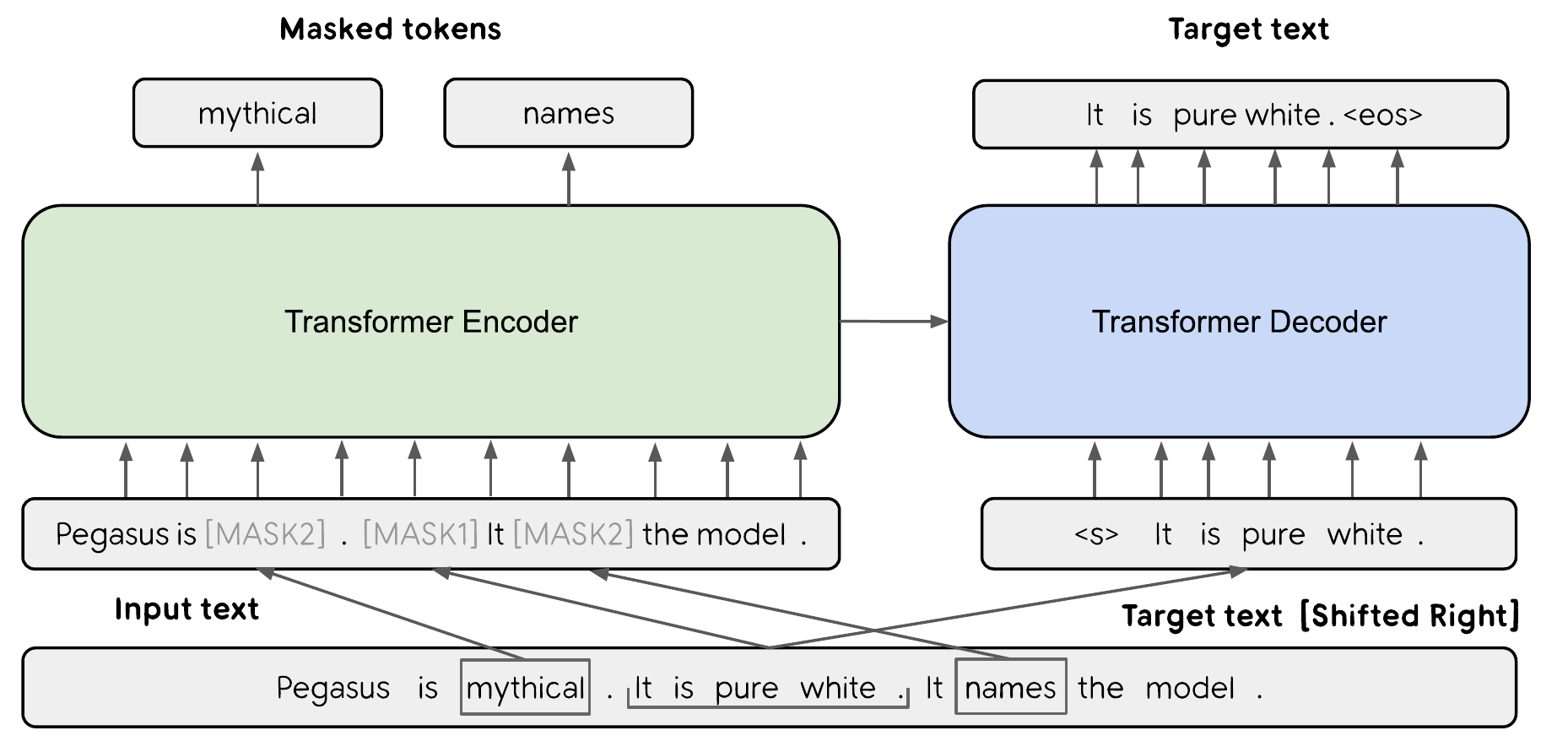}
\caption{
The base architecture of PEGASUS is a standard Transformer encoder-decoder. 
Both GSG and MLM are applied simultaneously to this example as
pre-training objectives. Originally there are three sentences. One sentence is masked with \texttt{[MASK1]} and used as target generation text (GSG). The other two sentences remain in the input, but some tokens are randomly masked by \texttt{[MASK2]} (MLM).
}
\label{fig:architecture}
\end{figure}

\section{Introduction}
\label{sec:intro}
Text summarization aims at generating accurate and concise summaries from input document(s).  In contrast to extractive summarization which merely copies informative fragments from the input, abstractive summarization may generate novel words. A good abstractive summary covers principal information in the input  and is linguistically fluent. 

In abstractive summarization, sequence-to-sequence \citep{seq2seq} has become a dominant framework using encoder-decoder architectures based on RNNs \citep{chung2014empirical,Hochreiter:1997:LSM:1246443.1246450} and more recently Transformers \citep{vaswani2017attention}. 
Most prior work on neural abstractive summarization relied on large-scale, high-quality datasets of supervised document-summary pairs \citep{hermann2015cnndm}
and achieved promising results \citep{rush-etal-2015-neural, nallapati-etal-2016-abstractive, see2017cnndm}. 
In recent years, there has been increased interest in collecting new summarization datasets that have more abstractive summaries \citep{shashi-2018-xsum}, have longer documents,  \citep{cohan-etal-2018-discourse-scientificpapers,sharma-etal-2019-bigpatent}, utilize multiple documents \citep{fabbri-etal-2019-multinews}, and are
sourced from diverse domains \citep{Grusky2018newsroom,koupaee2018wikihow,kim-etal-2019-reddittifu,kornilova-eidelman-2019-billsum,zhang-tetreault-2019-aeslc};
however, there has been little work on systematic evaluation of models across these broad settings.

Contemporaneously, the adoption of Transformer models \citep{vaswani2017attention} pre-trained using self-supervised
objectives on large text corpora \citep{radford2018improving,devlin_2019} have improved performance on many NLP tasks \citep{Wang_2018,Rajpurkar_2016}.

Recent work leveraging such pre-training for
  Transformer-based sequence-to-sequence models \citep{unilm,song2019mass,rothe2019leveraging,lewis2019bart,raffel2019exploring} has extended the success to text generation, including abstractive summarization.
  
In this work, we study pre-training objectives specifically for abstractive text summarization and evaluate on 12 downstream datasets spanning news \citep{hermann2015cnndm,shashi-2018-xsum,Grusky2018newsroom,rush-etal-2015-neural,fabbri-etal-2019-multinews}, science  \citep{cohan-etal-2018-discourse-scientificpapers}, short stories \citep{kim-etal-2019-reddittifu}, instructions \citep{koupaee2018wikihow}, emails \citep{zhang-tetreault-2019-aeslc}, patents \citep{sharma-etal-2019-bigpatent}, and legislative bills \citep{kornilova-eidelman-2019-billsum}. 
We find that masking whole sentences from a document
and generating these gap-sentences from the rest of the document
works well as a pre-training objective for downstream summarization tasks.
In particular, choosing putatively important sentences outperforms
lead or randomly selected ones.
We hypothesize this objective is suitable for abstractive
summarization as it closely resembles the
downstream task, encouraging whole-document understanding and summary-like generation.
We call this self-supervised objective Gap Sentences Generation (GSG).
Using GSG to pre-train a Transformer encoder-decoder on large corpora of documents (Web and
news articles) results in our method,
Pre-training with Extracted Gap-sentences for Abstractive SUmmarization Sequence-to-sequence models, or \emph{PEGASUS}.

With our best \largemodelsize parameter model trained
on the recently introduced C4 \citep{raffel2019exploring} corpus we equal or exceed state-of-the-art on the 12 summarization
tasks we consider. We further push forward the state-of-the-art using a newly collected text corpus comprised
of news-like articles we call \gnews, 
including the highly competitive XSum and CNN/DailyMail summarization datasets.

Large-scale document-summary datasets are rare and in practice there is a mismatch
between research datasets and real-world use-cases where collecting summaries is 
expensive; the most common setting is that of low-resource summarization. We simulate
this setting and show that our model is able to adapt very quickly when fine-tuning with small numbers
of supervised pairs, obtaining state-of-the-art results in 6 datasets with only 1000 examples.

Qualitatively we observed high quality outputs from our best models
and validated this in human evaluation studies. We found that PEGASUS
summaries are at least as good as reference summaries for the datasets
we assessed -- XSum, CNN/DailyMail, and Reddit TIFU -- even at low-levels
of supervision.

To summarize our contributions:
\begin{itemize}
    \item 
    We propose a new self-supervised pre-training objective for abstractive summarization, gap-sentences generation, and study strategies for selecting
    those sentences.
    \item We evaluate the proposed pre-training objective on a broad range of downstream summarization tasks, with careful
    ablations to choose the best model settings, which we use to train a \largemodelsize parameter PEGASUS model that surpasses or is on-par with the state-of-the-art on all 12 downstream datasets considered.
    
    \item We show how good abstractive summarization performance can be achieved
    across broad domains with very little supervision by fine-tuning
    the PEGASUS model and surpassing previous state-of-the-art results on many tasks with
    as little as 1000 examples.
    
    \item We conducted human evaluation studies to validate our
    experimental design and demonstrate
    human-level summarization performance on XSum, CNN/DailyMail, and Reddit TIFU.
\end{itemize}



\section{Related Work}
\label{sec:related_works}

\citet{dai_2015, ramachandran-etal-2017-unsupervised} used
LM and autoencoder pre-training on in-domain data to improve performance
of RNN sequence models.
However, the combination of pre-training with much larger
external text corpora (such as Wikipedia, books, or Web-pages) and Transformer-based sequence models
has led to a dramatic improvement in performance when
fine-tuned for
both natural language understanding and text generation tasks \citep{radford2018improving,devlin_2019,rothe2019leveraging,yang2019xlnet,joshi2019spanbert,song2019mass,unilm,lewis2019bart}.
Most similar to our approach are Transformer encoder-decoder models pre-trained on some masked input pre-training objective.

\paragraph{MASS} \citep{song2019mass} proposed masked sequence-to-sequence generation that reconstructs a sentence fragment given the remaining part of the sentence. A single sentence fragment was randomly selected.

\paragraph{UniLM} \citep{unilm} proposed jointly training on three types of language modeling tasks: unidirectional (left-to-right and right-to-left), bidirectional (word-level mask, with next sentence prediction), and sequence-to-sequence (word-level mask) prediction.

\paragraph{T5}\citep{raffel2019exploring} generalized the text-to-text framework to a variety of NLP tasks and showed the advantage of scaling up model size (to 11 billion parameters) and pre-training corpus, introducing C4,
a massive text corpus derived from Common Crawl, which we also use in some of our 
models. T5 was pre-trained with randomly corrupted text spans of varying mask ratios and sizes of spans.  

\paragraph{BART} \citep{lewis2019bart} introduced a denoising autoencoder to pre-train sequence-to-sequence models. BART corrupted text with an arbitrary noising function and learned to reconstruct the original text. For generation tasks, the noising function was text infilling which used single mask tokens to mask random sampled spans of text. 

In contrast to MASS, UniLM, BART and T5, the proposed PEGASUS masks multiple whole sentences rather than smaller continuous text spans. In our final objective we deterministically
choose sentences based on importance, rather than randomly. As in T5, PEGASUS does not reconstruct full input sequences, and only generates
the masked sentences as a single output sequence.
In this work we focus entirely on downstream summarization (generative) tasks and
do not evaluate on NLU classification tasks.

There has been some work on the low-resource, summarization setting using
the CNN/DailyMail dataset.
\citet{gpt2} showed that a large Transformer language model pre-trained on Web text
could generate summaries if prompted with "TL;DR", achieving a ROUGE-2 of 8.27 
on CNN/DailyMail. \citet{khandelwal2019sample}
pre-trained a Transformer language model on Wikipedia, and fine-tuned
using 3000 examples, achieving 13.1 ROUGE-2.



\section{Pre-training Objectives}
\label{sec:methodology}

We propose a new pre-training objective, GSG,
in this work, but
for comparison, we also evaluate BERT's masked-language model objective, in isolation
and in conjunction with GSG.

\subsection{Gap Sentences Generation (GSG)}
\label{sec:gsg}
We hypothesize that using a pre-training
objective that more closely resembles
the downstream task leads to better
and faster fine-tuning performance.
Given our intended use for abstractive summarization, our proposed pre-training
objective involves
generating summary-like text from
an input document. In order to leverage
massive text corpora for pre-training, 
we design a sequence-to-sequence  self-supervised objective
in the absence of abstactive summaries.
A naive option
would be to pre-train as an extractive
summarizer; however, such a procedure would
only train a model to copy sentences, thus not suitable for abstractive summarization.

Inspired by recent success in masking
words and contiguous spans \citep{joshi2019spanbert,raffel2019exploring}, 
we select and mask whole sentences from
documents, and concatenate the gap-sentences
into a pseudo-summary. 
 The corresponding position of each selected gap sentence is replaced by a mask token \texttt{[MASK1]} to inform the model. \emph{Gap sentences ratio}, or \emph{GSR},
 refers to the number of selected gap sentences to the total number of sentences in the document, which is similar to \emph{mask rate} in other works.
 
 To even
more closely approximate a summary,
we select sentences that appear to be important/principal to the document.
The resulting objective has both the
empirically
demonstrated benefits of masking, and
anticipates the form of the downstream task.

We consider 3 primary strategies for selecting  $m$ gap sentences without
replacement from a document, $D=\{x_i\}_n$, comprised
of $n$ sentences: 

\paragraph{Random} Uniformly select $m$ sentences at random.

\paragraph{Lead} Select the first $m$ sentences.

\paragraph{Principal} Select top-$m$ scored
sentences according to importance.  As a proxy
for importance we compute ROUGE1-F1 \citep{lin-2004-rouge}  between
the sentence and the rest of the document,
$
    s_i=rouge(x_i, D \setminus \{x_i\}), \forall i
$.

In this formulation sentences are scored independently (\textbf{Ind}) and the top $m$ selected.
We also consider selecting them sequentially (\textbf{Seq}) as in \citet{nallapati2017extractive}
by greedily maximizing the ROUGE1-F1 between
selected sentences, ${S \cup \{x_i\}}$, and  remaining sentences, ${D \setminus (S \cup \{x_i\})}$ as in Algorithm \ref{algo:seq}.

\begin{algorithm} \caption{Sequential Sentence Selection}
\label{algo:seq}
\begin{algorithmic}[1]
\STATE  $S:=\emptyset$
\FOR{$j \gets 1$ to $m$}  
    \STATE $s_i := rouge\Big(S \cup \{x_i\}, D \setminus (S \cup \{x_i\})\Big)$ \par
    $\forall i~~s.t.~~x_i \notin S$
    \STATE $k := \argmax_i \{s_i\}_n$
    \STATE $S := S \cup \{x_k\}$ 
\ENDFOR
\end{algorithmic}
\end{algorithm}

When calculating ROUGE1-F1, we also consider n-grams as a set (\textbf{Uniq}) instead of double-counting identical n-grams
as in the original implementation (\textbf{Orig}). 
This results in four variants of the principal sentence selection strategy, choosing Ind/Seq and Orig/Uniq options. 

An example containing lead, random and principal gap sentence selection strategies are shown in Figure~\ref{fig:sentence_selection_example}. 

\begin{figure}[h!]
\centering
\begin{mdframed}
\small
\textcolor{amaranth}{INVITATION ONLY We are very excited to be co-hosting a major drinks reception with our friends at Progress. This event will sell out, so make sure to register at the link above.}
\textcolor{blue}{Speakers include Rajesh Agrawal, the London Deputy Mayor for Business, Alison McGovern, the Chair of Progress, and Seema Malhotra MP.}
\textcolor{darkgreen}{Huge thanks to the our friends at the ACCA, who have supported this event.} 
The Labour Business Fringe at this year's Labour Annual Conference is being co-sponsored by Labour in the City and the Industry Forum. 
\textcolor{blue}{Speakers include John McDonnell, Shadow Chancellor, and Rebecca Long-Bailey, the Shadow Chief Secretary to the Treasury, and our own Chair, Kitty Ussher.} 
\textcolor{darkgreen}{Attendance is free, and refreshments will be provided.}
\end{mdframed}
\caption{An example of sentences (from the C4 corpus) selected by \textcolor{darkgreen}{Random}, \textcolor{amaranth}{Lead} and \textcolor{blue}{Ind-Orig} respectively. Best viewed in color.}
\label{fig:sentence_selection_example}
\end{figure}

\subsection{Masked Language Model (MLM)}
\label{sec:mlm}
Following BERT, we select 15\% tokens in the input text, and the selected tokens are (1) 80\% of time replaced by a mask token \texttt{[MASK2]},
or (2) 10\% of time replaced by a random token, or (3) 10\% of time unchanged. 
We apply MLM to train the Transformer encoder as the sole pre-training objective or along with GSG. When MLM is the sole pre-training objective, the Transformer decoder shares all parameters with encoder when fine-tuning on downstream tasks following \citet{rothe2019leveraging}.

Figure~\ref{fig:architecture}
simultaneously shows how both
GSG and MLM are applied to the same example
when used in conjunction.
However, we found that MLM does not improve downstream tasks at large number of pre-training steps (section \ref{sec:exp_mlm}), and chose not to include MLM in the final model \pegasuslarge (section \ref{sec:exp_large_model}).

\section{Pre-training Corpus}
For pre-training we considered two large text corpora:
\begin{itemize}
    \item \textbf{C4}, or the Colossal and Cleaned version of Common Crawl,   introduced in \citet{raffel2019exploring}; consists of text from  350M Web-pages (750GB).
    \item  \textbf{\gnews}, a dataset of 1.5B articles (3.8TB) collected from news
    and news-like websites from 2013-2019. A whitelist of domains
    ranging from high-quality news publishers to lower-quality
    sites such as high-school newspapers, and blogs was 
    curated and used to seed a web-crawler. Heuristics were used to 
    identify news-like articles, and only the main article
    text was extracted as plain text.
\end{itemize}

\section{Downstream Tasks/Datasets}
\label{sec:exp_downstream_datasets}

For downstream summarization, we only used
public abstractive summarization datasets, and access
them through TensorFlow Summarization Datasets \footnote{\url{https://www.tensorflow.org/datasets/catalog/overview}}, which
provides publicly reproducible code for dataset processing and
train/validation/test splits. We used train/validation/test ratio of 80/10/10 if no split was provided, and 10\% train split as validation if there was no validation split.

\textbf{XSum} \citep{shashi-2018-xsum} consists of 227k
BBC articles from 2010 to 2017 covering a wide variety of subjects along with professionally written single-sentence summaries.

\textbf{CNN/DailyMail} \citep{hermann2015cnndm} dataset contains 93k articles from the CNN, and 220k articles the Daily Mail newspapers. Both publishers supplement their articles with bullet point summaries.
We use the non-anonymized variant used in \citet{see2017cnndm}.

\textbf{NEWSROOM} \citep{Grusky2018newsroom} is a large dataset containing 1.3M article-summary pairs written by authors and editors in the newsrooms of 38 major publications between 1998 and 2017.

\textbf{Multi-News} \citep{fabbri-etal-2019-multinews} is a multi-document summarization dataset consisting of 56k pairs of news articles and their human-written summaries from the site newser.com. 

\textbf{Gigaword} \citep{rush-etal-2015-neural} contains 4M
examples  extracted from news articles (seven publishers) from the Gigaword corpus \citep{graff2003gigaword}. The task is to generate
the headline from the first sentence.

\textbf{arXiv, PubMed} \citep{cohan-etal-2018-discourse-scientificpapers} are two long document datasets of scientific publications from arXiv.org (113k) and PubMed (215k). The task is to generate the abstract from the paper body.

\textbf{BIGPATENT} \citep{sharma-etal-2019-bigpatent} consists of 1.3 million  U.S. patents along with human summaries under nine patent classification categories.

\textbf{WikiHow} \citep{koupaee2018wikihow} is a large-scale dataset of instructions from the online WikiHow.com website.
Each of 200k examples consists of multiple instruction-step paragraphs along with a summarizing sentence. The task is to generate the concatenated summary-sentences from the paragraphs.

\textbf{Reddit TIFU} \citep{kim-etal-2019-reddittifu} contains  120K posts of informal stories from the online discussion forum Reddit, more specifically the TIFU sub-reddit from 2013-Jan to 2018-Mar. The sub-reddit posts strictly follow the rule of writing a descriptive "TL;DR" summary and has higher quality than \citep{volske-etal-2017-tl} (which used more subreddits) based on our manual inspection. We uses the TIFU-long subset (using TLDR as summaries) in the work.

\textbf{AESLC} \citep{zhang-tetreault-2019-aeslc} consists of 18k email bodies and their subjects from the Enron corpus \citep{Klimt:2004:ECN:3108498.3108522}, a collection of email messages of employees in the Enron Corporation.

\textbf{BillSum} \citep{kornilova-eidelman-2019-billsum} contains 23k US Congressional bills and human-written reference summaries from the 103rd-115th (1993-2018) sessions of Congress. We do not use the California test set
which is out-of-distribution.

Following \citeauthor{Grusky2018newsroom}, 
the number of examples and extractive fragment coverage/density for all downstream datasets is illustrated
in Appendix~\ref{appendix:dataset}. 

\section{Experiments}
\label{sec:experiments}
In a similar strategy to \citet{raffel2019exploring}, to save time and computation we conducted pre-training ablation experiments using a reduced-size model with \basemodelsize parameters, \textbf{\pegasusbase},  smaller batch size, and only 4 of 12 datasets
before scaling up pre-training with the best settings to the final \largemodelsize parameters, \textbf{\pegasuslarge}. The datasets (XSum, CNN/DailyMail, WikiHow and Reddit TIFU) were chosen for diversity in abstractiveness, writing style, and size.

\pegasusbase had $L=12, H=768, F=3072, A=12$ and \pegasuslarge had $L=16, H=1024, F=4096, A=16$, where $L$ denotes the number of layers for encoder and decoder (i.e. Transformer blocks), $H$ for the hidden size, $F$ for the feed-forward layer size and $A$ for the number of self-attention heads. 
We pre-trained \pegasusbase with a batch size of $256$ and \pegasuslarge with a batch size of $8192$. 
We refer to \pegasusbase without pre-training as \textbf{\transformerbase}.

We used sinusoidal positional encoding following ~\citet{vaswani2017attention}.
For optimization, both pre-training and fine-tuning used Adafactor \citep{shazeer2018adafactor} with square root learning rate decay and dropout rate of 0.1.

We used greedy-decoding for studies in Section \ref{sec:ablations}, and used beam-search with a length-penalty, $\alpha$, as in \citet{wu2016wordpiece} for
the final large model.

All experiments' hyper parameters can be found in Appendix~\ref{appendix:hparams}
and reported numbers are in Appendix~\ref{appendix:exps} and \ref{appendix:low_resource}.

\subsection{Ablations on \pegasusbase}
\label{sec:ablations}
We used \pegasusbase to evaluate choices of pre-training corpus, pre-training objective, and vocabulary size.
For reproducibility, we evaluated the latter two using the publicly available C4 corpus.

Note that the y-axis in  Figures~\ref{fig:exp_copora},~\ref{fig:exp_gsg},~\ref{fig:exp_vocab} 
are normalized by the left-most bar using 
${\frac{1}{3}(\frac{R1}{R1_\text{base}} + \frac{R2}{R2_\text{base}} + \frac{RL}{RL_\text{base}})}$
where $R1$, $R2$, $RL$ are ROUGE F1 scores and $R1_\text{base}$, $R2_\text{base}$, $RL_\text{base}$ are the scores of the configuration corresponding to the first bar.

With more pre-training steps, the model observed more documents in the pre-training corpus. A \pegasusbase model trained for 500k (highest we tried) steps did not observe all training examples on C4 nor \gnews.   
Appendix~\ref{appendix:pre-train_steps} shows the number of pre-training steps had an unsurprisingly positive impact on downstream dataset performance.
We used 500k steps for the ablation studies and the large model.

\subsubsection{Pre-training Corpus}
\label{sec:exp_pre-training_corpus}

\begin{figure}[tbh]
\centering
\includegraphics[width=0.45\textwidth]{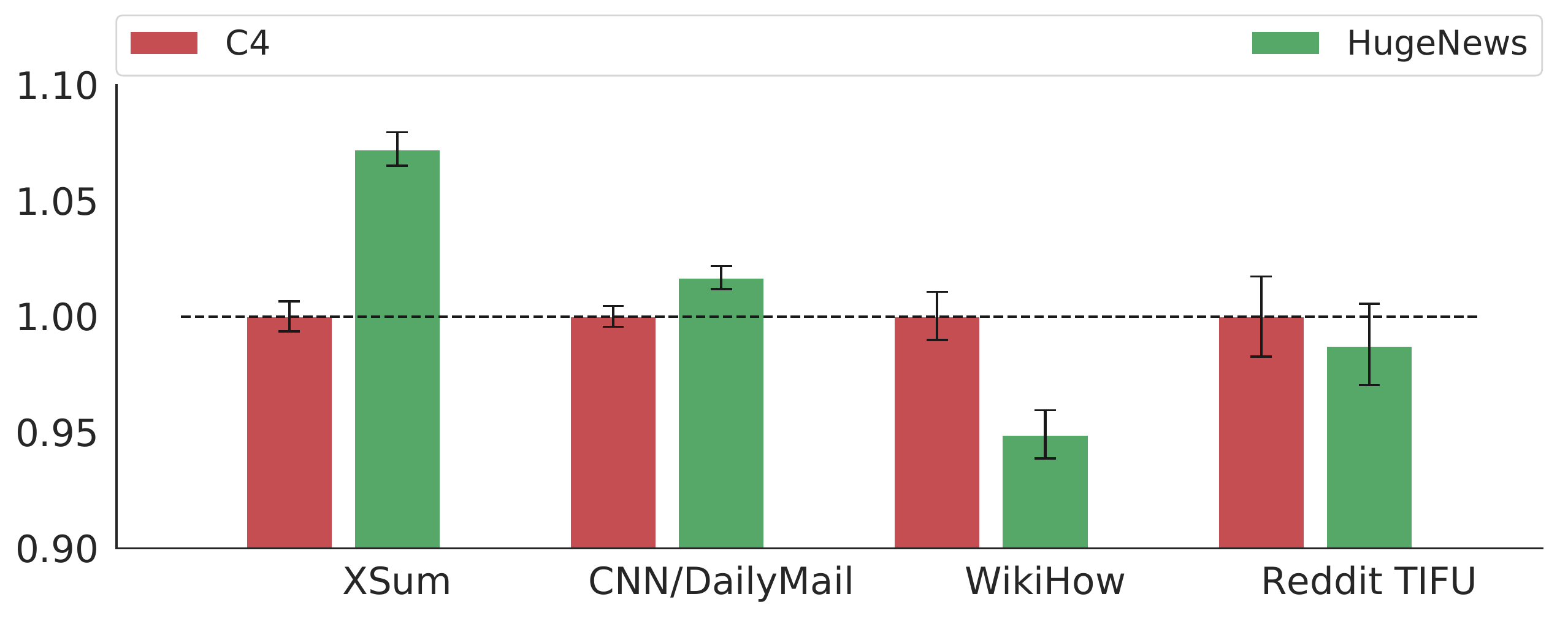}
\caption{Effect of pre-training corpus. \pegasusbase pre-trained on C4 (350M Web-pages) and \gnews (1.5B news-like documents).
}
\label{fig:exp_copora}
\end{figure}


Figure \ref{fig:exp_copora} shows that pre-training on \gnews was more effective than C4 on the two news downstream datasets, while the non-news informal datasets (WikiHow and Reddit TIFU) prefer the pre-training on C4. This suggests pre-training models transfer more effectively to downstream tasks when their domains are aligned better.

\subsubsection{Effect of Pre-training Objectives}
\label{sec:exp_pre-training_objectives}

\begin{figure*}[tbh]
\centering
\begin{subfigure}[t]{0.48\textwidth}
\includegraphics[width=\textwidth]{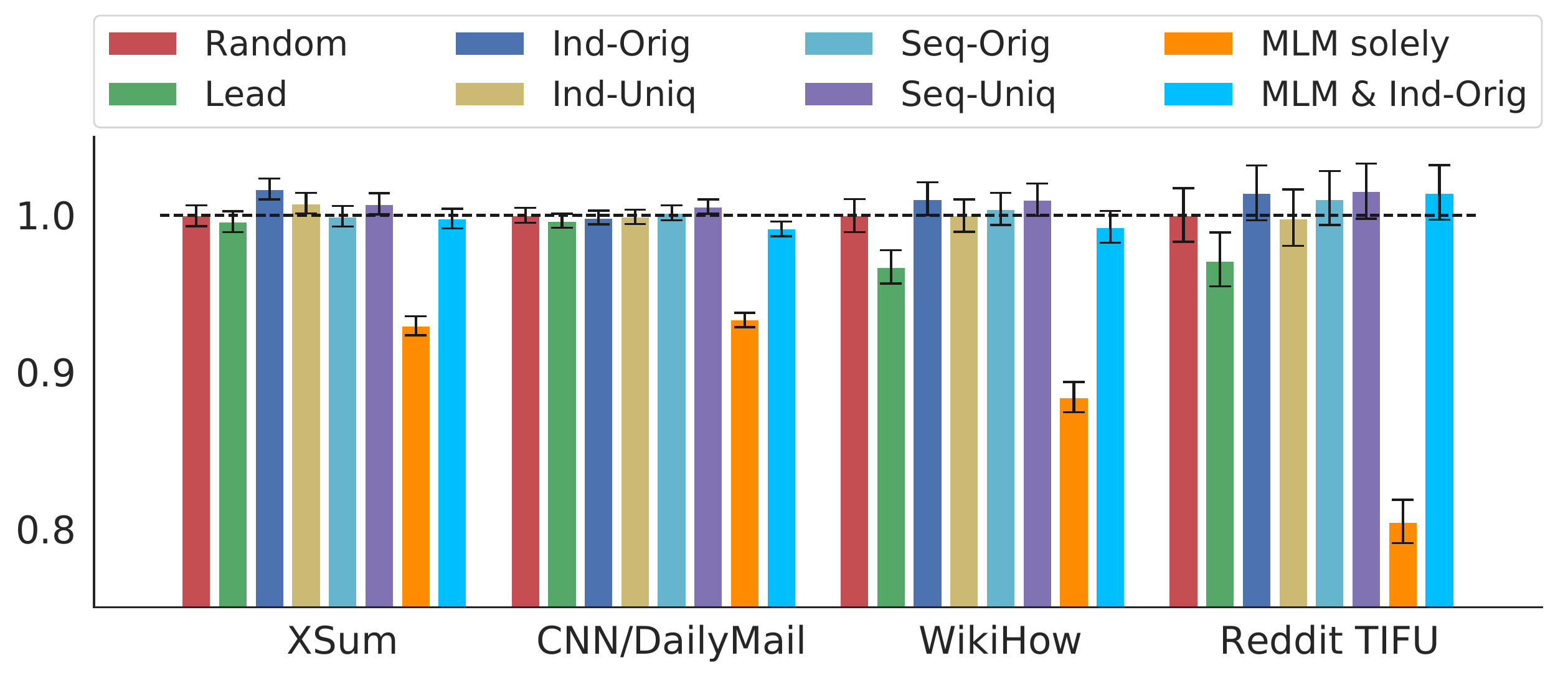}
\caption{Effect of pre-training objectives (30\% GSR). }
\label{fig:exp_objectives}
\end{subfigure}
\hfill
\begin{subfigure}[t]{0.48\textwidth}
\includegraphics[width=\textwidth]{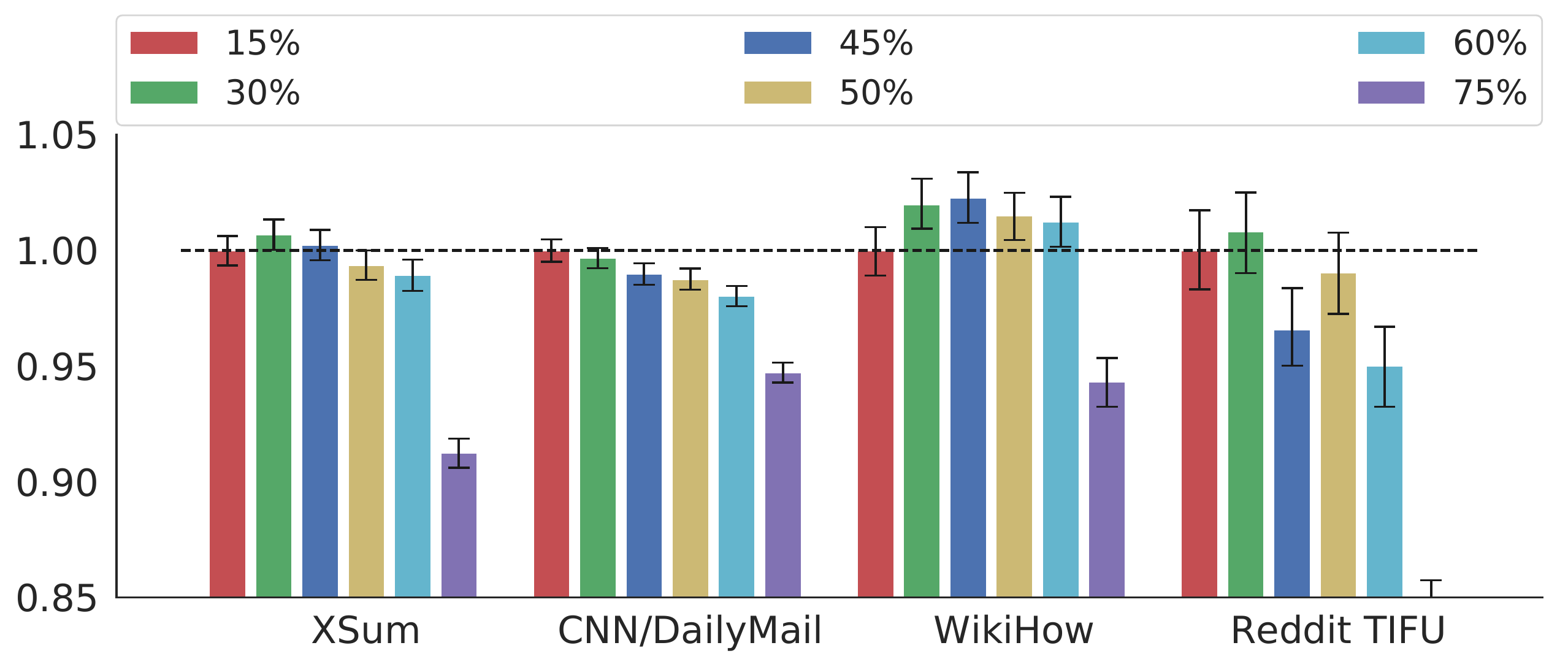}
\caption{Effect of gap sentences ratio  with GSG (Ind-Orig). }
\label{fig:exp_masking_rate}
\end{subfigure}
\caption{Effect of pre-training settings with \pegasusbase pre-trained on C4. }
\label{fig:exp_gsg}
\end{figure*}

\paragraph{GSG}\label{sec:exp_gsg}
We compared six variants of GSG (Lead, Random, Ind-Orig, Ind-Uniq, Seq-Orig, Seq-Uniq) while choosing 30\% sentences as gap sentences. As shown in Figure \ref{fig:exp_objectives},
Ind-Orig achieved the best performance followed by Seq-Uniq. Ind-Orig and Seq-Uniq were consistently better (or similar) than Random and Lead across the four downstream datasets.
Lead had decent performance on the two news datasets but was significantly worse on the two non-news datasets, which agrees findings of lead bias in news datasets \citep{see2017cnndm, Zhong_2019}. 
The results suggest choosing principal sentences works best for
downstream summarization tasks, and we chose Ind-Orig for the \pegasuslarge.

\label{sec:exp_masking_rate}
A significant hyper-parameter in GSG is the gap-sentences ratio (GSR). A low GSR makes the pre-training less challenging and computationally efficient. On the other hand, choosing gap sentences at a high GSR loses contextual information necessary to guide the generation. 
We compared GSRs from 15\% to 75\%. For a fair comparison, the original documents were truncated to have up to 400 words. The \emph{maximum input length}, $L_{input}$ in the encoder and the \emph{maximum target length}, $L_{target}$ in the decoder were set as 512 tokens.

Figure \ref{fig:exp_masking_rate} shows that different downstream datasets had slightly different optima. The best performance always had GSR lower than 50\%. The model with 15\% gap sentences achieved the highest ROUGE scores on CNN/DailyMail, while XSum/Reddit TIFU and WikiHow did better with 30\% and 45\% respectively.
When scaling up to \pegasuslarge (Section~\ref{sec:exp_large_model}), we chose an effective GSR of 30\%.  

\paragraph{MLM}\label{sec:exp_mlm}
As mentioned, the MLM objective can either be applied solely or together with GSG. 
We jointly trained MLM with GSG Ind-Orig (MLM \& Ind-Orig), which masks 30\% sentences and extra 15\% tokens in unselected sentences, as shown in Figure \ref{fig:architecture}.
Figure \ref{fig:exp_objectives} shows that the model pre-trained with MLM alone performed significantly worse  and MLM \& Ind-Orig had similar performance as Random. 
Interestingly, when comparing MLM \& Ind-Orig to Ind-Orig, we empirically observed MLM improved fine-tuning performance at early pre-training checkpoints (100k - 200k steps), but inhibited further gains with more pre-training steps (500k). Therefore, we chose not to include MLM in \pegasuslarge.

\subsubsection{Effect of Vocabulary}
\label{sec:exp_vocab}

\begin{figure}[tbh]
\centering
\includegraphics[width=0.45\textwidth]{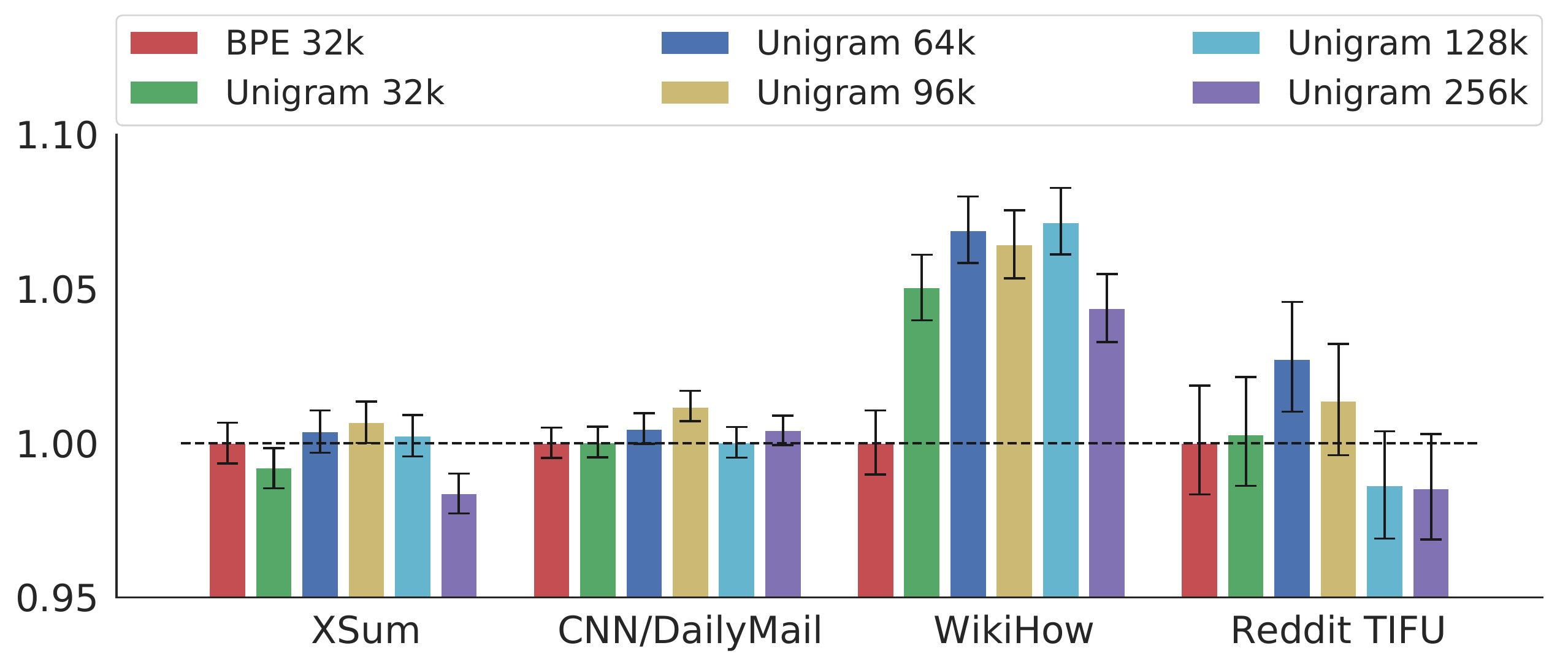}
\caption{Effect of vocabulary with \pegasusbase trained on C4 (15\% GSR, Ind-Orig). 
}
\label{fig:exp_vocab}
\end{figure}

We compared two tokenization methods\footnote{Implemented in https://github.com/google/sentencepiece}: Byte-pair-encoding algorithm (\textbf{BPE}) \citep{wu2016wordpiece,sennrich-etal-2016-neural}, and SentencePiece Unigram algorithm (\textbf{Unigram}) proposed in \citet{kudo2018sentencepiece}. We evaluated Unigram with different vocabulary sizes ranging from 32k to 256k. In these experiments, models were pre-trained for 500k steps on the C4 corpus with the Ind-Orig objective and 15\% GSR. As shown in Figure \ref{fig:exp_vocab}, BPE and Unigram were comparable on news datasets while Unigram outperformed BPE on non-news datasets, especially WikiHow. 
On XSum and CNN/DailyMail, Unigram 96k achieved the highest ROUGE scores. On WikiHow and Reddit TIFU, the best configurations were Unigram 128k and 64k respectively. 
Therefore, we used the overall best vocabulary option Unigram 96k in \pegasuslarge.

\begin{table*}[tb]
  \begin{minipage}{\textwidth}
  \scriptsize
  \centering
  \renewcommand{\arraystretch}{1.3}
  \begin{tabular}{cc|ccc|cc}
    \hline
    R1/R2/RL        & \makecell{Dataset\\size}    &   \transformerbase  & \pegasusbase   & Previous SOTA     & \makecell{\pegasuslarge\\(C4)} & \makecell{\pegasuslarge\\(\gnews)} \\\hline
    XSum            & 226k  & 30.83/10.83/24.41 & 39.79/16.58/31.70 & 45.14/22.27/37.25 &
    45.20/22.06/36.99 &
    \textbf{47.21/24.56/39.25} \\
    CNN/DailyMail   & 311k  & 38.27/15.03/35.48 & 
    41.79/18.81/38.93 & \textbf{44.16}/21.28/40.90 &
    43.90/21.20/40.76 & \textbf{44.17/21.47/41.11} \\
    NEWSROOM        & 1212k & 40.28/27.93/36.52 & 42.38/30.06/38.52 & 39.91/28.38/36.87 &
    \textbf{45.07/33.39/41.28} &    \textbf{45.15/33.51/41.33} \\
    Multi-News      & 56k   & 34.36/5.42/15.75  & 42.24/13.27/21.44 & 43.47/14.89/17.41 & 46.74/17.95/24.26 & \textbf{47.52/18.72/24.91} \\
    Gigaword        & 3995k & 35.70/16.75/32.83 & 36.91/17.66/34.08 & \textbf{39.14/19.92/36.57} & 38.75\textbf{/19.96}/36.14 & \textbf{39.12/19.86}/36.24 \\ \hline
    WikiHow         & 168k  & 32.48/10.53/23.86 & 36.58/15.64/30.01 & 28.53/9.23/26.54  & \textbf{43.06/19.71/34.80} & 41.35/18.51/33.42 \\
    Reddit TIFU     & 42k   & 15.89/1.94/12.22  & 24.36/6.09/18.75  & 19.0/3.7/15.1     & \textbf{26.54/8.94/21.64}  & \textbf{26.63/9.01/21.60} \\ 
    BIGPATENT       & 1341k & 42.98/20.51/31.87 & 43.55/20.43/31.80 & 37.52/10.63/22.79 & \textbf{53.63/33.16/42.25} & 53.41/32.89/42.07 \\
    arXiv           & 215k  & 35.63/7.95/20.00  & 34.81/10.16/22.50 & 41.59/14.26/23.55 & \textbf{44.70/17.27/25.80} & \textbf{44.67/17.18/25.73} \\
    PubMed          & 133k  & 33.94/7.43/19.02  & 39.98/15.15/25.23 & 40.59/15.59/23.59 & \textbf{45.49/19.90/27.69} & 45.09/19.56/27.42 \\ 
    AESLC           & 18k   & 15.04/7.39/14.93  & 34.85/18.94/34.10 & 23.67/10.29/23.44 & \textbf{37.69/21.85/36.84} & 37.40/21.22/36.45 \\
    BillSum         & 24k   & 44.05/21.30/30.98 & 51.42/29.68/37.78 & 40.80/23.83/33.73 & \textbf{57.20}/39.56/\textbf{45.80} & \textbf{57.31/40.19/45.82} \\ \hline 
  \end{tabular}
  \end{minipage}
  \caption{Results of \pegasuslarge and \pegasusbase on all downstream datasets compared with the previous SOTA, which are fetched from \cite{lewis2019bart,shi-etal-2019-leafnats,fabbri-etal-2019-multinews,koupaee2018wikihow,kim-etal-2019-reddittifu,subramanian2019extractive,song2019mass,zhang-tetreault-2019-aeslc,kornilova-eidelman-2019-billsum}. We only compared with previous abstractive models except on BillSum which had extractive results only. BIGPATENT, arXiv, PubMed and Multi-News datasets contain very long summaries and we truncate them to 256 tokens, in similar range compared to   \citep{sharma-etal-2019-bigpatent,cohan-etal-2018-discourse-scientificpapers,fabbri-etal-2019-multinews, goodman2019multistage}. Best ROUGE numbers on each dataset and numbers within $0.15$ of the best numbers are bolded.}
  \label{tab:large_model}
\end{table*}

\begin{table*}[tb]
  \scriptsize
  \centering
  \renewcommand{\arraystretch}{1.3}
  \begin{tabular}{c|ccc}
    \hline
    R1/R2/RL & XSum & CNN/DailyMail & Gigaword \\\hline 
    BERTShare \cite{rothe2019leveraging}    &  38.52/16.12/31.13 & 39.25/18.09/36.45 & 38.13/19.81/35.62 \\
    MASS \cite{song2019mass}                &  39.75/17.24/31.95 & 42.12/19.50/39.01 & 38.73/19.71/35.96 \\
    UniLM \cite{unilm}                      &  - & 43.33/20.21/40.51 & 38.45/19.45/35.75 \\
    BART \cite{lewis2019bart}               &  45.14/22.27/37.25 & \textbf{44.16}/21.28/40.90 & - \\
    T5 \cite{raffel2019exploring}           &  - & 43.52/\textbf{21.55}/40.69 & - \\ \hline
    $\text{PEGASUS}_{\text{LARGE}}$ (C4)    &  45.20/22.06/36.99 & 43.90/21.20/40.76 & 38.75\textbf{/19.96/36.14}  \\
    $\text{PEGASUS}_{\text{LARGE}}$ (\gnews) &  \textbf{47.21/24.56/39.25} & \textbf{44.17/21.47/41.11} & \textbf{39.12/19.86/36.24} 

    \\\hline
  \end{tabular}
  \caption{A comparison of $\text{PEGASUS}_{\text{LARGE}}$ with other pretrained models on XSum, CNN/DailyMail and Gigaword. Best ROUGE numbers  and numbers within $0.15$ of the best numbers are bolded.}
  \label{tab:large_model_selected}
\end{table*}

\subsection{Larger Model Results}
\label{sec:exp_large_model}

Compared with \pegasusbase, the large model \pegasuslarge had increased capacity from larger hidden size ($H:768\rightarrow1024$, $F:3072\rightarrow4096$, $A: 12\rightarrow 16$), number of layers ($L:12\rightarrow16$) and traversed much more data, due to larger batch size ($B: 256\rightarrow8192$) (same number of pre-training steps, 500k).
We adopted the best practices found in the \pegasusbase ablation studies using the GSG (Ind-Orig) pre-training objective without MLM and Unigram vocabulary size of 96k. In total, \pegasuslarge had \largemodelsize parameters. 

To encourage the model to copy, which is an important aspect of the more extractive datasets, we 
left 20\% of selected sentences unchanged in the input instead of replacing with \texttt{[MASK1]}.
 We increased the GSR to 45\% to achieve a similar number of ``gaps'' as the optimal 30\% found above. We reported the performance of the models pre-trained on \gnews and C4 separately. We conducted a simple hyper-parameter sweep of learning rate
 and length penalty, $\alpha$,  when fine-tuning \pegasuslarge on each downstream dataset.

CNN/DailyMail, Multi-News, arXiv, PubMed, BIGPATENT datasets contain input documents longer than the maximum input length ($L_{input} = 512$ tokens) in pre-training. 
This would present a problem for position embeddings which would never be
updated for longer input lengths,
but 
we confirm
the postulation that sinusoidal positional encodings \citep{vaswani2017attention} generalize well when fine-tuning \pegasuslarge beyond the input lengths observed in training up to $L_{input}=1024$ tokens. 
Since average input length in BIGPATENT, arXiv, PubMed and Multi-News are well beyond 1024 tokens, further scaling up $L_{input}$ or applying a two-stage approach \citep{liu2018generating} may improve performance even more, although this is 
outside the scope of this work.

Tables \ref{tab:large_model} and \ref{tab:large_model_selected} show the performance improvements of \pegasusbase and \pegasuslarge on downstream datasets. 
While \pegasusbase exceeded current state-of-the-art on many datasets, \pegasuslarge 
achieved better than state-of-the-art results on all downstream datasets using \gnews,
although C4 performed better on WikiHow.

The improvement from a Transformer model without pre-training (\transformerbase) to \pegasuslarge was more significant on smaller datasets. For example, the ROUGE2-F1 scores nearly tripled on AESLC and quintupled on Reddit TIFU. The large jumps in performance suggest that small text summarization datasets benefit the most from pre-training. We further investigate low resource summarization in Section~\ref{sec:low_resource}.

\begin{figure*}[bthp!]
\centering
\includegraphics[trim=300 100 300 0,clip,width=\textwidth]{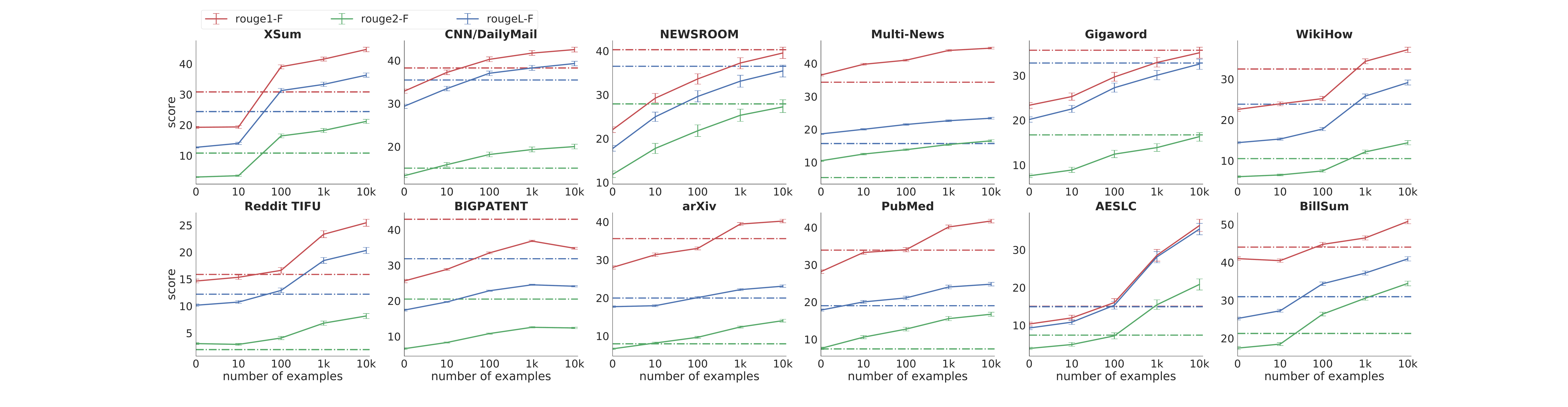}
\caption{Fine-tuning with limited supervised examples. The solid lines are \pegasuslarge fine-tuned on 0 (zero shot), 10, 100, 1k,10k examples. The dashed lines are \transformerbase models, equivalent in capacity as \pegasusbase and trained using the full supervised datasets, but with no pre-training. 
All numbers are reported in Appendix~\ref{appendix:low_resource}.
}
\label{fig:low_resource}
\end{figure*}

\subsection{Zero and Low-Resource Summarization}
\label{sec:low_resource}
In real-world practice, it is often difficult to collect a large number of supervised examples to train or fine-tune a summarization model. To simulate the low-resource summarization setting, we picked the first $10^k$ ($k=1,2,3,4$) training examples from each dataset to fine-tune \pegasuslarge (\gnews) .
We fine-tuned the models up to 2000 steps with batch size 256, learning rate 0.0005, and picked the checkpoint with best validation performance.
In Figure.~\ref{fig:low_resource}, in 8 out of 12 datasets, with just 100 examples  \pegasuslarge could be fine-tuned to generate summaries at comparable quality to \transformerbase trained on the full supervised datasets ranging from 20k to 200k examples. 
\pegasuslarge also beat previous state-of-the-art results on 6 out of 12 datasets with only
1000 fine-tuning examples.

On CNN/DailyMail, 
with half the number of parameters
\pegasuslarge demonstrated much better zero-shot
(ROUGE2-F=13.28)
performance than GPT-2 (ROUGE2-F=8.27).
Using only 1000 examples, \pegasuslarge achieved ROUGE2-F of 19.35,
much higher than the 13.1 obtained in  
\citet{khandelwal2019sample}  with 3000 examples.

\subsection{Qualitative Observations and Human Evaluation}
\begin{table}[]
\resizebox{1\columnwidth}{!}{%
\begin{tabular}[width=0.45\textwidth]{llll}
\hline
\textbf{Datasets} & \textbf{XSum}         & \textbf{CNN/DailyMail} & \textbf{Reddit TIFU}  \\
 & mean (p-value)         & mean (p-value) & mean (p-value)  \\ 
\hline
\hline
\textbf{Experiment 1: pretrain comparison} & &  \\  
Human-written       & 3.0 (-)        & 3.1 (-)        & 3.2 (-)        \\
\hline
\pegasuslarge (\gnews)       & \textbf{3.0} (0.6)      & \textbf{3.6} (0.0001) & \textbf{3.2} (0.7)      \\
\pegasuslarge(C4)          & \textbf{3.1} (0.7)      & \textbf{3.5} (0.009) & \textbf{3.1} (0.3)      \\
\transformerbase & 2.0   (3e-10) & \textbf{2.9} (0.06)     & {1.4} (5e-23)  \\
\hline 
\hline 
\textbf{Experiment 2: low resource} &         &  &   \\ 
Human-written       & 3.2 (-)        &  3.2(-)        & 3.3 (-)        \\
\hline
\pegasuslarge (\gnews) 10 examples      & \textbf{2.8} (0.1)      & \textbf{3.4} (0.007) & 2.6 (0.006)      \\
\pegasuslarge (\gnews) 100 examples               & \textbf{3.2} (0.5)      & \textbf{3.4} (0.08) & 2.1 (4e-8)      \\
\pegasuslarge (\gnews) 1000 examples      & \textbf{3.4} (0.3) & \textbf{3.6} (0.07)     & 2.7 (0.01)  \\
\pegasuslarge (\gnews) full supervision & \textbf{3.4} (0.3) & \textbf{3.3} (0.1) & \textbf{2.8} (0.05)\\
\hline

\end{tabular}
}

\caption{Human evaluation side-by-side results on Likert (1-5) scale (higher is better). 
Scores are bolded if they are not worse than human-level performance by $p<0.01$. 
}
\label{tab:human_eval_models}
\end{table}

Overall, we observed high-linguistic quality (in terms of fluency and coherence), closely emulating 
the style of ground-truth summaries.
While some previous work suggested that maximum likelihood training 
results in repetitive text in model outputs  \citep{welleck2019neural}
we found this to be rare in our outputs and did not require additional
counter-measures to mitigate dis-fluencies.

Although ROUGE clearly has its draw-backs \citep{kryscinski-etal-2019-neural},
over-penalizing abstractive approaches compared to extractive ones and having
no sense of linguistic quality,
we found that 
choosing perplexity-optimized models using aggregated ROUGE (rather than directly 
optimizing ROUGE as in \citet{paulus2017deep}) resulted in qualitatively 
good models.
Randomly sampled  (by a program) model decodes across all datasets and a broad range of
ROUGE scores can be found in Appendix~\ref{appendix:model_ouputs}.We found that even low-ROUGE model summaries often were high-quality, Figure \ref{fig:qualitative_cnn_example}.

To assess how close \pegasuslarge is to human performance we conducted human evaluation experiments 
on Amazon Mechanical Turk
comparing model summaries with (human) reference summaries given the input document. The examples were drawn from the XSum, CNN/DailyMail, and Reddit TIFU datasets; the first two were chosen due to their popularity in past work, and the third was chosen for its significant difference in style.
Workers were asked to rate the summaries on a 1-5 scale, with higher
being better (full experiment details provided in Appendix \ref{appendix:human})
and a paired t-test was used to assess whether scores were significantly different from human. 

In the first experiment,
\pegasuslarge(\gnews), \pegasuslarge(C4), and \transformerbase were
compared with reference summaries; in the second experiment,
\pegasuslarge(\gnews) fine-tuned using 10, 100, 1000, and all 
supervised examples were compared with references; the results are shown in Table \ref{tab:human_eval_models}.
According to the significance level of $p<0.01$, both \pegasuslarge(\gnews) and \pegasuslarge(C4) outputs were at least as good as the reference summaries in all cases.
Even at low-levels of supervision \pegasuslarge(\gnews) was not measurably worse than human summaries on XSum and CNN/DailyMail. 
In the Reddit TIFU case, however, perhaps due to its diverse writing styles,
human performance required full supervision.

\subsection{Test-set Overlap with Pre-training Corpus}

The pre-training corpora are a large collection of documents from the Internet and potentially have overlap with the downstream test sets. In this section, we measured the extent of overlap between the pre-training corpus and downstream datasets. We also studied if the pre-trained model was able to exploit memorization to achieve higher performance on the downstream datasets.

\begin{figure}[tbh]
\centering
\includegraphics[width=0.45\textwidth]{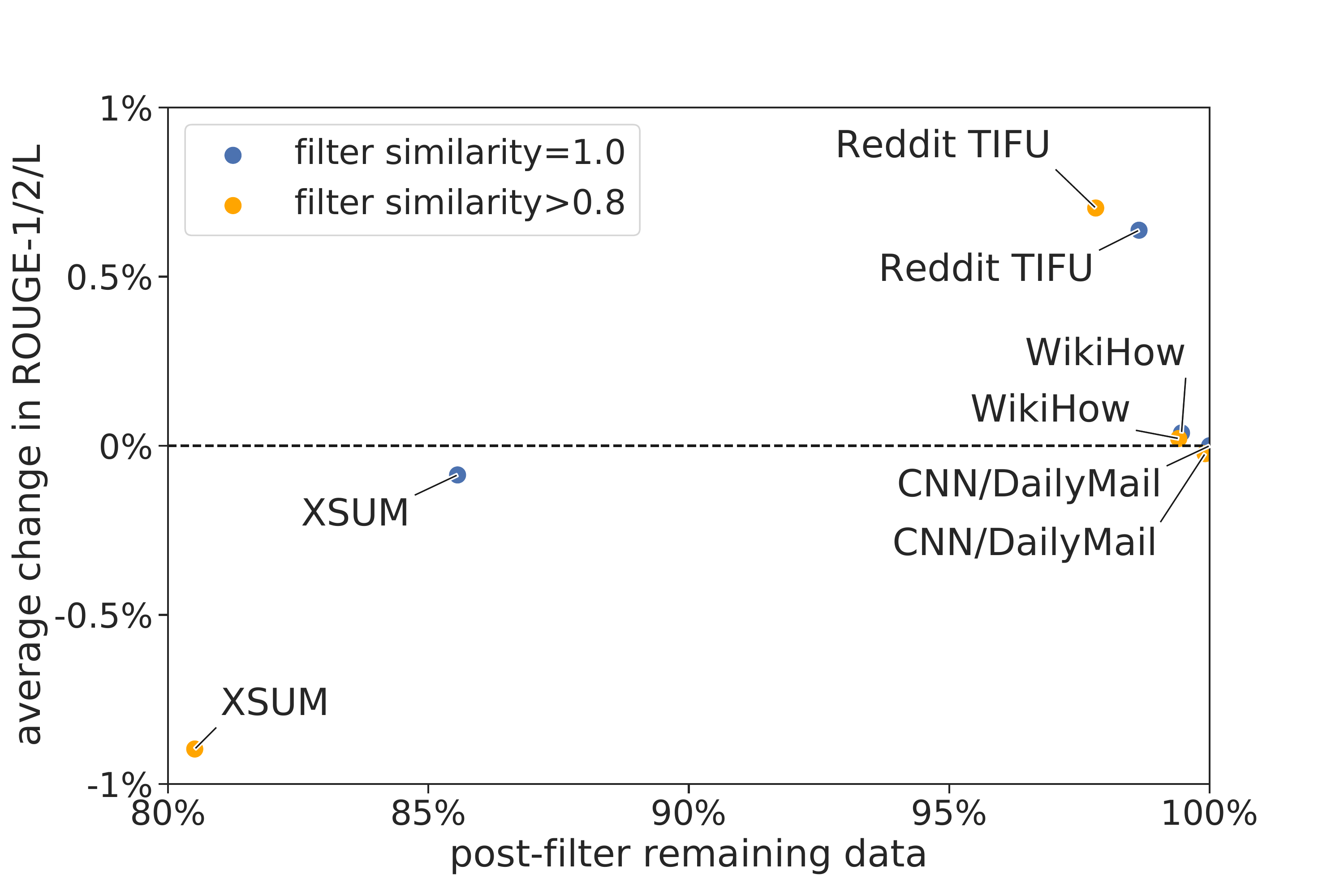}
\caption{Percentage of overlap between C4 and downstream test sets, and ROUGE score changes after removing those overlapped examples in test sets. 
}
\label{fig:filter}
\end{figure}

To measure the overlap, we calculated similarities between all pairs of downstream test set targets and pre-training documents. We use the ROUGE-2 recall as a similarity measure (common 2-grams / test set targets 2-grams). It is not necessarily exact match even if the similarity score is 1.0. We filtered all test set examples that have similarity to any pre-training example above a threshold, and recalculated the ROUGE scores on the remaining test set. In Figure~\ref{fig:filter}, we conducted this study on the pre-training corpus C4 and test set of XSum, CNN/Dailymail, Reddit TIFU and WikiHow, with a similarity threshold of 1.0 and 0.8. Results show that only XSum has significant amount of overlap 15\% to 20\%, and filtering those examples does not change ROUGE scores more than 1\%.
We also manually examined those overlapped examples with similarity of 1.0, and found that the models produce very different summaries compared to the human written ones, suggesting that there was no clear memorization.

\subsection{Additional \pegasuslarge Improvements}
Following our experiments on \pegasuslarge pre-trained on C4 and \gnews, 
we pre-trained a \pegasuslarge model on both corpora and stochastically sampled important sentences.
The \pegasuslarge (mixed,stochastic) model includes the changes:
(1) The model was pre-trained on the mixture of C4 and \gnews weighted by their number of examples.
(2) The model dynamically chose gap sentences ratio uniformly between 15\%-45\%.
(3) Importance sentences were stochastically sampled with 20\% uniform noise on their scores.
(4) The model was pre-trained for 1.5M steps instead of 500k steps, as we observed slower convergence of pre-training perplexity.
(5) The SentencePiece tokenizer was updated to encode the newline character.
The \pegasuslarge (mixed, stochastic) model achieved best results on almost all downstream tasks, as shown in Table~\ref{tab:mixed}.

\begin{table}[tb]
\begin{minipage}{.5\textwidth}
\scriptsize
\centering
\renewcommand{\arraystretch}{1.3}
\begin{tabular}{ccc}
\hline
XSum & CNN/DailyMail & NEWSROOM \\
47.60/24.83/39.64 & 44.16/21.56/41.30 & 45.98/34.20/42.18  \\ \hline
Multi-News & Gigaword & WikiHow \\ 
47.65/18.75/24.95 & 39.65/20.47/36.76 & 46.39/22.12/38.41 \\ \hline
Reddit TIFU & BIGPATENT & arXiv \\ 
27.99/9.81/22.94 & 52.29/33.08/41.66 $\ddagger$ & 44.21/16.95/25.67 \\ \hline
PubMed & AESLC & BillSum \\ 
45.97/20.15/28.25 & 37.68/21.25/36.51 & 59.67/41.58/47.59  \\ \hline
\end{tabular}
\end{minipage}
\caption{Results (ROUGE-1/ROUGE-2/ROUGE-L F scores) of \pegasuslarge (mixed, stochastic) on downstream datasets. $\ddagger$ 
We updated the BIGPATENT dataset to preserve casing, some format cleanings are also changed. }
\label{tab:mixed}
\end{table}

\section{Conclusion}
\label{sec:conclusion}
In this work, we proposed PEGASUS, a sequence-to-sequence model with gap-sentences generation as a pre-training objective tailored for abstractive text summarization.
We studied several gap-sentence selection methods and identified principle sentence selection as the optimal strategy.
We demonstrated the effects of the pre-training corpora, gap-sentences ratios, vocabulary sizes and scaled up the best configuration to achieve state-of-the-art results
on all 12 diverse downstream datasets considered.
We also showed that our model was able to adapt to unseen
summarization datasets very quickly, achieving strong results in as little as 1000 examples. We finally showed our model summaries achieved human performance on multiple datasets using human evaluation. 

\section{Code and Model Checkpoints Release}
The training code and instructions for using model checkpoints can be found at

\pegasuscode

\section*{Acknowledgments}
We thank Anastassia Kornilova, Eva Sharma, Shashi Narayan,
 Adam Roberts, Etienne Pot, and the Google News team for assistance with datasets, and
 Carey Radebaugh, David Grangier, Doug Eck, and Samy Bengio for reviewing the manuscript.  

\bibliography{icml2020}
\bibliographystyle{icml2020}

\clearpage
\appendix
\counterwithin{figure}{section}
\counterwithin{table}{section}

\onecolumn

\section{Datasets Statistics}
\label{appendix:dataset}

Following \citeauthor{Grusky2018newsroom}, we calculate extractive fragment coverage/density for all downstream datasets.
They were defined as
$$coverage=\frac{1}{S}\sum_{f\in F(A,S)}|f|$$ $$density=\frac{1}{S}\sum_{f\in F(A,S)}|f|^2$$
where $A$ is article, $S$ is summary, and $f \in F(A,S)$ are extractive fragments. High density indicates more extractive datasets and low coverage suggests more novel words in the summary.

\begin{figure*}[tbh]
\centering
\includegraphics[width=0.75\textwidth]{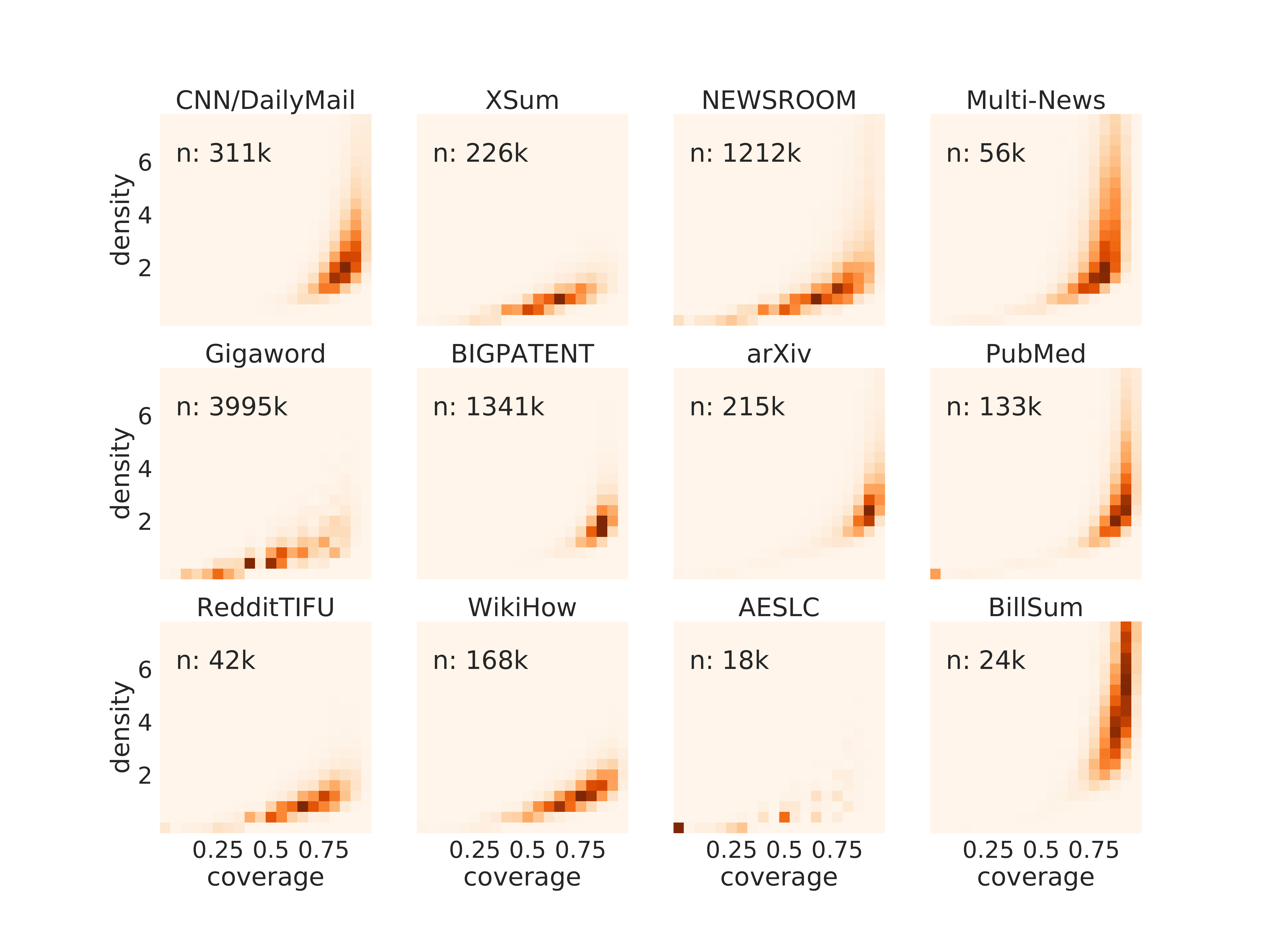}
\caption{A comparison of extractive fragment coverage and density of downstream datasets. The darker blocks indicate higher percentages and the n is the number of examples in the dataset.}
\label{fig:exp_downstream}
\end{figure*}

\section{Pre-training Steps}
\label{appendix:pre-train_steps}
\begin{figure*}[tbh]
\begin{subfigure}[b]{\textwidth}
\includegraphics[width=0.32\textwidth]{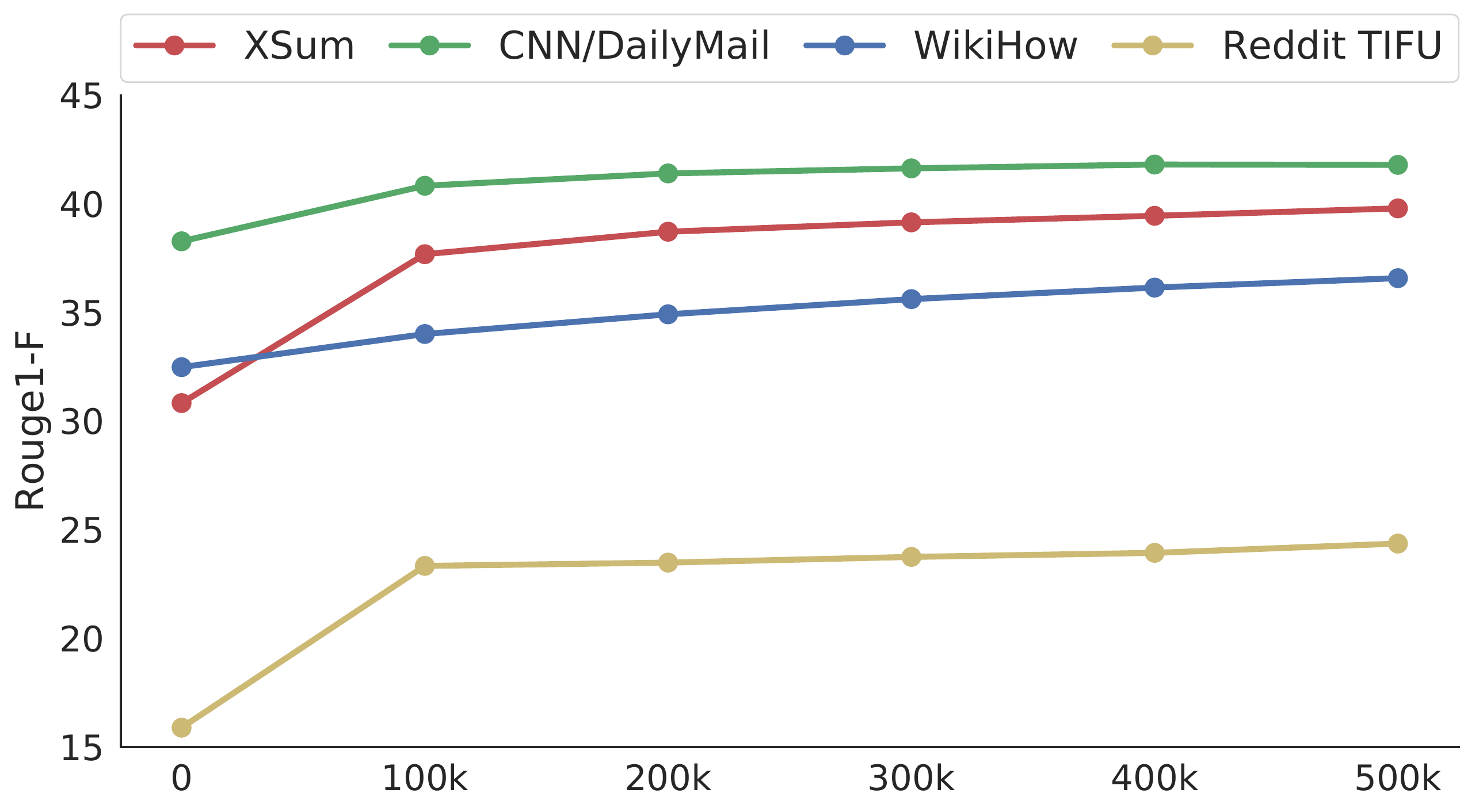}
\includegraphics[width=0.32\textwidth]{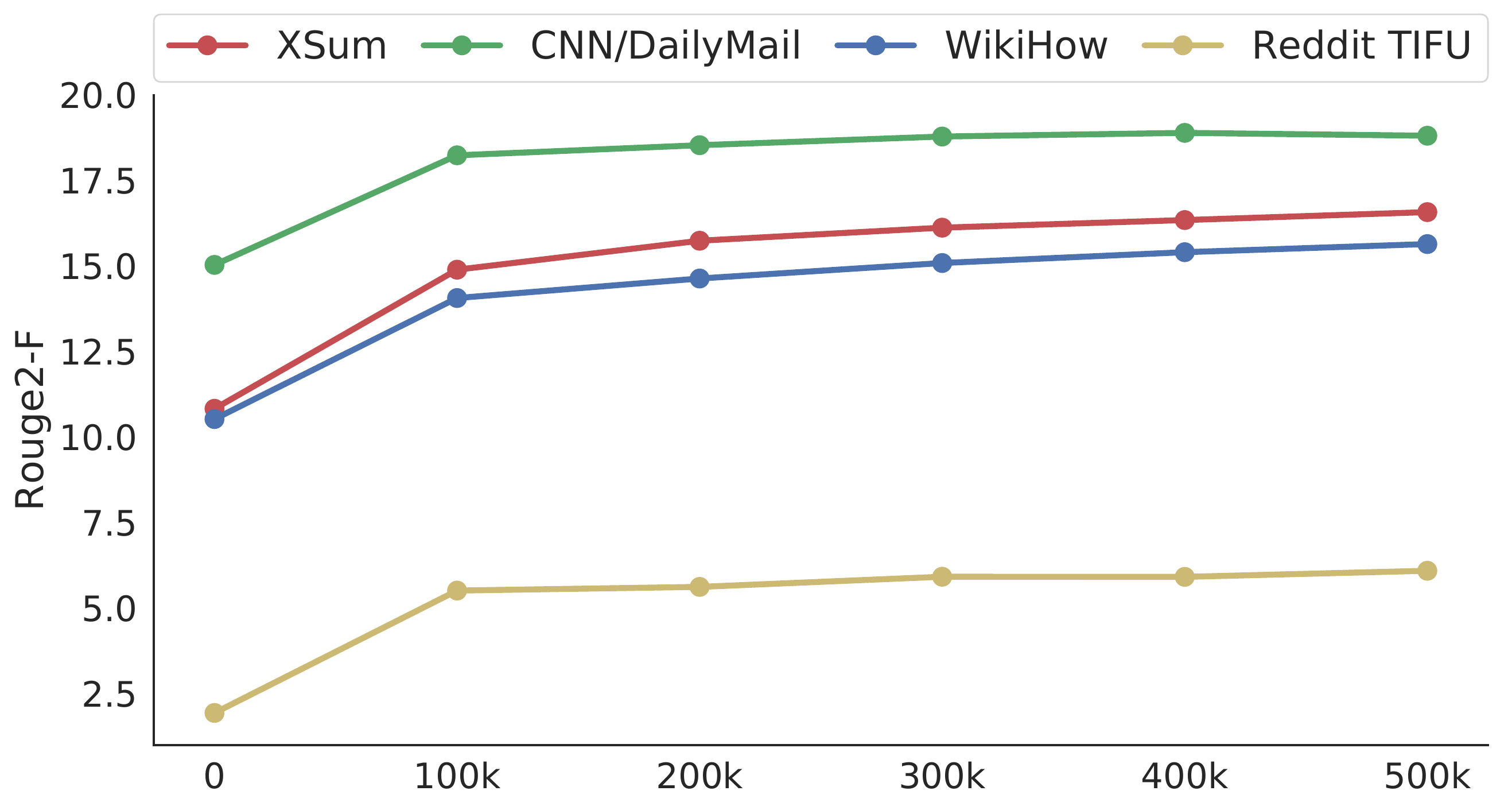}
\includegraphics[width=0.32\textwidth]{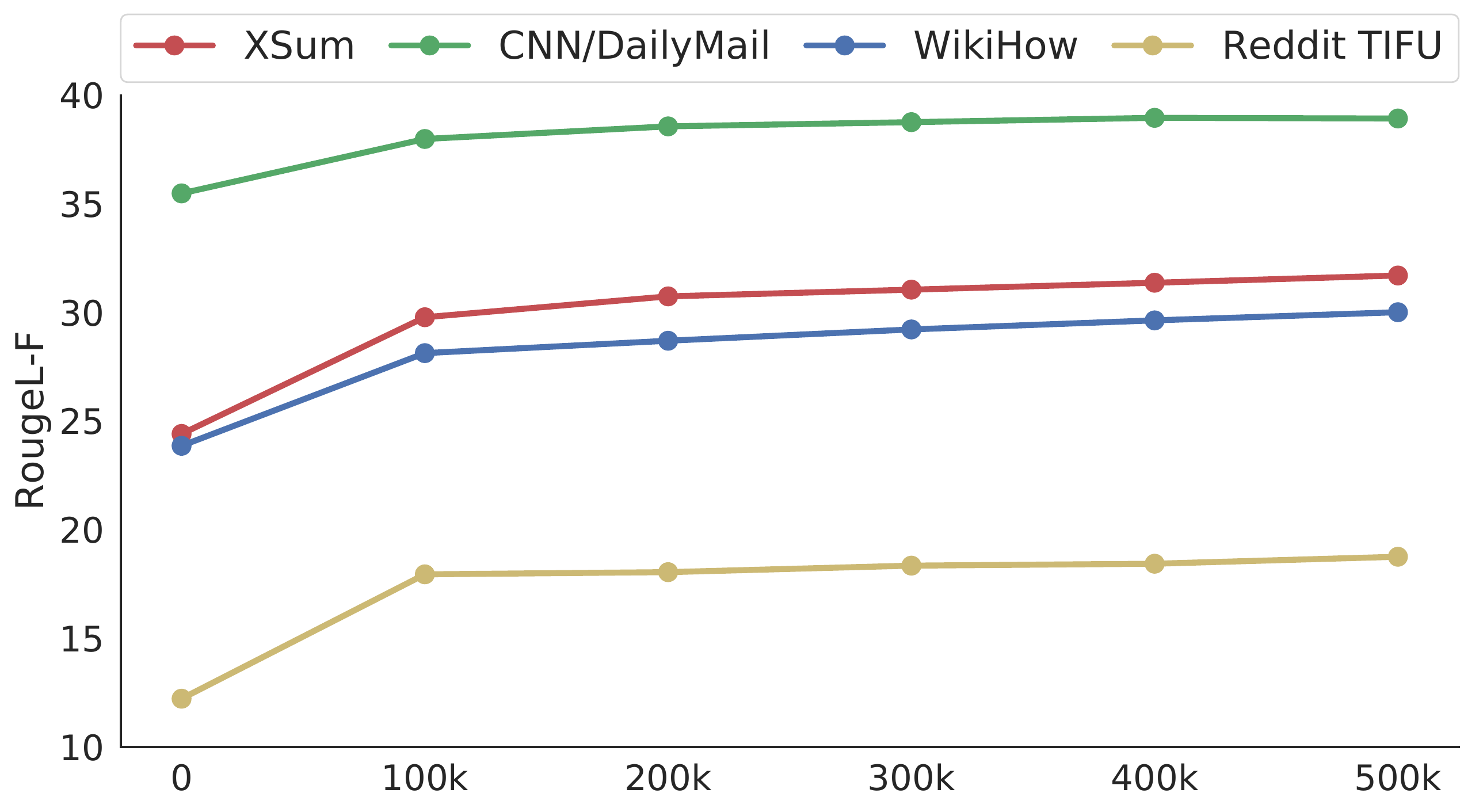}
\end{subfigure}
\caption{
Performance increase on downstream datasets as \pegasusbase trains for more steps on C4.}
\label{fig:exp_pre-training_steps}
\end{figure*}

\clearpage
\section{PEGASUS Hyper Parameters}
\label{appendix:hparams}
\begin{table*}[htb!]
  \scriptsize
  \centering
  \renewcommand{\arraystretch}{1.3}
  \begin{tabular}{ccccccccc}
    \hline
    \multicolumn{9}{c}{\textbf{Pre-training (default unless otherwise specified in section \ref{sec:experiments})}} \\\hline
    Model & \makecell{Learning \\ rate} & \makecell{Label \\ smoothing} & Num of steps & Batch size & Objective & Corpus & \makecell{Max input\\tokens} & \makecell{Max target\\tokens}\\\hline
    $\text{PEGASUS}_{\text{BASE}}$      & 0.1 & 0.0 & 500k & 256    & Ind-Orig & c4 & 512 & 256 \\
    $\text{PEGASUS}_{\text{LARGE}}$     & 0.1 & 0.0 & 500k & 8192   & Ind-Orig & c4 or \gnews & 512 & 256 \\\hline
    \multicolumn{9}{c}{\textbf{Fine-tuning of $\text{PEGASUS}_{\text{BASE}}$ in Figure \ref{fig:exp_copora}, \ref{fig:exp_gsg}, \ref{fig:exp_vocab},  \ref{fig:exp_pre-training_steps} and Table \ref{tab:large_model}}} \\\hline
    Dataset & \makecell{Learning \\ rate} & \makecell{Label \\ smoothing} & Num of steps & Batch size & Beam size & Beam alpha & \makecell{Max input\\tokens} & \makecell{Max target\\tokens}\\\hline
    XSum            & 5e-4 & 0.1 & 50k  & 256 & 1 & - & 512 & 64 \\
    CNN/DailyMail   & 5e-4 & 0.1 & 50k  & 256 & 1 & - & 512 & 128 \\
    NEWSROOM        & 5e-4 & 0.1 & 50k  & 256 & 1 & - & 512 & 128 \\
    Multi-News      & 5e-4 & 0.1 & 50k  & 256 & 1 & - & 512 & 256 \\
    WikiHow         & 5e-4 & 0.1 & 50k  & 256 & 1 & - & 512 & 256 \\
    Reddit TIFU     & 5e-4 & 0.1 & 50k  & 256 & 1 & - & 512 & 128 \\
    BIGPATENT       & 0.01 & 0.1 & 300k & 256 & 1 & - & 512 & 256 \\
    arXiv           & 5e-4 & 0.1 & 50k  & 256 & 1 & - & 512 & 256 \\
    PubMed          & 5e-4 & 0.1 & 50k  & 256 & 1 & - & 512 & 256 \\
    Gigaword        & 5e-4 & 0.1 & 50k  & 256 & 1 & - & 128 & 32 \\
    AESLC           & 5e-4 & 0.1 & 50k  & 256 & 1 & - & 512 & 32 \\
    BillSum         & 5e-4 & 0.1 & 50k  & 256 & 1 & - & 512 & 256 \\\hline
    \multicolumn{9}{c}{\textbf{\transformerbase in Table \ref{tab:large_model}}} \\\hline
    Dataset & \makecell{Learning \\ rate} & \makecell{Label \\ smoothing} & Num of steps & Batch size & Beam size & Beam alpha & \makecell{Max input\\tokens} & \makecell{Max target\\tokens}\\\hline
    BIGPATENT   & 0.01 & 0.1 & 300k & 256 & 1 & - & 512 & 256 \\
    AESLC       & 5e-4 & 0.1 & 300k & 256 & 1 & - & 512 & 32 \\
    Others      & 5e-3 & 0.1 & 300k & 256 & 1 & - & \multicolumn{2}{c}{Same as $\text{PEGASUS}_{\text{BASE}}$}  \\ \hline
    \multicolumn{9}{c}{\textbf{Fine-tuning of $\text{PEGASUS}_{\text{LARGE}}$ in Table \ref{tab:large_model} and \ref{tab:large_model_selected}}} \\\hline
    Dataset & \makecell{Learning \\ rate} & \makecell{Label \\ smoothing} & Num of steps & Batch size & Beam size & Beam alpha & \makecell{Max input\\tokens} & \makecell{Max target\\tokens}\\\hline
    XSum(C4)            & 1e-4  & 0.1 & 130k     & 256   & 8 & 0.8   & 512   & 64    \\
    XSum(\gnews)            & 1e-4  & 0.1 & 80k     & 256   & 8 & 0.8   & 512   & 64    \\
    CNN/DailyMail(C4)   & 5e-5  & 0.1 & 220k    & 256   & 8 & 0.8   & 1024   & 128   \\
    CNN/DailyMail(\gnews)   & 5e-5  & 0.1 & 170k    & 256   & 8 & 0.9   & 1024   & 128   \\
    NEWSROOM        & 4e-4  & 0.1 & 104k    & 256   & 8 & 0.8     & 512   & 128   \\
    Multi-News      & 5e-5  & 0.1 & 80k     & 256   & 8 & 0.9   & 1024   & 256   \\
    WikiHow         & 8e-4  & 0.1 & 50k     & 256   & 8 & 0.6   & 512   & 256   \\
    Reddit TIFU     & 1e-4  & 0.1 & 12k     & 256   & 8 & 0.6   & 512   & 128   \\
    BIGPATENT       & 5e-3  & 0.1 & 300k    & 256   & 8 & 0.7     & 1024  & 256   \\
    arXiv           & 8e-4  & 0.1 & 74k     & 256   & 8 & 0.8   & 1024  & 256   \\
    PubMed          & 2e-4  & 0.1 & 100k    & 256   & 8 & 0.8   & 1024  & 256   \\
    Gigaword        & 8e-4  & 0.1 & 90k     & 256   & 8 & 0.6   & 128   & 32    \\
    AESLC           & 2e-4  & 0.1 & 16k     & 256   & 8 & 0.6   & 512   & 32    \\
    BillSum         & 2e-4  & 0.1 & 100k    & 256   & 8 & 0.8   & 1024  & 256   \\\hline
    \multicolumn{9}{c}{\textbf{Fine-tuning of $\text{PEGASUS}_{\text{LARGE}}$ in Figure~\ref{fig:low_resource} }}\\\hline
    Dataset & \makecell{Learning \\ rate} & \makecell{Label \\ smoothing} & Num of steps & Batch size & Beam size & Beam alpha & \makecell{Max input\\tokens} & \makecell{Max target\\tokens}\\\hline
    all   & 5e-4 & 0.1 & 2k & 256 & 1 & - & \multicolumn{2}{c}{Same as $\text{PEGASUS}_{\text{BASE}}$} \\\hline
  \end{tabular}
  \caption{Hyperparamters of the pre-training and fine-tuning stages reported in section \ref{sec:experiments}. The hyperparameters of fine-tuning $\text{PEGASUS}_{\text{LARGE}}$ were decided by grid search while others were decided by empirically default commonly used values. Max input/target tokens correspond to $L_{input}$ and $L_{target}$ in Section~\ref{sec:experiments}.
  }
  \label{tab:hyperparamters}
\end{table*}

\clearpage
\section{Experiment Figures' Numbers}
\label{appendix:exps}
\begin{table*}[hbt!]
\small
  \centering
  \renewcommand{\arraystretch}{1.3}
  \begin{tabular}{ccccc}
    \hline
    \multicolumn{5}{c}{\textbf{ROUGE scores reported in Figure \ref{fig:exp_copora}}} \\\hline
    & XSum & CNN/DailyMail & WikiHow & Reddit TIFU \\
    & R1/R2/RL & R1/R2/RL &  R1/R2/RL & R1/R2/RL \\\hline 
    Pre-trained on c4    & 39.79/16.58/31.70 & 41.79/18.81/38.93 & 36.58/15.64/30.01 & 24.36/6.09/18.75 \\ 
    Pre-trained on \gnews & 41.63/18.47/33.48 & 42.34/19.22/39.49 & 34.93/14.67/28.63 & 24.11/5.99/18.57 \\  \hline
    \multicolumn{5}{c}{\textbf{ROUGE scores reported in Figure \ref{fig:exp_objectives}}} \\\hline
    & XSum & CNN/DailyMail & WikiHow & Reddit TIFU \\
    & R1/R2/RL & R1/R2/RL &  R1/R2/RL & R1/R2/RL \\\hline 
    Random      & 39.28/16.23/31.21 & 41.80/18.91/38.88 & 36.27/15.47/29.67 & 24.04/6.01/18.47 \\ 
    Lead        & 39.22/16.12/31.09 & 41.70/18.78/38.85 & 35.30/14.79/28.85 & 23.48/5.78/18.00 \\ 
    Ind-Orig   & 39.79/16.58/31.70 & 41.79/18.81/38.93 & 36.58/15.64/30.01 & 24.36/6.09/18.75 \\ 
    Ind-Uniq  & 39.50/16.41/31.41 & 41.79/18.83/38.94 & 36.26/15.47/29.69 & 24.10/5.98/18.41 \\ 
    Seq-Orig   & 39.22/16.27/31.11 & 41.88/18.89/39.02 & 36.39/15.57/29.74 & 24.09/6.15/18.55 \\ 
    Seq-Uniq  & 39.50/16.39/31.40 & 41.98/19.03/39.11 & 36.69/15.61/29.95 & 24.25/6.17/18.67 \\ 
    MLM solely  & 37.22/14.48/29.62 & 39.33/17.34/36.65 & 32.20/13.19/27.05 & 21.00/3.96/16.27 \\ 
    MLM \& Ind-Orig & 39.08/16.21/31.20 & 41.48/18.70/38.63 & 35.99/15.29/29.57 & 24.19/6.16/18.70 \\\hline 
    \multicolumn{5}{c}{\textbf{ROUGE scores reported in Figure \ref{fig:exp_masking_rate}}} \\\hline
    & XSum & CNN/DailyMail & WikiHow & Reddit TIFU \\
    & R1/R2/RL & R1/R2/RL &  R1/R2/RL & R1/R2/RL \\\hline 
    15\%  & 39.47/16.32/31.30 & 41.88/18.98/38.97 & 35.63/15.08/29.23 & 24.06/5.91/18.52 \\ 
    30\%   & 39.61/16.51/31.48 & 41.83/18.82/38.96 & 36.26/15.47/29.69 & 24.05/6.05/18.55 \\ 
    45\%  & 39.43/16.42/31.36 & 41.57/18.67/38.69 & 36.39/15.46/29.85 & 23.47/5.61/18.01 \\ 
    50\%   & 39.19/16.20/31.16 & 41.49/18.60/38.64 & 36.15/15.36/29.56 & 23.92/5.83/18.33 \\ 
    60\%   & 39.06/16.08/31.08 & 41.27/18.40/38.42 & 36.04/15.34/29.47 & 23.14/5.50/17.74 \\ 
    75\%  & 36.94/14.21/29.14 & 40.17/17.52/37.37 & 34.32/13.72/27.96 & 21.72/4.32/16.45 \\  \hline
    \multicolumn{5}{c}{\textbf{ROUGE scores reported in Figure \ref{fig:exp_vocab}}} \\\hline
    & XSum & CNN/DailyMail & WikiHow & Reddit TIFU \\
    & R1/R2/RL & R1/R2/RL &  R1/R2/RL & R1/R2/RL \\\hline 
    BPE 32k         & 39.23/16.17/31.13 & 41.86/18.97/38.97 & 35.22/14.88/28.87 & 24.04/6.04/18.57 \\ 
    Unigram 32k     & 38.94/15.99/30.97 & 41.75/19.08/38.91 & 36.94/15.68/30.28 & 24.17/6.07/18.54 \\ 
    Unigram 64k     & 39.17/16.33/31.24 & 41.89/19.19/39.03 & 37.58/16.02/30.71 & 24.47/6.32/18.90 \\ 
    Unigram 96k     & 39.33/16.40/31.24 & 42.22/19.31/39.34 & 37.38/15.94/30.63 & 24.10/6.22/18.73 \\
    Unigram 128k    & 39.26/16.27/31.14 & 41.76/19.08/38.89 & 37.66/16.04/30.83 & 23.74/5.95/18.33 \\ 
    Unigram 256k    & 38.55/15.92/30.62 & 41.98/19.11/39.08 & 36.94/15.49/30.08 & 23.63/5.95/18.33  \\ \hline
    \multicolumn{5}{c}{\textbf{ROUGE scores reported in Figure \ref{fig:exp_pre-training_steps}}} \\\hline
    & XSum & CNN/DailyMail & WikiHow & Reddit TIFU \\
    & R1/R2/RL & R1/R2/RL &  R1/R2/RL & R1/R2/RL \\\hline 
    No pretraining  & 30.83/10.83/24.41 & 38.27/15.03/35.48 & 32.48/10.53/23.86 & 15.89/1.94/12.22 \\ 
    100k-step       & 37.68/14.89/29.78 & 40.83/18.24/37.99 & 34.01/14.07/28.13 & 23.33/5.52/17.95 \\ 
    200k-step       & 38.72/15.74/30.74 & 41.40/18.53/38.57 & 34.91/14.64/28.70 & 23.48/5.62/18.05 \\ 
    300k-step       & 39.15/16.12/31.05 & 41.63/18.79/38.76 & 35.61/15.09/29.22 & 23.75/5.92/18.35 \\ 
    400k-step       & 39.45/16.34/31.37 & 41.81/18.89/38.95 & 36.14/15.41/29.64 & 23.93/5.92/18.43 \\ 
    500k-step       & 39.79/16.58/31.70 & 41.79/18.81/38.93 & 36.58/15.64/30.01 & 24.36/6.09/18.75 \\ \hline 
  \end{tabular}
  \caption{The raw ROUGE1-F1, ROUGE2-F1 and ROUGEL-F1 scores reported in corresponding figures.}
  \label{tab:raw_rouge_scores}
\end{table*}

\clearpage
\section{Low Resource Numbers}
\label{appendix:low_resource}
\begin{table*}[hbt!]
\scriptsize
\centering
\renewcommand{\arraystretch}{1.3}
\begin{tabular}{ccccccc}
\hline
Dataset & 0 examples & 10 examples & 100 examples & 1k examples & 10k examples & previous SOTA \\
 & $R_1/R_2/R_L$ & $R_1/R_2/R_L$ & $R_1/R_2/R_L$ & $R_1/R_2/R_L$ & $R_1/R_2/R_L$  &  $R_1/R_2/R_L$    \\ \hline
XSum & 19.27/3.00/12.72 & 19.39/3.45/14.02 & 39.07/16.44/31.27 & 41.55/18.23/33.29 & 44.71/21.20/36.31 & 45.14/22.27/37.25 \\
CNN/DailyMail & 32.90/13.28/29.38 & 37.25/15.84/33.49 & 40.28/18.21/37.03 & 41.72/19.35/38.31 & 42.54/20.04/39.32  & 44.16/21.28/40.90 \\
NEWSROOM & 22.06/11.86/17.76 & 29.24/17.78/24.98 & 33.63/21.81/29.64 & 37.26/25.34/33.12 & 39.54/27.25/35.45 & 39.91/28.38/36.87 \\
Multi-News & 36.54/10.52/18.67 & 39.79/12.56/20.06 & 41.04/13.88/21.52 & 44.00/15.45/22.67 & 44.70/16.57/23.43 & 43.47/14.89/17.41 \\
Gigaword & 23.39/7.59/20.20 & 25.32/8.88/22.55 & 29.71/12.44/27.30 & 32.95/13.90/30.10 & 35.13/16.36/32.61 & 38.73/19.71/35.96 \\
WikiHow & 22.59/6.10/14.44 & 23.95/6.54/15.33 & 25.24/7.52/17.79 & 34.35/12.17/25.84 & 37.22/14.41/29.15 & 28.53/9.23/26.54 \\
Reddit TIFU & 14.66/3.06/10.17 & 15.36/2.91/10.76 & 16.64/4.09/12.92 & 23.34/6.85/18.46 & 25.47/8.18/20.33 & 19.0/3.7/15.1 \\
BIGPATENT & 25.61/6.56/17.42 & 28.87/8.30/19.71 & 33.52/10.82/22.87 & 36.85/12.58/24.54 & 34.81/12.39/24.13 & 37.52/10.63/22.79 \\
arXiv & 28.05/6.63/17.72 & 31.38/8.16/17.97 & 33.06/9.66/20.11 & 39.46/12.38/22.20 & 40.24/14.04/23.11 & 41.59/14.26/23.55\\
PubMed & 28.17/7.57/17.85 & 33.31/10.58/20.05 & 34.05/12.75/21.12 & 40.15/15.56/24.05 & 41.75/16.74/24.80 & 40.59/15.59/23.59 \\
AESLC & 10.35/3.86/9.29 & 11.97/4.91/10.84 & 16.05/7.20/15.32 & 28.58/15.45/28.14 & 36.47/20.85/35.53 & 23.67/10.29/23.44 \\
BillSum & 41.02/17.44/25.24 & 40.48/18.49/27.27 & 44.78/26.40/34.40 & 46.47/30.58/37.21 & 50.81/34.49/40.96 & 40.80/23.83/33.73 \\
\hline
\end{tabular}
\caption{The ROUGE1-F1, ROUGE2-F1 and ROUGEL-F1 scores of low resource summarization reported in Figure~\ref{fig:low_resource} along with previous SOTA in Table~\ref{tab:large_model}.
With 100 examples, \pegasuslarge beats previous SOTA on ROUGE2-F1 metrics on BIGPATENT, Reddit TIFU, and BillSum dataset. 
With 1000 examples, \pegasuslarge beats previous SOTA metrics on Multi-News, WikiHow, Reddit TIFU, BigPatent, AESLC and BillSum.
}
\label{tab:low_resource_rouge_scores}
\end{table*}

\clearpage
\section{Human Evaluation Details}
\label{appendix:human}

\begin{figure*}[tbh]
\centering
\includegraphics[width=1.0\textwidth]{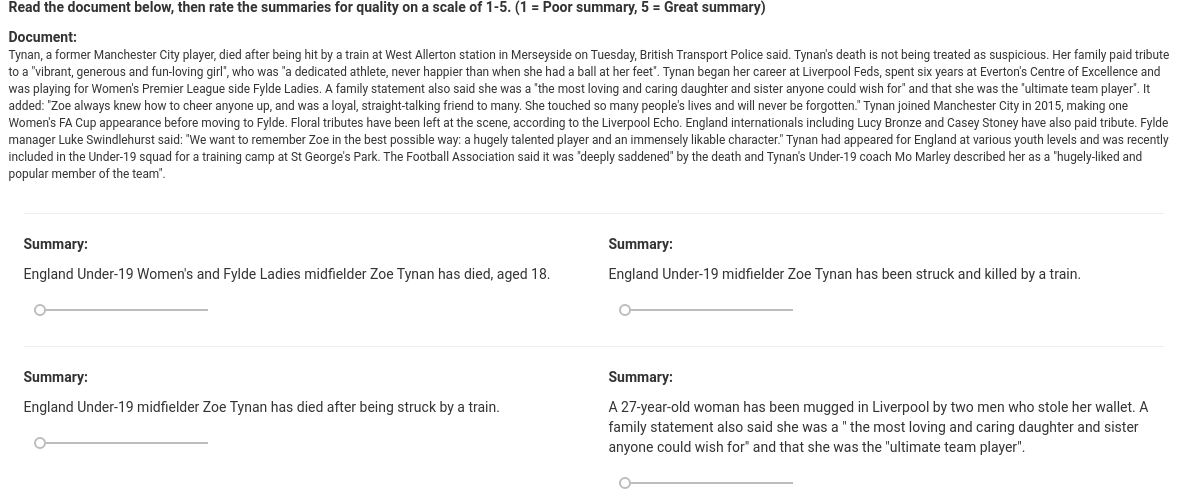}
\caption{A screenshot of the Amazon MTurk HIIT.}
\label{fig:mturk}
\end{figure*}

In all human evaluation experiments we used the same task template
shown in Figure \ref{fig:mturk}, where workers were asked to rate
4 summaries for a document on a scale of 1 (poor summary) to 5 (great summary). The order in which the summaries are presented for each task was random per example. Each task was independently done by 3 different workers and we retained the median score across workers for each summary. We paid 1 USD per task and used the following critieria for
workers to ensure high-quality:
\begin{itemize}
    \item Location: US
    \item Minimum approval rate: 95\%
    \item Minimum HIITs: 1000
\end{itemize}

With this criteria we observed high reproducibility in the conclusions of the huamn evaluation. Multiple runs of the same experiment with different workers meeting this criteria yielded very similar results. The HITT template is provided at \pegasuscode.

In experiment 1, the four summaries corresponded to 3 models (\pegasuslarge pre-trained on \gnews, C4, and \transformerbase) that were fine-tuned using all the supervised examples along with the reference (human) summary. We sampled 100 examples from each dataset (XSum, CNN/DailyMail, Reddit TIFU).

In experiment 2, we evaluated 4 models  (\pegasuslarge pre-trained
on \gnews fine-tuned using
different amounts of supervision, 10, 100, 1000, and all examples)  alongside the human summary. To do this with the same template,
for each example we randomly selected 4 out of the 5 summaries. This
resulted in fewer ratings per model, but did not increase the
work (and cost) of the task.

We used a paired t-test to determine statistical significance
when comparing the ratings of two sets of summaries.

\clearpage
\section{Example of summary with relatively low ROUGE2-F but qualitatively good.}

This figure shows an example
model summary from the CNN/DailyMail dataset exhibiting high fluency,
coherence, although highly abstractive, and only ROUGE2-F of 16. The model understood that 
the football team "Chelsea" could be paraphrased as "Jose Mourinho's side" and
"The Blues" and highlighted the same four matches to be played.

\begin{figure*}[htb!]
\begin{mdframed}
\small
    
    \textbf{Document:} chelsea will face paris saint-germain, the french team who knocked jose mourinho’s side out of the champions league this season, in a pre-season friendly in july. the blues, who were sent crashing out on away goals at the last-16 stage following a 2-2 draw at stamford bridge, will play psg in north carolina on july 25. it is one of three games mourinho’s side will feature in across the pond as they gear up to defend a probable premier league title. john terry leads the celebrations as chelsea close in on the premier league title with a 0-0 draw at arsenal . eden hazard, the pfa player of the year, will line-up for chelsea when they travel to the usa in the summer . new york red bulls - july 22 - new jersey . paris saint-germain - july 25 - charlotte, north carolina . barcelona - july 28 - washington d.c. fiorentina - august 5 - stamford bridge . chelsea, 10 points ahead of arsenal with just four games to play, will also face the new york red bulls on july 22 and spanish giants barcelona six days later in washington. chelsea fans will then get to see their side before the premier league campaign kicks-off with a friendly against fiorentina at stamford bridge on august 5. all four matches mark chelsea’s participation in this summer’s pre-season international champions cup with manchester united, who mourinho’s side will not face, la galaxy, porto and san jose earthquakes also involved. ‘i’m pleased we are able to announce our fixtures for what promises to be an exciting summer,' said chelsea chairman bruce buck. ‘as promised, we face some excellent opposition across several iconic venues in the united states and to top it off we are delighted to be hosting fiorentina at stamford ... ... ...  \\
    
    \textbf{Ground-truth:}
    chelsea to play three matches inside six days in the united states . they will face new york red bulls, paris saint-germain and barcelona . fiorentina will then travel to stamford bridge for friendly on august 5 . four matches will make up chelsea's participation in champions cup . read: chelsea interested in £43m antoine griezmann .
    \\ 
    
    \textbf{Model:}  
    jose mourinho's side will play psg in north carolina on july 25 . chelsea will also face the new york red bulls and barcelona . the blues will play fiorentina at stamford bridge on august 5 .
    
\normalsize
\end{mdframed}

\caption{A CNN/DailyMail \pegasuslarge model summary with relatively low
ROUGE2-F of 16, but qualitatively quite good, and factually accurate.
}

\label{fig:qualitative_cnn_example}

\end{figure*}

\clearpage
\section{Abstractiveness of Summaries}

\begin{figure}[tbh]
\centering
\includegraphics[width=\textwidth]{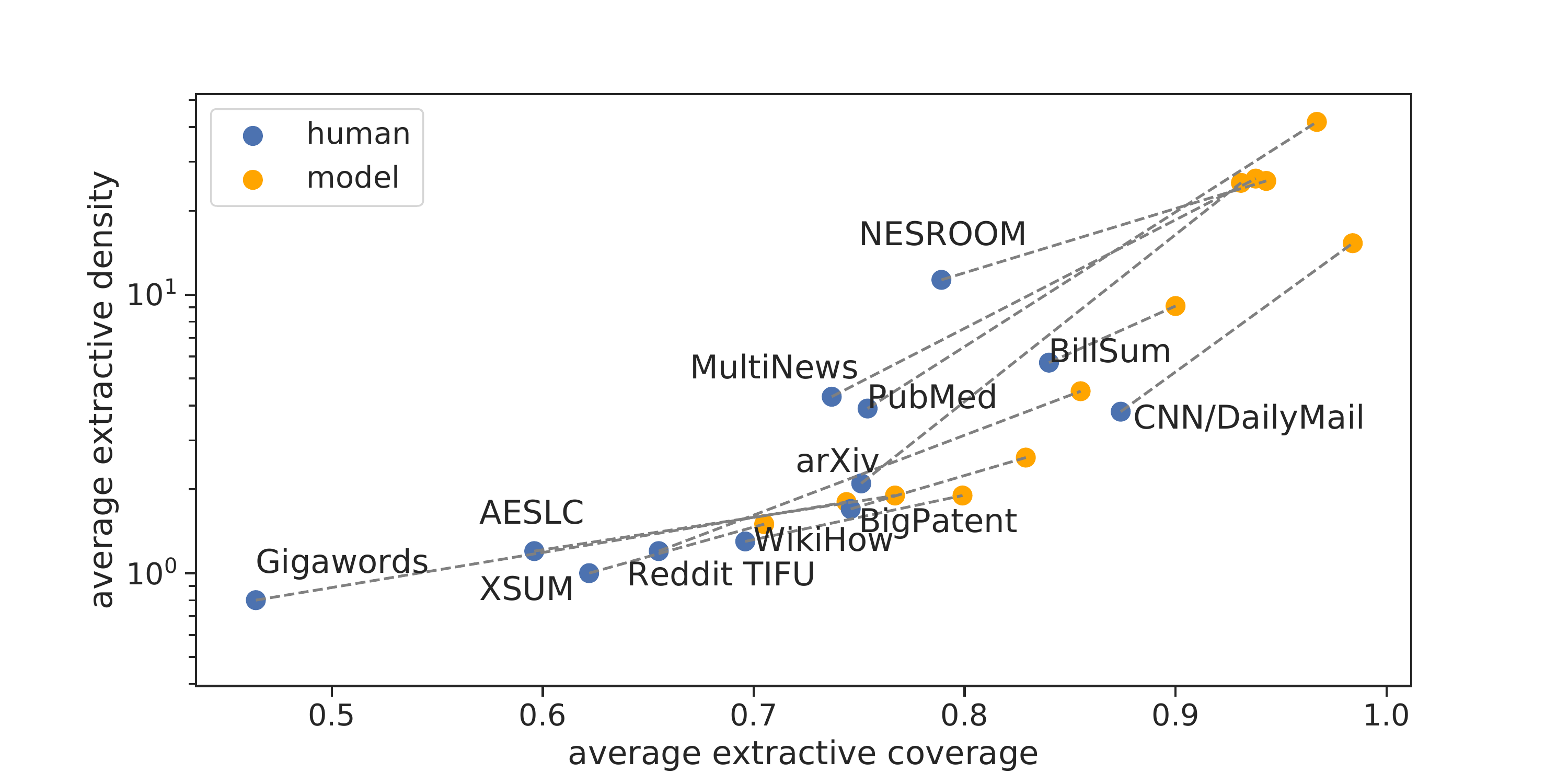}
\caption{Comparison of abstractiveness of human written and model generated summaries.
}
\label{fig:abstractiveness}
\end{figure}

We compared the abstractiveness of model generated summaries with the human-written ones for all downstream datasets. We measured abstractiveness of summaries using average values of extractive coverage and extractive density \citep{Grusky2018newsroom} on each dataset. More abstractive summaries have smaller extractive coverage (more novel words) and smaller extractive density (smaller spans copied from inputs). Figure~\ref{fig:abstractiveness} shows that the summaries generated by models were all less abstractive than the human-written counterparts. However, the models that were finetuned on more abstractive datasets, such as XSum and Reddit TIFU, could generate more abstractive summaries than human-written ones on other datasets.

\section{Example Model Outputs}
\label{appendix:model_ouputs}

Model outputs were selected (and \LaTeX~ tables generated) automatically by a program
in the following way: (1) pick first 300 examples of triplets (document, gold summary, model output) from the dataset test split; (2) rank the examples by ROUGE1-F1/ROUGE2-F1/ROUGEL-F1 metrics in descending order; (3) divide the examples into 2-10 buckets depending on the documents lengths; (4) randomly pick one example from each bucket.

We filtered out examples that contain bad words from the link \footnote{
\url{https://github.com/LDNOOBW/List-of-Dirty-Naughty-Obscene-and-Otherwise-Bad-Words/blob/master/en}}. Input documents were truncated at 300 words for visualization. Each page shows examples from one dataset sampled by one ROUGE metric.

\begin{table*}[bth!]
\scriptsize
\centering
\renewcommand{\arraystretch}{1.3}
\begin{tabular}{c|p{0.85\textwidth}}
\hline
\multicolumn{2}{c}{XSum} \\ \hline
\makecell{Document\\(ID \#187)} & 
GP practices are being paid to help local NHS groups limit the number of patient referrals and cut costs, the doctors' magazine Pulse found. Appointments affected include scans and consultations with specialists - including those for cancer patients. The British Medical Association said such incentives were "misguided". At least nine clinical commissioning groups (CCGs) were offering GP practices payments for hitting targets, according to Pulse's investigation. In one case, Birmingham South Central CCG was offering practices more than   £11,000 to reduce new outpatient attendances, follow-ups, A\&E attendances and emergency admissions by 1\%, compared with 2014/15. It said the schemes, which exclude cancer referrals, were designed to "incentivise best quality practice" and "drive improvements in the quality of primary medical care". "Our priority is to ensure that patients have access to services that they need, when they need them," said a spokesperson for Birmingham South Central CCG. Another CCG told Pulse it had considered the "full impact" of the incentive scheme and was "confident that there is no conflict of interest". Pulse said that one scheme had already been looked at by the General Medical Council, the body which regulates medical standards in the UK, after local GP leaders expressed their concern. The magazine pointed out that initial hospital referrals for cancer patients should happen within two weeks of a GP first suspecting the condition. Dr Chand Nagpaul, chairman of the GPs committee of the doctors' trade union the British Medical Association, told BBC Radio 4's Today programme that such schemes were a "financial contaminant" to patient-doctor trust. He said: "It's short-sighted and misguided of CCGs to introduce such mechanisms, because they do lead to the potential for patients questioning the motives of GP referrals. "We believe it is far more appropriate for CCGs to introduce clinical pathways that ... ... ... 
\\ \hline
Gold & 
Some doctors in England are being offered thousands of pounds to cut the number of patients being sent to hospital, an investigation has found.
\\ \hline
Model & 
Thousands of pounds are being paid to GPs in England to avoid sending patients to hospital, an investigation has found.
\\ \hline
ROUGE1-F1 & 
68.18
\\ \hline
\makecell{Document\\(ID \#206)} & 
The striker took his tally to five goals in two games when hitting the opener and winner in an ABAX Stadium thriller. Marriott fired Posh ahead two minutes into the second half with a low shot that crept in courtesy of a kiss off the inside of the far post. But Rotherham were soon back on level terms as Kieffer Moore, who like Marriott hit a hat-trick in his previous league appearance, headed in captain Lee Frecklington's left-wing cross. Marriott then steered a Marcus Maddison cross against a post to be denied by the woodwork for the second time after seeing an early effort hit the underside of the crossbar. Rotherham thought they had hit the front moments later only to see their celebrations cut short by a raised flag when Jamie Proctor nodded in from close range. But a cracking contest was settled with 15 minutes to go when Marriott got the better of dithering defender Semi Ajayi and lobbed a glorious finish over stranded Rotherham goalkeeper Richard O'Donnell. Match report supplied by the Press Association. Match ends, Peterborough United 2, Rotherham United 1. Second Half ends, Peterborough United 2, Rotherham United 1. Corner, Rotherham United. Conceded by Michael Doughty. Kieffer Moore (Rotherham United) is shown the yellow card for hand ball. Corner, Rotherham United. Conceded by Michael Doughty. Attempt missed. Joe Newell (Rotherham United) right footed shot from a difficult angle on the left is close, but misses to the left. Attempt saved. Andrew Hughes (Peterborough United) left footed shot from outside the box is saved in the top left corner. Joe Mattock (Rotherham United) is shown the yellow card for a bad foul. Idris Kanu (Peterborough United) wins a free kick in the defensive half. Foul by Joe Mattock (Rotherham United). Attempt missed. Jack Marriott (Peterborough United) ... ... ... 
\\ \hline
Gold & 
Jack Marriott hit a second-half double as Peterborough continued their perfect League One start by beating Rotherham.
\\ \hline
Model & 
Jack Marriott continued his goalscoring form in League One with a brace as Peterborough beat Rotherham 2-1.
\\ \hline
ROUGE1-F1 & 
50.00
\\ \hline
\makecell{Document\\(ID \#276)} & 
Thieves took the animal from Silver Star Pets in Pengam, along with dog leads, dog coats, fish tanks, pet food and dog grooming equipment worth £35,000. The incident is one of four break-ins that occurred at the Britannia Enterprise Centre some time between Tuesday evening and Wednesday morning. Police are appealing for information. The other units broken into were Auto Stitch, George Monumental Masons and Pride and Groom where damage was caused to the doors but nothing was stolen. A taxi sign, however, was also stolen from one of the vehicles parked at the centre.
\\ \hline
Gold & 
An African Grey parrot valued at £900 has been stolen from a pet shop in Caerphilly.
\\ \hline
Model & 
A Yorkshire terrier worth £100,000 has been stolen from a pet shop in Stoke-on-Trent.
\\ \hline
ROUGE1-F1 & 
48.48
\\ \hline
\makecell{Document\\(ID \#70)} & 
Matthew Gillard, of Connsbrook Avenue in east Belfast, pleaded not guilty to charges including kidnapping and false imprisonment. The defendant, 25, also denied charges of common assault, assaulting a police officer and driving dangerously. The charges relate to incidents in east Belfast and Comber on Saturday 4 April. Newtownards Magistrates Court heard that the defendant and the woman were in his car in east Belfast when he began questioning her about who she was seeing. A detective giving evidence in court said the woman tried to get out of the Seat Toledo car but the defendant allegedly drove off at speed, through a red light on Bloomfield Avenue. When the car stopped on the Belfast Road in Comber, the woman escaped along a lane but was carried back to the car by the defendant, the court heard. An off-duty police sergeant saw this and went to help the woman. When the sergeant tried to intervene, the defendant allegedly drove his car at the officer, forcing him to get out of the way. The sergeant was able to pull the keys from the ignition through the car's open window, but the defendant wrenched them from his grasp. He then drove to the Grand Parade area of east Belfast where the woman was released. He is also alleged to have sent the woman a message threatening that if she went to police about the incident he would "ruin her life in every possible way". The court heard that the defendant handed himself over to police on Wednesday, despite being aware since Saturday that he was wanted by the PSNI. But during police interviews he refused to answer questions put to him. An application for bail was made but this was refused. Mr Gillard will appear again in court on 1 May.
\\ \hline
Gold & 
A man accused of kidnapping his partner allegedly drove at a police officer who tried to save her, a court has heard.
\\ \hline
Model & 
A man has appeared in court charged with kidnapping and assaulting a woman.
\\ \hline
ROUGE1-F1 & 
34.29
\\ \hline
\makecell{Document\\(ID \#124)} & 
Johnson made his film debut in the 1950s and featured in numerous films, alongside stars such as Frank Sinatra, Laurence Olivier and Charlton Heston. A founder member of the Royal Shakespeare Company, Johnson played several lead roles including Romeo and Mark Anthony in Julius Caesar. He also appeared in several TV dramas such as Lewis and Silent Witness. Johnson died after a short illness at the Royal Marsden Hospital in Chelsea, London. He is survived by his wife Lynne, who he married in 2004, and his four children. Johnson was born in Upminster, Essex, and he left his training at the Royal Academy of Dramatic Art (RADA) to join Sir John Gielgud's company. He joined the Royal Navy during World War Two and then made his film debut in 1959, when he appeared in the MGM film Never So Few, starring Frank Sinatra and Gina Lollobrigida. He also appeared in The Haunting (1963) and Khartoum (1966), opposite Laurence Olivier and Charlton Heston. His family said he was offered and turned down the role of James Bond after playing British spy Bulldog Drummond in Deadlier Than the Male (1967) and its sequel Some Girls Do (1969). His most recent film credits include The Boy in the Striped Pyjamas (2008) and Lara Croft: Tomb Raider (2001).
\\ \hline
Gold & 
British actor Richard Johnson, whose career spanned film, theatre and TV, has died aged 87, his family has said.
\\ \hline
Model & 
Actor Roger Johnson, best known for his roles in Deadlier Than the Male and Some Girls Do, has died at the age of 93.
\\ \hline
ROUGE1-F1 & 
27.91
\\ \hline
\end{tabular}
\caption{Generated summaries by $\text{PEGASUS}_{\text{LARGE}}$ (\gnews) on XSum sampled by ROUGE1-F1.}
\end{table*}

\begin{table*}[bth!]
\scriptsize
\centering
\renewcommand{\arraystretch}{1.3}
\begin{tabular}{c|p{0.85\textwidth}}
\hline
\multicolumn{2}{c}{XSum} \\ \hline
\makecell{Document\\(ID \#255)} & 
The 34-year-old has been absent from the club since a training-ground row after the 5-1 Scottish Premiership loss at Celtic on 10 September. Alongside agent Eddy Jennings, he attended a meeting with club officials last week. Barton, who joined from Burnley in May, has repeatedly said he wants to return and fight for his Rangers place. "Joey Barton has been told he will return to full-time training following the conclusion of a disciplinary procedure," read a statement on Rangers' website. "The Auchenhowie [training base] management team will inform the player of appropriate training arrangements to which he must adhere for as long as he remains a Rangers player. "Neither the club, nor the player, will comment further." Barton is also facing Scottish FA charges for breaching betting rules. He was charged with placing 44 bets on games between 1 July and 15 September. The former Manchester City, Newcastle United and QPR player, who has one England cap, left Burnley under freedom of contract at the end of last season, signing a two-year contract at Ibrox. He has made eight appearances for Mark Warburton's side, who are second in the Premiership. Barton was not available for comment but said earlier on Wednesday on Twitter that it was "a lovely day for a \@Iceman\_Hof session and a run!" in reference to Dutch ice bath training guru Wim Hof. BBC Scotland's Richard Wilson The fact that Joey Barton is returning to full-time training does not necessarily mean the resumption of his Rangers career. Under Fifa rules, a player under contract at a club must be provided with training facilities. The fact that the player was suspended reflects the view of the management team about the serious nature of a training-ground row in September. Any return to the first-team squad would need to be ... ... ... 
\\ \hline
Gold & 
Rangers say midfielder Joey Barton "will return to full-time training" following a club-imposed suspension.
\\ \hline
Model & 
Rangers midfielder Joey Barton is to return to full-time training after being suspended by the club.
\\ \hline
ROUGE2-F1 & 
38.71
\\ \hline
\makecell{Document\\(ID \#7)} & 
Operation Anagram was formed by Strathclyde Police in 2006 after Tobin raped and murdered 23-year-old Polish student Angelika Kluk in Glasgow. It also helped convict him of the murders of 18-year-old Dinah McNicol and 15-year-old Vicky Hamilton. Police said that while the operation was being scaled back it would never end. Det Sup David Swindle, who formed the operation, said anyone with information regarding Peter Tobin would still be able to contact the Anagram incident room via e-mail which would be monitored daily. "Since Operation Anagram was formed in 2006, I could never have believed the momentum and magnitude it would have developed into," he said. "The public response and assistance from the media has been unbelievable. "The long-running investigation has brought some solace to the families of Vicky Hamilton and Dinah McNicol in knowing what happened to their loved ones." Det Sup Swindle said that although Anagram had not uncovered other murder victims of Tobin to date, he was "proud" of its "success in bringing closure" to some bereaved families. He added: "I leave Strathclyde Police after 34 years in the knowledge that we have tried our best to find out the truth of what Tobin has done and also with confidence that the Anagram processes built up over the last four and a half years ensures a life long awareness of this cowardly, vicious serial killer Tobin. "Any new information coming to light will continue to be researched which may bring similar closure to other families. "The search for the truth about what Peter Tobin has done will continue." Tobin is serving three life sentences for the murders of Vicky Hamilton, Dinah McNicol and Angelika Kluk. Ms Kluk was raped and murdered at a church in Glasgow in 2006. The bodies of Ms McNicol, from Essex, and Ms ... ... ... 
\\ \hline
Gold & 
A dedicated police operation set up to examine the activities of serial killer Peter Tobin is to be wound down.
\\ \hline
Model & 
A police investigation into serial killer Peter Tobin is being scaled back.
\\ \hline
ROUGE2-F1 & 
26.67
\\ \hline
\makecell{Document\\(ID \#267)} & 
Norah Boyle, 85, suffered a head injury as Sabrina Duncan and Benter Ouma put her to bed at The Green Nursing Home in Kings Norton. She died 23 days later. The workers delayed calling 999 and pretended Mrs Boyle knocked her head on the head board. They must do 160 hours' unpaid work and pay   £500 court costs each. The pair were not prosecuted for being responsible for the fall, but for being "grossly negligent in their response". Mrs Boyle died in hospital after developing pneumonia. In sentencing, judge Mark Wall said it was impossible to say whether reporting the incident any earlier would have made any difference to Mrs Boyle's eventual death. Her daughter, Ellen Boyle, said: "I'm appalled that that's what they got for what happened to my mum. "I'm appalled that my mother's life is only worth 12 months of a community order." Speaking after the court hearing Det Sgt Victoria Lee said the pair had delayed calling an ambulance while they came up with a cover story. "While [Duncan and Ouma] plotted, Mrs Boyle laid in bed with a serious head injury, her head bleeding onto the pillow," she said. "Most of us have relatives who are frail, disabled or vulnerable     we expect them to be cared for professionally and compassionately." Duncan, 40, of Shartlands Close, Cotteridge, and Ouma, 31, of Summerfield Crescent, Edgbaston, had pleaded guilty at an earlier hearing at Birmingham Crown Court to neglect and ill-treatment.
\\ \hline
Gold & 
Two care workers who admitted neglect after a pensioner fell from a hoist at a Birmingham care home have been sentenced to 12-month community orders.
\\ \hline
Model & 
Two care home workers who admitted covering up the death of a woman who fell in her bed have been given community orders.
\\ \hline
ROUGE2-F1 & 
25.53
\\ \hline
\makecell{Document\\(ID \#234)} & 
She will play Denker, a lady's maid to Dame Maggie Smith's character, the Dowager Countess of Grantham. Johnston, who has also appeared in Waking the Dead and Coronation Street, joins new stars Richard E Grant and Anna Chancellor, both of whom will play guests of the Granthams at Downton. The hit period drama will return to screens this autumn. Series four of the show, which followed the wealthy Grantham family and their servants, achieved an average of 11.9 million viewers in the UK. The very British drama has also been a huge hit in the US, winning both Emmy Awards and Golden Globes. More than 26 million viewers watched series four on Masterpiece on PBS, making it one of the highest rating shows on American television. Previous high profile guest stars include Shirley Maclaine who played Martha Levinson, Lady Grantham's mother, and Oscar-nominated actor Paul Giamatti who appeared in last year's Christmas special as her "maverick, playboy" son. Series five will also feature 24 star Rade Sherbedgia as a Russian refugee who has fled the revolution after World War 1. Earlier this year, executive producer Gareth Neame promised it would have "all the usual highs and lows, romance, drama and comedy".
\\ \hline
Gold & 
The Royle Family actress Sue Johnston is the latest star to join series five of ITV's Downton Abbey in a guest role.
\\ \hline
Model & 
Former Emmerdale actress Sian Johnston is to join the fifth series of Downton Abbey.
\\ \hline
ROUGE2-F1 & 
17.14
\\ \hline
\makecell{Document\\(ID \#70)} & 
Matthew Gillard, of Connsbrook Avenue in east Belfast, pleaded not guilty to charges including kidnapping and false imprisonment. The defendant, 25, also denied charges of common assault, assaulting a police officer and driving dangerously. The charges relate to incidents in east Belfast and Comber on Saturday 4 April. Newtownards Magistrates Court heard that the defendant and the woman were in his car in east Belfast when he began questioning her about who she was seeing. A detective giving evidence in court said the woman tried to get out of the Seat Toledo car but the defendant allegedly drove off at speed, through a red light on Bloomfield Avenue. When the car stopped on the Belfast Road in Comber, the woman escaped along a lane but was carried back to the car by the defendant, the court heard. An off-duty police sergeant saw this and went to help the woman. When the sergeant tried to intervene, the defendant allegedly drove his car at the officer, forcing him to get out of the way. The sergeant was able to pull the keys from the ignition through the car's open window, but the defendant wrenched them from his grasp. He then drove to the Grand Parade area of east Belfast where the woman was released. He is also alleged to have sent the woman a message threatening that if she went to police about the incident he would "ruin her life in every possible way". The court heard that the defendant handed himself over to police on Wednesday, despite being aware since Saturday that he was wanted by the PSNI. But during police interviews he refused to answer questions put to him. An application for bail was made but this was refused. Mr Gillard will appear again in court on 1 May.
\\ \hline
Gold & 
A man accused of kidnapping his partner allegedly drove at a police officer who tried to save her, a court has heard.
\\ \hline
Model & 
A man has appeared in court charged with kidnapping and assaulting a woman.
\\ \hline
ROUGE2-F1 & 
6.06
\\ \hline
\end{tabular}
\caption{Generated summaries by $\text{PEGASUS}_{\text{LARGE}}$ (\gnews) on XSum sampled by ROUGE2-F1.}
\end{table*}

\begin{table*}[bth!]
\scriptsize
\centering
\renewcommand{\arraystretch}{1.3}
\begin{tabular}{c|p{0.85\textwidth}}
\hline
\multicolumn{2}{c}{XSum} \\ \hline
\makecell{Document\\(ID \#198)} & 
Media playback is not supported on this device Craig Cathcart put the visitors ahead before substitute Simon Church won and scored an 89th-minute penalty. "There were lots of positives out of it even if we'd have come off and lost 1-0. They had a good mentality and attitude," said Coleman. Wales face another Euro 2016 warm-up game against Ukraine in Kiev on Monday. "We look forward to our next challenge now," added Coleman. "The team will change up again, and we'll see how they go again." Striker Church, currently on loan at Scottish Premiership side Aberdeen from Reading, was delighted with his equaliser from the spot. "Northern Ireland were a tough side to play against. They've obviously done well to get where they are and it was a tough game," he said. "We wanted to do well because it was the last time a Wales crowd would see us before the Euros and we wanted to put in a good performance. "I've just got to keep going now and hopefully score some goals. This is a great squad to be part of."
\\ \hline
Gold & 
Wales manager Chris Coleman said he was pleased with his team's performance after they came from behind to draw 1-1 with Northern Ireland in Cardiff.
\\ \hline
Model & 
Wales manager Chris Coleman praised his side's attitude after they came from behind to draw 1-1 with Northern Ireland in Cardiff.
\\ \hline
ROUGEL-F1 & 
80.00
\\ \hline
\makecell{Document\\(ID \#25)} & 
The Senegal international, 26, joined for £9m from Lille in July 2015 and played 35 times as Villa were relegated from the Premier League last season. Other interested clubs have until the end of July to make a bid for Gueye. "If he wants to go, we are powerless," club chairman Dr Tony Xia posted on Twitter. Gueye only missed three league games for Villa in 2015-16 and scored his only goal for the club in their FA Cup fourth round win against Wycombe in January. It is believed the fee that has activated the departure clause is about £7m, with Villa bracing themselves for further offers.
\\ \hline
Gold & 
Aston Villa cannot stop midfielder Idrissa Gueye leaving the club after Everton triggered a release clause in the player's contract.
\\ \hline
Model & 
Aston Villa midfielder Idrissa Gueye has triggered a clause in his contract that will allow him to leave the club this summer.
\\ \hline
ROUGEL-F1 & 
46.51
\\ \hline
\makecell{Document\\(ID \#279)} & 
The early work, carried out on mice and pigs, reveals the protein-infused patch encourages the growth of healthy cells and leads to less scarring. Scarring can be common after a heart attack, making the heart pump less effectively and sometimes fail. Writing in the journal Nature, researchers say the patch may one day revolutionise treatment. During an attack, muscle cells in the heart die because of a lack of blood flow and scientists believe repairing or replacing some of these cells may help reduce long-term damage. In this trial an international team of researchers soaked a collagen patch in a protein known as Fstl1 and stitched it on to the hearts of animals who had experienced heart attacks. Though the protein occurs naturally in healthy hearts, it becomes depleted in a key layer of the heart after an attack. Two weeks later the hearts began to grow fresh muscle cells and new blood vessels, while showing signs of pumping more effectively. Prof Pilar Ruiz-Lozano at Stanford University (which has patented the patch), said: "Many were so sick prior to getting the patch that they would have been candidates for heart transplantation. "The hope is that a similar procedure could eventually be used in human heart attack patients who suffer severe heart damage." Commenting on the study in Nature, Prof Gordana Vunjak-Novakovic at Columbia University, said the work "could lead to entirely new modalities for treating heart infarction". But she cautioned that further studies needed to be done to understand whether this type of approach would work on larger animals and ultimately humans.
\\ \hline
Gold & 
A prototype patch could help the repair the damage caused by a heart attack, scientists say.
\\ \hline
Model & 
Scientists say they have developed a synthetic patch that can repair damaged hearts after a heart attack.
\\ \hline
ROUGEL-F1 & 
36.36
\\ \hline
\makecell{Document\\(ID \#177)} & 
Around 155 countries are expected to formally sign the deal at the UN, setting in motion events that could see the treaty operational within a year. The UN says the expected record turnout for the signing shows overwhelming global support for tackling rising temperatures. But some environmentalists have dismissed the event as a "distraction". Despite the absence of President Obama, around 60 world leaders are expected here at UN headquarters, including French President Francois Hollande and Prime Minister Trudeau from Canada. But their signatures alone will not be enough to make the Paris agreement operational. The legal requirements mean that each country will have to go through a process of ratification. For some this will require nothing more than the assent of the political leader as in the example of the United States. Others though, such as India and Japan, will have to take the document to their parliaments; some may need new laws. The European Union is expected to lag behind on this issue as it has not yet agreed with the 28 member states on how emissions cuts will be shared out. Each member state will also have to ratify the deal individually. Some countries, including the Marshall Islands, Palau, Fiji and Switzerland, have already completed this step and will be able to formally join the agreement on April 22. To become operational, the treaty needs at least 55 countries representing at least 55\% of global emissions to complete all the steps. While this is a tough threshold to reach an unusual coalition of interests is making it possible. Firstly President Obama is keen to ensure the deal is operational before his successor takes office next January. If the next President wants to take the US out of an established treaty they will have to wait for four ... ... ... 
\\ \hline
Gold & 
The first significant step to putting the Paris Climate Agreement into practice will take place on Friday.
\\ \hline
Model & 
World leaders are gathering in New York to sign the Paris Agreement on climate change, despite US President Barack Obama not attending.
\\ \hline
ROUGEL-F1 & 
25.64
\\ \hline
\makecell{Document\\(ID \#186)} & 
There could be "serious implications" for Gloucestershire Police, the police and crime commissioner has warned. Cotswold District Council's proposal to form a unity authority with West Oxfordshire has proved controversial. But CDC says the plans - dubbed 'Coxit' - are at an early stage but aim to improve accountability. CDC leader Lynden Stowe has said Gloucestershire's "historic borders" would remain under the proposals, which aim to improve the "administration and the delivery of council services". He told BBC Radio Gloucestershire a few weeks ago: "We would expect the police to patrol up to the existing county borders, exactly as now, so if there's a crime in Cirencester, Gloucestershire Police attend and if there's a burglary in Burford, Thames Valley Police attend." But Gloucestershire's PCC Martin Surl said he had taken legal advice on the "Coxit" proposal, and warned it could "signal the end of Gloucestershire Police as we know it". Under the Local Government and Public Involvement in Health Act 2007, he said the Cotswolds could become part of Thames Valley Police District. "If they break away to a new district, we can't continue policing just half a district," he told the BBC. "So, either we start taking some of Thames Valley's area in Oxfordshire, or they start policing parts of Gloucestershire. Neither want that ." A spokesman for Cotswold District Council said they "fully understand the initial concerns" but they were "best addressed through the detailed studies that will now take place to assess the feasibility of the unitary proposal". Mr Surl, an independent, is standing for re-election as Gloucestershire PCC on 5 May, alongside Labour's Barry Kirby and the Conservatives' Will Windsor Clive.
\\ \hline
Gold & 
The Cotswolds could be policed by the Thames Valley force if plans to "break away" from Gloucestershire County Council go ahead, it is claimed.
\\ \hline
Model & 
Plans to create a new district in the Cotswolds have been criticised by a police chief.
\\ \hline
ROUGEL-F1 & 
15.00
\\ \hline
\end{tabular}
\caption{Generated summaries by $\text{PEGASUS}_{\text{LARGE}}$ (\gnews) on XSum sampled by ROUGEL-F1.}
\end{table*}

\begin{table*}[bth!]
\scriptsize
\centering
\renewcommand{\arraystretch}{1.3}
\begin{tabular}{c|p{0.85\textwidth}}
\hline
\multicolumn{2}{c}{CNN/DailyMail} \\ \hline
\makecell{Document\\(ID \#134)} & 
a us citizen has been killed in a mortar attack in yemen after he traveled to the country in an attempt to extricate his pregnant wife and daughter from the civil war there and fly them to california, family say. jamal al-labani was an oakland gas station owner, his cousin mohammed alazzani told kpix-tv. according to alazzani, al-labani was in yemen visiting his pregnant wife and the couple's two-and-a-half-year-old daughter. alazzani told kpix al-labani was trying to get his family out of the war-torn middle eastern nation and take them to oakland - but he couldn't because the us has withdrawn its diplomatic staff and the country has shut down most airports. rebels from the houthi islamist group have been battling to take aden, a last foothold of fighters loyal to saudi-backed president abd-rabbu mansour hadi. they have advanced to the city center despite 11 days of air strikes by a saudi-led coalition of mainly gulf air forces. scroll down for video . family: jamal al-labani was in yemen visiting his pregnant wife and the couple's 2 1/2-year-old daughter . attack: jamal al-labani's family has said he was struck by mortar shrapnel after leaving a mosque tuesday and soon died . sunni muslim saudi arabia launched the air strikes on march 26 in an attempt to turn back the iran-allied shi'ite houthis, who already control yemen's capital sanaa, and restore some of hadi's crumbling authority. the air and sea campaign has targeted houthi convoys, missiles and weapons stores and cut off any possible outside reinforcements - although the houthis deny saudi accusations that they are armed by tehran. career: al-labani reportedly wished to take his family to oakland, though there was no way to do so given what has been happening in yemen . the fighting has failed so far ... ... ... 
\\ \hline
Gold & 
jamal al-labani was a oakland, california, gas station owner, as well as a husband and a father-of-three . al-labani traveled to yemen in an attempt to extricate his pregnant wife and daughter from the civil war there and fly them to california . he was unable to because the us withdrew its diplomatic staff in february . yemen also recently shut down most of its airports . al-labani was struck by mortar shrapnel after leaving a mosque tuesday in aden and soon died . al-labani's cousin has said houthi forces launched the mortar shelling 
\\ \hline
Model & 
jamal al-labani was in yemen visiting his pregnant wife and the couple's two-and-a-half-year-old daughter . al-labani's family said he was struck by mortar shrapnel after leaving a mosque tuesday and soon died . he was with his teenage nephew, who was also killed, when the attack occurred . al-labani's sons from an earlier marriage reside in california .
\\ \hline
ROUGE1-F1 & 
50.93
\\ \hline
\makecell{Document\\(ID \#256)} & 
kevin de bruyne's agent expects to go 'around the world' discussing his client as interest in the wolfsburg midfielder increases ahead of the summer transfer window. patrick de koster, who has confirmed that he held talks with manchester city earlier this season, admits that he could receive 20 phone calls a day about the belgium international as clubs prepare to strengthen their squads before the start of next season. city are keen to sign the ex-chelsea midfielder as they look to reshape their squad but de koster insists no decision has been made and that the 23-year-old could even remain at the volkswagen arena. wolfsburg midfielder has attracted interest from manchester city, bayern munich and paris saint-germain . de bruyne tussles with schalke's sead kolasinac during wolfsburg's 1-1 draw at the volkswagen arena . speaking to the guardian, de koster said: 'for the moment, there are no formal discussions. of course in the next few weeks i will be going around the world to talk about the situation with kevin but this is just informal information. 'i will talk to everybody but kevin is very, very happy with wolfsburg and the way they have treated him since he arrived from chelsea last year. 'there are still five games to play and hopefully they can make sure of a place in the champions league next season so it is a little bit too early to be making any decisions. he has a four-year contract at wolfsburg so we will have to see what they want to do. 'i have met the people from manchester city and we know each other. i've never spoken to anyone from manchester united but a lot of other clubs have been in touch to find out some general information.' bundesliga champions bayern munich and ligue 1 ... ... ... 
\\ \hline
Gold & 
patrick de koster will go 'around the world' to talk about kevin de bruyne . the wolfsburg midfielder is wanted by manchester city and bayern munich . de koster has admitted having talks with city chiefs this season . but he has not spoken to manchester united about a move for his client . de bruyne remains happy at wolfsburg and could yet remain at the club .
\\ \hline
Model & 
kevin de bruyne has attracted interest from manchester city, bayern munich and psg . patrick de koster expects to go 'around the world' discussing his client . de bruyne has scored 10 league goals and provided 17 assists this season .
\\ \hline
ROUGE1-F1 & 
49.50
\\ \hline
\makecell{Document\\(ID \#141)} & 
matthew kenney smoked flakka and then ran naked . a florida man who was high on a designer drug called flakka stripped and ran naked through traffic in fort lauderdale to escape from imaginary killers who he believed stole his clothes and wanted to murder him. matthew kenney, 34, told police he smoked flakka before he streaked though traffic early on saturday evening while only wearing a pair of sneakers. flakka, which can be injected, snorted, smoked, swallowed or taken with other substances, has been nicknamed '\$5 insanity' for its mind-bending effects and cheap cost. after he was arrested, kenney told police he would 'rather die than be caught by these unknown people', the sun sentinel reported. he added that 'if i got hit by a car they would stop chasing me' according to a fort lauderdale police reported. kenney has previous arrests for disorderly conduct, making a riot and possession of a controlled substance. he was hospitalized for a psychiatric evaluation. flakka is usually made from the chemical alpha-pvp, a synthetic version of the stimulant cathinone. that is the same type of chemical that is used to make bath salts. scroll down for video . kenney, 34, ran though traffic early on saturday evening while only wearing sneakers in fort lauderdale, florida . the suspect said he was escaping imaginary killers who he believed stole his clothes and wanted to murder him . the use of flakka a designer drug that can be even stronger than crystal meth or bath salts, is up in florida . flakka resembles a mix of crack cocaine and meth and it has a a strong odor 'like a sweaty sock', wpbf 25 news reported. once ingested, the drug causes a feeling of euphoria, hallucinations and sometimes psychosis or even superhuman strength. the high ... ... ... 
\\ \hline
Gold & 
matthew kenney, 34, said he smoked flakka before he went streaking . was arrested on saturday after run through fort lauderdale, florida . drug is made from same version of stimulant used to produce bath salts . it causes euphoria, hallucinations, psychosis and superhuman strength . kenney has prior arrests and was hospitalized for a psychiatric evaluation .
\\ \hline
Model & 
matthew kenney, 34, told police he smoked flakka before he streaked through traffic in fort lauderdale while only wearing a pair of sneakers . he said he was escaping imaginary killers who he believed stole his clothes and wanted to murder him . kenney has previous arrests for disorderly conduct, making a riot and possession of a controlled substance .
\\ \hline
ROUGE1-F1 & 
40.00
\\ \hline
\makecell{Document\\(ID \#197)} & 
i yield to no one in my love of the old days — warm beer, cricket on the village green, bobbies on bicycles two by two, all that — but it's rare a chance arises to compare the rose-tinted past with the brave new world, as it did on saturday evening when sky's high-octane premier league coverage went head-to-head with arsenal v reading in the fa cup semi-final on the bbc. as we know, the premier league has the money and prestige, but what the fa cup has is history, and boy does the bbc love a bit of history? lest you were in any doubt, its coverage of the semi-final kicked off with footage of the late sir laurence olivier doing the st crispin's day speech from the film of henry v ('we happy few, we band of brothers,' and so on). gary lineker, alan shearer, jason roberts and ian wright fronted the bbc's coverage at wembley . bbc presenter lineker prepares to present the match of the day 50th anniversary special broadcast . reading defender nathaniel chaloboah (left) chases arsenal midfielder aaron ramsey (right) on saturday . gunners forward alexis sanchez celebrates after scoring his side's winning goal in the fa cup semi-final . stand-in match of the day presenter gabby logan (left) with pundits phil neville and robbie savage (right) the excuse, i guess, was that reading's nickname is the royals, but as the second-tier team are also known as the biscuitmen. the bbc even gave us a moment of history at half-time with a breakdown, something that used to be a regular feature of tv outside broadcasts, when a feature on crystal palace's shock semi-final victory over liverpool 25 years ago came to a shuddering halt halfway through. in such instances, the bbc is lucky to ... ... ... 
\\ \hline
Gold & 
the weekend saw bbc's fa cup coverage compete with sky's premier league . it was a refreshing throwback to see the bbc's use of archive footage . gary lineker remains one of the bbc's prized assets and they must keep him .
\\ \hline
Model & 
arsenal beat reading 1-0 in the fa cup semi-final at wembley on saturday . the bbc presented the match of the day 50th anniversary special . the price of live premier league football is now way beyond the reach of the bbc, which may be the clincher in gary lineker's future .
\\ \hline
ROUGE1-F1 & 
29.47
\\ \hline
\end{tabular}
\caption{Generated summaries by $\text{PEGASUS}_{\text{LARGE}}$ (\gnews) on CNN/DailyMail sampled by ROUGE1-F1.}
\end{table*}
\begin{table*}[bth!]
\scriptsize
\centering
\renewcommand{\arraystretch}{1.3}
\begin{tabular}{c|p{0.85\textwidth}}
\hline
\multicolumn{2}{c}{CNN/DailyMail} \\ \hline
\makecell{Document\\(ID \#298)} & 
(cnn)the tulsa county reserve deputy who fatally shot a man instead of using his taser turned himself in to authorities tuesday at the tulsa county jail. video shows reserve deputy robert bates announcing he is going to deploy his taser after an undercover weapons sting on april 2 but then shooting eric courtney harris in the back with a handgun. bates was charged with second-degree manslaughter monday. he surrendered tuesday morning, accompanied by his attorney, clark brewster, and immediately posted bail of \$25,000. as he exited the jailhouse, bates paused in front of television cameras for a moment but did not speak. his attorney reiterated that he believes the charge against his client is unwarranted. the tulsa county sheriff's office says a sting operation caught harris illegally selling a gun. harris ran when officers came in for the arrest. authorities say bates thought he pulled out his taser but "inadvertently" fired his gun. harris' brother, andre harris, told cnn that he is pleased district attorney steve kunzweiler pressed charges. in his opinion, however, no type of force should have been used in the arrest of his brother. watching the video of the shooting, andre harris said he can see that three or more officers were already on top of his brother. that manpower should have been enough to arrest him, he said. "it was a situation where i didn't necessarily think that a taser should even be used," andre harris said. scott wood, another bates' attorney, has said the shooting was an "excusable homicide." investigators' efforts to defend bates and the other deputies involved in the arrest have sparked a mounting chorus of criticism online. harris' relatives are demanding an independent investigation of what they call unjustified brutality. they're also questioning why the 73-year-old bates -- the ceo of an ... ... ... 
\\ \hline
Gold & 
reserve deputy robert bates surrenders to authorities, posts bail of \$25,000 . bates is charged with second-degree manslaughter in the killing of eric harris .
\\ \hline
Model & 
tulsa county reserve deputy robert bates turns himself in to authorities . bates is charged with second-degree manslaughter in the death of eric courtney harris .
\\ \hline
ROUGE2-F1 & 
54.17
\\ \hline
\makecell{Document\\(ID \#148)} & 
a former lager lout who ballooned to 24 stone has lost nearly half his body weight by giving up his favourite drink. rugby prop dale forrest, 26, of bolton, would sink up to 12 pints a night, but decided to shed the pounds after seeing a photo of himself while out with his fitness fanatic friends. in december 2013, the bank teller decided to ditch the booze, give up his favourite fatty readymeals and greasy takeaways and hit the gym. dale forrest would drink up to 12 pints a night - and ballooned to 24 stone - before giving up beer to lose weight . dale, pictured at his slimmest, was worried he would look like a 'beached whale' next to his friends on holiday . since then, mr forrest, who had a holiday planned with his mates and didn’t want to look like a ‘beached whale’ next to them, has lost 10.5 stone. he said: ‘i can now go on nights out and feel confident - even without the dutch courage. ‘i no longer feel like people are laughing at me and all it took was a bit of willpower.’ mr forrest struggled with his weight from a young age and in adulthood fell into unhealthy habits. before shedding the pounds mr forrest ate a cheese and sausage bap for breakfast, a big daddy box meal from kfc for lunch and a meat feast pizza for dinner. dale, pictured sitting down, said he would consume beer, alcopops and shots regularly on nights out . dale, pictured before his weight loss (right) and after losing 10 stone, gave up greasy takeaways for healthy foods and started going to the gym regularly, and saw the pounds fall off . despite being a rugby player, dale weighed 24 stone due to his ... ... ... 
\\ \hline
Gold & 
dale forrest would go out drinking regularly and would eat fatty food . dined on cheese and sausage bap for breakfast and kfc for lunch . decided to lose weight after seeing photos of him next to slim friends . started going to the gym and eating healthy foods and lost 10 stone .
\\ \hline
Model & 
dale forrest would drink up to 12 pints a night on nights out with friends . 26-year-old from bolton worried he would look like a 'beached whale' on holiday . in december 2013 he decided to lose weight after seeing a photo of himself . ditched the booze and started going to the gym and lost more than 10 stone .
\\ \hline
ROUGE2-F1 & 
26.17
\\ \hline
\makecell{Document\\(ID \#260)} & 
chelsea will face paris saint-germain, the french team who knocked jose mourinho’s side out of the champions league this season, in a pre-season friendly in july. the blues, who were sent crashing out on away goals at the last-16 stage following a 2-2 draw at stamford bridge, will play psg in north carolina on july 25. it is one of three games mourinho’s side will feature in across the pond as they gear up to defend a probable premier league title. john terry leads the celebrations as chelsea close in on the premier league title with a 0-0 draw at arsenal . eden hazard, the pfa player of the year, will line-up for chelsea when they travel to the usa in the summer . new york red bulls - july 22 - new jersey . paris saint-germain - july 25 - charlotte, north carolina . barcelona - july 28 - washington d.c. fiorentina - august 5 - stamford bridge . chelsea, 10 points ahead of arsenal with just four games to play, will also face the new york red bulls on july 22 and spanish giants barcelona six days later in washington. chelsea fans will then get to see their side before the premier league campaign kicks-off with a friendly against fiorentina at stamford bridge on august 5. all four matches mark chelsea’s participation in this summer’s pre-season international champions cup with manchester united, who mourinho’s side will not face, la galaxy, porto and san jose earthquakes also involved. ‘i’m pleased we are able to announce our fixtures for what promises to be an exciting summer,' said chelsea chairman bruce buck. ‘as promised, we face some excellent opposition across several iconic venues in the united states and to top it off we are delighted to be hosting fiorentina at stamford ... ... ... 
\\ \hline
Gold & 
chelsea to play three matches inside six days in the united states . they will face new york red bulls, paris saint-germain and barcelona . fiorentina will then travel to stamford bridge for friendly on august 5 . four matches will make up chelsea's participation in champions cup . read: chelsea interested in £43m antoine griezmann .
\\ \hline
Model & 
jose mourinho's side will play psg in north carolina on july 25 . chelsea will also face the new york red bulls and barcelona . the blues will play fiorentina at stamford bridge on august 5 .
\\ \hline
ROUGE2-F1 & 
16.09
\\ \hline
\makecell{Document\\(ID \#73)} & 
she's a best-selling singer, actress, beauty buff and one of the world's most stylish stars. and now, rita ora - who is only 24 and has already made the transition from a kosovan-born fledgling singer to one of the globe's most successful stars - has channeled her passion for fashion into a new adidas range. for her latest collaboration with adidas originals, rita has taken the brand's classics and put her own bold spin on them. the result? a collection that is as daring and vibrant as the designer herself. scroll down for video . rita ora has channeled her passion for fashion into a new adidas range, so femail caught up with the global star to find out her influences and plans for the future . speaking to femail about the inspiration behind her new range, which is emblazoned with graphic dragon prints and lands on may 1, rita said: 'the dragon print is all about my love of travel and soaking up different cultures. the print itself is inspired by asian culture and one of its most recognisable symbols. 'i’ve contrasted that with the fit, basketball style cuts, which were inspired by american culture. my white smoke pack is about the body-mind-soul connection, about taking a moment for reflection. the graphic comes from the visual smoke creates when burning incense.' the talented star, who also recently unveiled a beauty range for rimmel, couldn't be happier with the final result. she said: 'i love being able to go into design meetings with all these crazy ideas and then be able to create a product that hasn’t existed before. from the initial stages to seeing my fans wearing the clothes and shoes - it’s been an amazing experience.' the talented star, who also recently unveiled a beauty range for rimmel, ... ... ... 
\\ \hline
Gold & 
rita, 24, has designed range for adidas originals . designs are inspired by asian culture, she tells femail . star says she's excited to see what the future holds for her .
\\ \hline
Model & 
rita, 24, has teamed up with adidas originals . has taken brand's classics and put her own bold spin on them . promises there's 'a little something for everyone' in the range . cites marilyn monroe as her ultimate postergirl .
\\ \hline
ROUGE2-F1 & 
8.96
\\ \hline
\end{tabular}
\caption{Generated summaries by $\text{PEGASUS}_{\text{LARGE}}$ (\gnews) on CNN/DailyMail sampled by ROUGE2-F1.}
\end{table*}

\begin{table*}[bth!]
\scriptsize
\centering
\renewcommand{\arraystretch}{1.3}
\begin{tabular}{c|p{0.85\textwidth}}
\hline
\multicolumn{2}{c}{CNN/DailyMail} \\ \hline
\makecell{Document\\(ID \#291)} & 
it's truly squeaky bum time in the premier league relegation battle as just nine points separates the bottom seven teams. sportsmail asks some of the managers in and among the dog fight what they feel is required for them to avoid the drop this season. question: 'what will it take for your club to stay in the premier league?' chris ramsey (qpr) 'if we win three games i think we will stay up. it might take less. i'm not saying that's the definitive amount - but we really need to start winning, starting with this weekend. to do that, we need to concentrate right to the end and make sure that our performances stay similar with a bit more defensive resilience.' qpr manager chris ramsey (centre) feels they will avoid relegation if they win three more league games . tim sherwood (aston villa) 'i don't know how many points it will take. we've done alright. nothing's been achieved yet. i'm a new voice and given them a lot of belief and confidence that they are better than what they were showing. they have managed to score a few more goals and have real belief they can go to places and win. something was probably a little bit missing previously. i want to make sure that i don't take my foot off pedal. i'll make sure the players don't.' tim sherwood (left) believes his appointment at aston villa has given them a lot of belief and confidence . nigel pearson (leicester) 'we've put ourselves in a position now where we have a more realistic chance. that's where we are. if it raises optimism elsewhere, fine. if it applies a bit more pressure on other sides around us, fine also. but as far as i'm concerned it's about making sure we're back ... ... ... 
\\ \hline
Gold & 
just nine points separates the bottom seven clubs in the premier league . qpr boss chris ramsey says they need three more wins to survive . burnley host relegation rivals leicester in the league on saturday .
\\ \hline
Model & 
just nine points separate the bottom seven teams in the premier league . qpr manager chris ramsey feels they will avoid relegation if they win three more games . tim sherwood believes his appointment at aston villa has given them a lot of belief and confidence .
\\ \hline
ROUGEL-F1 & 
41.03
\\ \hline
\makecell{Document\\(ID \#139)} & 
a father whose 20-year-old daughter was found murdered in iowa last year made a desperate plea from china for u.s. authorities to do more in tracking down her killer, six months after the girl's body was discovered. and police in iowa seem to have now answered it, by issuing a warrant for the girl's boyfriend, according to reports. tong shao, a chemical engineering student at iowa state university, went missing in september 2014. after a three week search, police found her body stuffed in the trunk of her toyota camry in iowa city. shao's boyfriend, xiangnan li, 23, was listed as a person of interest in the case and is believed to have been the last person to see her alive, however he bought a one-way ticket to china in the days after his girlfriend went missing and has disappeared, cnn reported. wanted for murder: an arrest warrant has reportedly been issued for , xiangnan li, 23 (right), the boyfriend of tong shaom 20 (left), a university of iowa student found murdered in september after going missing . li had transferred to iowa from rochester institute of technology to be closer to tong. the two had meet studying english in beijing in 2011. they had checked into a hotel room together on september 5, 2014. the pair had stayed at the same hotel three times before and the owner knew them. according to police records obtained by cnn, two days earlier tong had accidentally called li - or 'pocket dialed' him - and he stayed on the line for 30 minutes, overhearing a conversation. tong was complaining about li to a friend and said things that 'were not nice', the records noted. the owner of the hotel told investigators li left the hotel either on the night of september 6 ... ... ... 
\\ \hline
Gold & 
tong shao, 20, was an international student from china attending iowa state university . her body was found in the trunk of her car in iowa city on september 26 . police believe it had been for three weeks . she died of blunt force trauma and asphyxiation . her boyfriend, xiangnan li, 23, was the last to see her, but flew to china on september 8, before shao was officially missing . according to tong's father, an arrest warrant has now been issued . however li has disappeared .
\\ \hline
Model & 
tong shao, 20, was found murdered in the trunk of her car in september . her boyfriend, xiangnan li, 23, was a person of interest in the case . li bought a one-way ticket to china in the days after tong went missing . he is believed to have been the last person to see her alive .
\\ \hline
ROUGEL-F1 & 
35.97
\\ \hline
\makecell{Document\\(ID \#293)} & 
a skier in switzerland proved that his dog is certainly his best friend when he brought it along to a snowy slope for a day of skiing. videoed descending the crisp ski runs of the small resort of minschuns in val müstair, adrian schaffner is initially featured ascending the mountain on a button lift. perched on his shoulders looking entirely at ease is his dog sintha – an appenzeller mix according to the owner, who noted alongside the upload of the original video that he only knows the breed of the dog's mother. the dog named sintha appears to be entirely at ease as it sits across its owners shoulders . once at the top, mr schaffner points his skis down the mountain and takes off at speed with the dog remaining calmly sat on his back. the dog appears to be enjoying the sensation of speed as it points its face into the wind and the camera angle changes to show the skier’s descent. after a long ski to the bottom, mr schaffner comes to a stop and the dog jumps from his shoulders and onto the ground. mr schaffner points his skis down the mountain and begins skiing at speed and the dog remains calmly sat on his back . the dog appears to be enjoying the sensation of speed and points its face into the wind . the skier smiles at the camera as the excited dog begins barking and running off in the snow. the video concludes with the dog – who obviously loves snow – chasing after some more skiers as they make their way down another section of mountain. discussing the video, mr schaffner wrote: ‘she grew up on a farm in the mountains pretty wild and more or less without any supervision. mr ... ... ... 
\\ \hline
Gold & 
adrian schaffner skis at speed with pet dog on his shoulders . dog called sintha appears content and leans into the wind . video concludes with dog jumping off and running in snow . footage was captured in ski resort in val müstair, switzerland .
\\ \hline
Model & 
adrian schaffner took his dog sintha on a day of skiing in minschuns, switzerland . the dog appears to be at ease as it sits across its owner's shoulders . after a long ski to the bottom, mr schaffner comes to a stop and the dog jumps from his shoulders and onto the ground . the skier smiles at the camera as the excited dog begins barking and running off in the snow .
\\ \hline
ROUGEL-F1 & 
23.01
\\ \hline
\makecell{Document\\(ID \#55)} & 
danny willett gave a rules official, who had been in his line-of-sight, a verbal blast which was clearly audible to spectators surrounding the 17th green at the masters on thursday. englishman willett vented his anger after his second shot from beyond the green trickled all the way across the putting surface and left the preacher's son facing a bogey or worse. 'of anyone you should know the rules,' willett shouted loudly at the official, who was sitting about 60 yards away in a golf cart, outside the gallery ropes. danny willett waits to play a shot on the fifth during the first round at 2015 masters on thursday . the 27-year-old englishman carded a one-under 71 during his first ever round at augusta . willett was still visibly angry as he left the green after salvaging a bogey, though he had calmed down by time he spoke to reporters about 30 minutes later. he said his ire had been raised because the official had been in his line-of-sight as he was preparing to play his shot. 'we were being timed (for slow play), which i can appreciate,' willett said after carding a one-under 71 at augusta national. 'it's a little bit tricky out here, so it takes a little bit of time, but you'd like to think the referee that's timing you knows exactly where to put his buggy and where not to put his buggy.' willett, 27, a two-time european tour winner, is playing in his first masters. willett was left frustrated with a rules official for getting in his line-of-sight during his round .
\\ \hline
Gold & 
englishman danny willett blasts timing referee for getting in line-of-sight . vented anger at official as he bogeyed the 17th at 2015 masters . willett carded a one-under 71 on opening round in first time at augusta .
\\ \hline
Model & 
danny willett gave a rules official a verbal blast during the first round . the englishman vented his anger after his second shot from beyond the green trickled all the way across the putting surface and left the preacher's son facing a bogey or worse . willett was still visibly angry as he left the green after salvaging a bogey, though he had calmed down by time he spoke to reporters about 30 minutes later . he said his ire had been raised because the official had been in his line-of-sight as he was preparing to play his shot .
\\ \hline
ROUGEL-F1 & 
13.14
\\ \hline
\end{tabular}
\caption{Generated summaries by $\text{PEGASUS}_{\text{LARGE}}$ (\gnews) on CNN/DailyMail sampled by ROUGEL-F1.}
\end{table*}

\begin{table*}[bth!]
\scriptsize
\centering
\renewcommand{\arraystretch}{1.3}
\begin{tabular}{c|p{0.85\textwidth}}
\hline
\multicolumn{2}{c}{NEWSROOM} \\ \hline
\makecell{Document\\(ID \#91)} & 
Penelope Cruz had just scored her first lead in a summer blockbuster, as a sexy, feisty swashbuckler opposite her old pal Johnny Depp. By Dan MacMedan, USA TODAY Penelope Cruz stars in the latest 'Pirates of the Caribbean 4' movie. By Dan MacMedan, USA TODAY Penelope Cruz stars in the latest 'Pirates of the Caribbean 4' movie. It was a physically arduous six-month shoot for a massive tentpole film, replete with swamp-wading and swordfighting. And right before filming began on Pirates of the Caribbean: On Stranger Tides, Cruz discovered buried treasure of a different sort. "I was pregnant through the whole movie. I found out at the beginning," says Cruz, who, with her typical directness, immediately broke the news to Depp and director Rob Marshall. "I found out before we started, and I wanted them to know because I did not want to go in with any secrets, for protection and for the honesty of my relationship with them. For six months they were all taking such good care of me." It helped that Cruz, 37, handled her pregnancy with pirate-worthy panache, having little morning sickness, nausea or dizziness. She avoided any dangerous stunts in the film, which opens today, and relied heavily on her dance background to learn the intricate fight choreography required of her mercenary buccaneer Angelica, who's vying with Depp's Jack Sparrow to find the Fountain of Youth. For Cruz, it felt "good to be working" while expecting, she says. "I had a lot of free days, and once in a while, I had a free week. (The shoot) was very balanced, very easy. I traveled around the world. The whole summer I spent in Hawaii. It was good. I have only good memories" of the shoot. Fast-forward to a May morning in Manhattan. Cruz's infant son ... ... ... 
\\ \hline
Gold & 
And she has scored her first lead in a summer blockbuster, as a sexy, feisty swashbuckler opposite her old pal Johnny Depp.
\\ \hline
Model & 
Penelope Cruz had just scored her first lead in a summer blockbuster, as a sexy, feisty swashbuckler opposite her old pal Johnny Depp.
\\ \hline
ROUGE1-F1 & 
84.44
\\ \hline
\makecell{Document\\(ID \#228)} & 
Seven species of bees in Hawaii have been classified as endangered, the first time the insect has been protected by federal law. The U.S. Fish \& Wildlife Service has granted seven species of yellow-faced bees native to the islands protection under the Endangered Species Act, which will hopefully “allow authorities to implement recovery programs, access funding and limit their harm from outside sources,” Gregory Koob of USFW told The Associated Press. The yellow-faced bee’s population faces a range of threats, like habitat destruction, invasive species both animal and vegetable, and all manner of natural phenomena such as wildfires, hurricanes, tsunamis and droughts. This news represents an important step for bees in general. They — and many other pollinating species like butterflies — are in danger across the world: About 40 percent of invertebrate pollinating species are facing extinction, according to a U.N. report released in February. This could have devastating effects on global agriculture, as about 75 percent of the world’s crops rely on pollination to grow. The ruling also grants protection to 39 plant species found in Hawaii and three other native animals: the band-romped storm-petrel, the orangeblack Hawaiian damselfly and the anchialine pool shrimp. Bees in certain parts of the U.S. are facing a different threat, as aerial spraying, aimed to combat Zika-spreading mosquitos, is killing hives en masse.
\\ \hline
Gold & 
A species of bee in Hawaii has been added to the endangered species list, a first for the insect.
\\ \hline
Model & 
Seven species of bees in Hawaii have been classified as endangered, the first time the insect has been protected by federal law. The U.S. Fish \& Wildlife Service has granted seven species of yellow-faced bees native to the islands protection under the Endangered Species Act, which will hopefully “allow authorities to implement recovery programs, access funding and limit their harm from outside sources,” Gregory Koob of USFW told The Associated Press.
\\ \hline
ROUGE1-F1 & 
28.57
\\ \hline
\makecell{Document\\(ID \#227)} & 
Culture Connoisseurs consistently offer thought-provoking, timely comments on the arts, lifestyle and entertainment. More about badges | Request a badge Washingtologists consistently post thought-provoking, timely comments on events, communities, and trends in the Washington area. More about badges | Request a badge This commenter is a Washington Post editor, reporter or producer. This commenter is a Washington Post contributor. Post contributors aren’t staff, but may write articles or columns. In some cases, contributors are sources or experts quoted in a story. More about badges | Request a badge Washington Post reporters or editors recommend this comment or reader post. You must be logged in to report a comment. You must be logged in to recommend a comment.
\\ \hline
Gold & 
PHOTOS: The legendary heavyweight boxer was one of the most important political, social and athletic figures of the 20th century.
\\ \hline
Model & 
A look at some of our favorite images of the week.
\\ \hline
ROUGE1-F1 & 
19.35
\\ \hline
\makecell{Document\\(ID \#98)} & 
You may have heard Skip Bayless is a Dallas Cowboys fan. And unlike in past years, that is worth celebrating (which Bayless does quite frequently on Undisputed). At 12-2 the Cowboys are running away from the rest of the NFC, have clinched a playoff berth and have a rookie duo in Dak Prescott and Ezekiel Elliott that appears poised to make a deep playoff push in 2017 and beyond. With such a bright future, many Cowboys fans will be searching for a way to suitably express their jubilation. Enter Bayless and DJ Steve Porter with How Bout Them Cowboys. There are a few observations to be made right off the bat. 1) What is that table on set made of and where can it be ordered? The abuse it has taken without breaking is remarkable. 2) There are an extraordinary number of famous people taken aback by Skip’s passion in 90 seconds. At a glance ... “I have a feeling I’m about to unleash, you’re Cowboy hating, is it Dak is it Zeke?” “My Cowboys have arrived, are you sleep deprived? You have to eat humble pie” “Monday after Monday, they just keep making plays, here we go, every dog has its day” “Dak attack, the MVP, Dak attack, you decree, ah ah ah, kee kee kee, week after week, I do agree” “Romo need to check that team chemistry, Dez Bryant chasing greatness oh that’s fancy” 3) A special consideration should be made by the Recording Academy to bestow an honorary Grammy to the whites of Shannon Sharpe’s eyes for their performance throughout the video. Is it a hit single? Only time will tell. Will the Cowboys use this as inspiration to reach their first Super Bowl since 1996? If they do, someone better reinforce that table.
\\ \hline
Gold & 
It won't end the QB controversy, but it is catchy.
\\ \hline
Model & 
Skip Bayless is a Dallas Cowboys fan. And unlike in past years, that is worth celebrating (which Bayless does quite frequently on Undisputed). At 12-2 the Cowboys are running away from the rest of the NFC, have clinched a playoff berth and have a rookie duo in Dak Prescott and Ezekiel Elliott that appears poised to make a deep playoff push in 2017 and beyond. With such a bright future, many Cowboys fans will be searching for a way to suitably express their jubilation. Enter Bayless and DJ Steve Porter with How Bout Them Cowboys. There are a few observations to be made right off the bat.
\\ \hline
ROUGE1-F1 & 
3.36
\\ \hline
\end{tabular}
\caption{Generated summaries by $\text{PEGASUS}_{\text{LARGE}}$ (\gnews) on NEWSROOM sampled by ROUGE1-F1.}
\end{table*}

\begin{table*}[bth!]
\scriptsize
\centering
\renewcommand{\arraystretch}{1.3}
\begin{tabular}{c|p{0.85\textwidth}}
\hline
\multicolumn{2}{c}{NEWSROOM} \\ \hline
\makecell{Document\\(ID \#193)} & 
2011 - 12:33 pm | Gregory Booth, the CEO of Zippo, which makes perhaps the world’s most iconic cigarette lighters, stopped by the Forbes studio recently to talk about how to sell a commodity product – at decidedly non-commodity prices. You must be logged in to post a comment Log in with your Forbes account
\\ \hline
Gold & 
The CEO of the world's most iconic maker of cigarette lighters talks about how to sell a commodity product -- at decidedly non-commodity prices.
\\ \hline
Model & 
Gregory Booth, the CEO of Zippo, which makes perhaps the world's most iconic cigarette lighters, stopped by the Forbes studio recently to talk about how to sell a commodity product - at decidedly non-commodity prices.
\\ \hline
ROUGE2-F1 & 
61.02
\\ \hline
\makecell{Document\\(ID \#32)} & 
TESTED 2013 Lincoln MKZ Hybrid WHAT IS IT? A Lincoln-branded upscale hybrid sedan with its Ford roots showing. HOW MUCH? \$36,820 base, \$42,415 as tested. The price is the same as an equivalent nonhybrid MKZ. WHAT MAKES IT RUN? A 141-horsepower 2-liter 4-cylinder, with an 88-kilowatt electric motor and a 1.4-kilowatt-hour lithium-ion battery for 188 net horsepower; continuously variable transmission. IS IT THIRSTY? The E.P.A. rating is 45 m.p.g. all around    in the city, on the highway and combined. ALTERNATIVES Lexus ES 300h, Ford Fusion Hybrid. Part of the fun of a hybrid car is listening for the gasoline engine to start and analyzing how smoothly it makes the transition from electric to gas. Unless you floor the accelerator, the MKZ Hybrid is so quiet it  s hard to tell what  s moving you forward. With active noise control, which sends sound-canceling frequencies through the infotainment system, the car errs on the side of isolation: if engagement with the road is your goal, this isn  t your ride. Rather, the appeal is the combination of an impressive mileage rating and an understated wood-and-leather ambience. Ford calls the MKZ Hybrid   America  s most fuel-efficient luxury sedan.   Consumers are showing an active interest in upscale gas sippers, a category that never existed in the past. This is the second iteration of the MKZ Hybrid, which shares a powertrain with the Ford Fusion Hybrid. Both cars were extensively reworked for 2013, with striking    but quite different    new designs. The MKZ  s extroverted styling is one of the best things about it. The winged grille treatment recalls a subtler form on the streamlined Stout Scarab of the 1930s. The lithium-ion battery replaces the nickel-metal-hydride pack in ... ... ... 
\\ \hline
Gold & 
The appeal of the 2013 Lincoln MKZ Hybrid is its impressive 45 miles per gallon rating and understated wood-and-leather ambience.
\\ \hline
Model & 
The appeal of the Lincoln MKZ Hybrid is the combination of an impressive mileage rating and an understated wood-and-leather ambience.
\\ \hline
ROUGE2-F1 & 
52.38
\\ \hline
\makecell{Document\\(ID \#85)} & 
Go up to your attic right now, or down to your basement, or to your late parents’ storage unit that you still pay for each month even though you’re unfamiliar with its exact contents. Wherever it is you keep things that are ancient and dusty and mysterious, go there immediately and see if an elderly African American woman stares back at you from the canvas of an oil painting. Because if she does, you may have found one of the world’s rarest paintings. John Kelly writes "John Kelly's Washington," a daily look at Washington's less-famous side. Born in Washington, John started at The Post in 1989 as deputy editor in the Weekend section. “It would be the only oil painting of an African American woman who came over on a slave ship. That’s how valuable it would be,” said Jim Johnston. Jim is a Bethesda lawyer and author. His infatuation with a different oil painting inspired him to write “From Slave Ship to Harvard: Yarrow Mamout and the History of an African American Family,” published in 2012 by Fordham University Press. Yarrow Mamout was taken from West Africa in the 18th century and sold into slavery in Maryland. He was owned by a Georgetown family but eventually gained his freedom. He was well known in Georgetown, a practicing Muslim who made bricks and owned land. His portrait was painted in 1819 by Charles Willson Peale and displayed in Peale’s Philadelphia museum. Yarrow was also painted by James Alexander Simpson, a Georgetown painter. Today the Alexander painting of Yarrow hangs in the Peabody Room of the Georgetown public library branch. But in 1825 it hung at Ninth and Pennsylvania NW. We know this because the newspapers at the time were full of mentions of a new attraction: the Columbia Museum. The ... ... ... 
\\ \hline
Gold & 
In 1825, a painting of an African American woman was displayed in Washington. Where is it now?
\\ \hline
Model & 
Go up to your attic right now, or down to your basement, or to your late parents’ storage unit that you still pay for each month even though you’re unfamiliar with its exact contents. Wherever it is you keep things that are ancient and dusty and mysterious, go there immediately and see if an elderly African American woman stares back at you from the canvas of an oil painting. Because if she does, you may have found one of the world’s rarest paintings.
\\ \hline
ROUGE2-F1 & 
6.00
\\ \hline
\makecell{Document\\(ID \#163)} & 
The age when North American clubs look to England and the wider British Isles with a child’s embrace, beseeching input from the mother country’s coaching bosom to take the domestic game on, has long since passed. Or so some critics would argue. A new, forward-thinking generation of coaches not steeped in the 4-4-2 and an unadorned, direct style of football rule the waves in this epoch, they postulate. And this legion of coaches are increasingly young, fresh and, most importantly, American. While there might be some historical merit to the spirit of the argument against the British-style coach, it is perhaps itself a little outdated, not to say somewhat harsh on someone like Carl Robinson and his stylish Vancouver Whitecaps, for instance. Some 20 years on from the birth of Major League Soccer, some of the men who as players helped mold the country’s top division into what it has become today now form an integral portion of the younger cohort of the domestic coaching ranks. Broadly successful elder statesmen like Bruce Arena and Sigi Schmid are followed by a seemingly blossoming rank and file. Head coaches such as Sporting Kansas City’s Peter Vermes. DC United’s Ben Olsen. New England Revolution’s Jay Heaps. Columbus Crew’s Gregg Berhalter. New York Red Bulls’ Jesse Marsch. Behind them, too, are some well-thought-of assistants. Then there are the firmly rooted foreign coaches partly schooled in MLS ways after successful stints in the league. Robinson would be one. As would FC Dallas head coach Oscar Pareja. Related: Would a pan-American Champions League benefit soccer in the US? Yet the country has yet to see a coach leave home shores and make a name with instant recognition. Amid the chattering classes of the global game’s chief draws in Europe, that is not always easy. It is ... ... ... 
\\ \hline
Gold & 
Bryan Kay: Plenty of Americans play abroad, but far fewer have made careers as coaches of top sides outside MLS. Could former USA coach Bob Bradley break that pattern?
\\ \hline
Model & 
American players are now a common presence in the Old World top divisions of England and Germany, but no American manager has yet made a name for himself
\\ \hline
ROUGE2-F1 & 
0.00
\\ \hline
\end{tabular}
\caption{Generated summaries by $\text{PEGASUS}_{\text{LARGE}}$ (\gnews) on NEWSROOM sampled by ROUGE2-F1.}
\end{table*}

\begin{table*}[bth!]
\scriptsize
\centering
\renewcommand{\arraystretch}{1.3}
\begin{tabular}{c|p{0.85\textwidth}}
\hline
\multicolumn{2}{c}{NEWSROOM} \\ \hline
\makecell{Document\\(ID \#157)} & 
Each Tuesday, WhatIfSports.com's NFL simulation engine generates predictions and box scores for every NFL game for the coming week. Our highly sophisticated algorithms simulate every play of every game to produce each team's likelihood to win. Each matchup is simulated 501 times. To account for injuries and roster moves announced late in the week, we will be re-simulating games on Thursdays through the season. Note: our predictions use the latest available team and player information, while NFL SimMatchup includes all players that were or will be available at any point during the 2015 season so that theoretical and "what if" scenarios can be simulated. Week 4 was a “prove it” week for multiple teams, and most passed the test. The undefeated Atlanta Falcons went up against the Houston Texans in our Game of the Week, and as predicted, they continued their hot start to the season with a big win. Our Lock of the Week was the Colts winning against the Jags, and even without Andrew Luck, Indy pulled it off. For the week, our NFL simulation engine finished 9-6 and 4-4 against the spread. Nailed It: The NFL simulation engine had no problems with the Denver-Minnesota game. The engine predicted the Broncos would win by an average score of 27-20. The actual final score? Broncos 23, Vikings 20. Wide Right: It’s fitting that the Buffalo Bills show up here at least once, right? The NFL simulation engine predicted the Bills would beat the Giants by an average score of 26-20 and win 64.4 percent of the time, but the Giants took care of business and upset the Bills 24-10 instead. Track our 2015 performance or view our accuracy last season. Week 5 Game to Watch Seattle at Cincinnati: The undefeated Bengals will play host to the reigning NFC champs ... ... ... 
\\ \hline
Gold & 
View WhatIfSports.com's NFL predictions for Week 5.
\\ \hline
Model & 
View WhatIfSports.com's NFL predictions for Week 5.
\\ \hline
ROUGEL-F1 & 
100.00
\\ \hline
\makecell{Document\\(ID \#127)} & 
Poverty comes in many forms: a middle-aged man making \$10 an hour, a single mother with mouths to feed, a person with a criminal record and few employers willing to look past it. But the path out is often the same: acquiring a new skill, whether it’s taking college classes or learning how to boost a lousy credit score. Within the next few years, nearly two-thirds of all job openings in the United States will require at least some sort of certification, according to Georgetown University research. But funding for career training has declined drastically. In Massachusetts, the state budget for programs that provide these services has been slashed in half in the past decade. Soon, however, lower-income residents may have more tools to lift themselves up. President Obama just launched an initiative to help workers upgrade skills, and last month, the White House held a summit with employers, nonprofits, tech innovators, and unions committed to expanding and improving educational opportunities for millions of workers. “It’s very, very, very difficult to move up without accumulating higher-level skills,” said Jerry Rubin, chief executive of Jewish Vocational Service, a Boston nonprofit that provides worker training. “Without them, people will be employed, but they will be at or below the poverty line.” Many who have lifted themselves out of desperate situations say they could not have gotten there on their own — or without a lot of effort. Here are a few of their stories. A study by a Boston nonprofit found that poor people often face high interest rates that make it nearly impossible to pay down debts. Nathaniel Awan spent last fall learning how to build a tool box and a spice rack, among other construction skills, just a few miles from where he grew up in the Four Corners neighborhood ... ... ... 
\\ \hline
Gold & 
Poverty comes in many forms, but the path out is often the same: acquiring a new skill.
\\ \hline
Model & 
Poverty comes in many forms: a middle-aged man making \$10 an hour, a single mother with mouths to feed, a person with a criminal record and few employers willing to look past it.
\\ \hline
ROUGEL-F1 & 
23.53
\\ \hline
\makecell{Document\\(ID \#258)} & 
The chief executive of Wells Fargo, one of the country’s largest banks, stumbled his way through more than four hours of blistering questioning by House members Thursday, struggling to defend the bank over a scandal involving its sales practices. For the second time in as many weeks, John G. Stumpf sat before lawmakers and apologized for the scandal that resulted in the firing of 5,300 employees and a \$185 million fine. This go-round didn’t go much better than the first time. "Fraud is fraud and theft is theft. What happened at Wells Fargo over the course of many years cannot be described any other way," said Rep. Jeb Hensarling (R-Tex.), chairman of the House Financial Services Committee. Wells Fargo has turned into a "school for scoundrels," said Rep. Carolyn B. Maloney (D-N.Y.). "I’ve come to the conclusion that Wells Fargo should be broken up,” said Rep. Maxine Waters (D-Calif.). "It’s too big to manage and I’m moving forward to break up the bank.” Stumpf, 63, stayed calm as he questioned by more than a dozen members of the committee. But he often struggled to finish a sentence before lawmakers cut him off to ask him another question or complain that he was dodging the current one. "I am fully accountable for all unethical sales practices in our retail banking business, and I am fully committed to fixing this issue, strengthening our culture, and taking the necessary actions to restore our customers’ trust," he told the committee. After being pummeled in the Senate last week, Stumpf told the House committee he recommended that the company’s board rescind \$41 million of unvested stock he had been awarded. Carrie Tolstedt, who led the bank's community banking unit where the misconduct occurred, will have to forfeit about \$19 million, and both executives will not ... ... ... 
\\ \hline
Gold & 
Stumpf agreed to forfeit \$41 million after the bank became entangled in sales scandal.
\\ \hline
Model & 
The chief executive of Wells Fargo, one of the country’s largest banks, stumbled his way through more than four hours of blistering questioning by House members Thursday, struggling to defend the bank over a scandal involving its sales practices.
\\ \hline
ROUGEL-F1 & 
14.81
\\ \hline
\makecell{Document\\(ID \#95)} & 
“Y'all think a little toy buzzer is going to keep the truth from tumbling out from these jaws?” With news that Olivia's (other) ex Edison Davis (Norm Lewis) is joining the race, Shondaland has completed the fantasy: two women, a Hispanic man, a black man and a racist redneck who is almost certainly but not actually Donald Trump. 'SCANDAL' RECAP SEASON 5, EPISODE 14: BILLIONAIRE BUFFOON RUNS FOR PRESIDENT — BUT NOT THE ONE YOU'RE THINKING OF Yes somehow, this election still isn't interesting. Cyrus (Jeff Perry) is fighting for control of Francisco Vargas (Ricardo Chavira) with the candidate's brother, which is mind-numbingly boring. Olivia (Kerry Washington) and Huck (Guillermo Diaz) have set up a focus group on Mellie's (Bellamy Young) campaign, which lends itself to a hilariously ridiculous new slogan: “Mellie Grant, woman of the people.” Remember when she got drunk on hooch and ate fried chicken for months because a secret society killed her son? Quinn (Katie Lowes) goes undercover as a WASP to gather information on Vanessa, because Liv's still freaking out about Jake's (Scott Foley) love life. And Hollis Doyle is shooting really big guns and posing for cameras. Election 2016, y'all. It's weird in all realities. Mellie's trek to Gettysburger to relate to the commonfolk backfires when she tells reporters that she and Karen (remember her?) used to go every Sunday after church. But — dun-dun-dun! — Gettysburger is closed on Sundays. So we have Burgergate. End the meme. End it now. 'SCANDAL' RECAP 5X12: FITZ IS DATING, CYRUS IS SCHEMING, OLIVIA IS SPYING After some more WASP undercover work by Quinn, we finally get some marginally interesting information: Jake is stealing money from Vanessa to give to Papa Pope (Joe Morton) to fund a Super Pac for Edison. Which, at least somewhat, explains why ... ... ... 
\\ \hline
Gold & 
With news that Olivia's (other) ex Edison Davis (Norm Lewis) is joining the race, Shondaland has completed the fantasy.
\\ \hline
Model & 
“Y'all think a little toy buzzer is going to keep the truth from tumbling out from these jaws?”
\\ \hline
ROUGEL-F1 & 
10.26
\\ \hline
\end{tabular}
\caption{Generated summaries by $\text{PEGASUS}_{\text{LARGE}}$ (\gnews) on NEWSROOM sampled by ROUGEL-F1.}
\end{table*}

\begin{table*}[bth!]
\scriptsize
\centering
\renewcommand{\arraystretch}{1.3}
\begin{tabular}{c|p{0.85\textwidth}}
\hline
\multicolumn{2}{c}{Multi-News} \\ \hline
\makecell{Document\\(ID \#114)} & 
Size really does seem to matter when it comes to cancer risk. Being tall undoubtedly has its benefits. You can see in a crowd and grab objects off high shelves. But with the good comes the bad. The taller you are, the higher your odds of developing cancer, and a new paper has added weight to this. key points Key points: Taller people have more cells in their body, as well as higher levels of a protein that encourages cells to divide and grow For every 10cm over the average height, a person's risk for cancer increases 10 per cent New analysis of data from big cancer studies supports this, and also finds a few specific cancers to be more or less strongly correlated with height Leonard Nunney, an evolutionary biologist at the University of California, Riverside, looked at massive cancer databases to find out how the number of cells in a person's body, using height as a proxy, might affect their risk of developing cancer. Reporting in the Proceedings of the Royal Society B, he found being taller and having more cells did mean more cancer overall: For every 10 centimetres over the average height, the risk of developing any cancer increased by around 10 per cent. This fits with previous studies. "If you were comparing a 5-foot guy to a basketball player who's over 7 feet tall, then that basketball player has around twice the risk of cancer across the board," Professor Nunney said. He also found that taller people were at much higher risk of melanoma, and women specifically had greater odds of developing thyroid cancer. But it doesn't mean tall people should panic. "Now, you can't do anything about your height, but what you can do is tell extremely tall individuals that they should be aware ... ... ... 
\\ \hline
Gold & 
– A new study on cancer won't make for pleasant reading for tall people. Consider this quote from lead researcher Leonard Nunney of the University of California Riverside: "If you were comparing a 5-foot guy to a basketball player who's over 7 feet tall, then that basketball player has around twice the risk of cancer across the board," he tells Australia's ABC. Yes, Nunney's study in the Proceedings of the Royal Society B suggests that taller people are at greater risk for cancer. The surprisingly simple reason? They have more cells, and thus more things go can wrong in those cells, explains the Guardian. Specifically, the study found that cancer risk rises by 10\% for every 4 inches people are above average height—5 feet 4 for women and 5 feet 9 for men, reports CNN. The findings held true for 18 of 23 cancers studied. "Now, you can't do anything about your height, but what you can do is tell extremely tall individuals that they should be aware of this, and if they have any concerns, to get checked," says Nunney. Cancers of the colon and kidney were among those with the strongest correlation to height. A researcher not involved with the study says it backs up previously seen links between height and cancer
\\ \hline
Model & 
– Taller people are more likely to develop cancer than shorter people, according to a new study. The study, published in the Proceedings of the Royal Society B, found that for every 10 centimeters over the average height of a person, the risk of developing cancer increases by 10\%. "If you were comparing a 5-foot guy to a basketball player who's over 7 feet tall, then that basketball player has around twice the risk of cancer across the board," Leonard Nunney, an evolutionary biologist at the University of California, Riverside, tells Australia's News Network. The study also found that taller people were at much higher risk of melanoma, thyroid cancer, and cervical cancer. "Now, you can't do anything about your height, but what you can do is tell extremely tall individuals that they should be aware of this, and if they have any concerns, to get checked," Nunney says. "It just suggests more vigilance, I think." One theory is that taller people have more cells in their bodies, which encourages them to grow. But Nunney also found that cancers in the skin, stomach, and mouth were more closely linked to height. "Across other cancers, there's no real geographical variation," he says.
\\ \hline
ROUGE1-F1 & 
65.88
\\ \hline
\makecell{Document\\(ID \#145)} & 
Image copyright Crossrail Image caption The plague victims' bones reveal clues to their harsh lives in medieval London Skeletons unearthed in London Crossrail excavations are Black Death victims from the great pandemic of the 14th Century, forensic tests indicate. Their teeth contain DNA from the plague bacterium Yersinia pestis and their graves have been dated to 1348-50. Records say thousands of Londoners perished and their corpses were dumped in a mass grave outside the City, but its exact location was a mystery. This discovery solves a 660-year-old mystery. It's a hugely important step forward Jay Carver , Lead archaeologist, Crossrail Archaeologists now believe it is under Charterhouse Square near the Barbican. They plan to expand their search for victims across the square - guided by underground radar scans, which have picked up signs of many more graves. Crossrail's lead archaeologist Jay Carver says the find "solves a 660-year-old mystery". "This discovery is a hugely important step forward in documenting and understanding Europe's most devastating pandemic," he said. "Further excavations will follow to see if - as we expect - we are coming across a much bigger mass burial trench." Media playback is unsupported on your device Media caption The teeth of the skeletons contain plague bacterial DNA Between 1347 and 1351 the "Great Pestilence" swept westward across Europe killing millions of people. It later became known as the Black Death. The plague The plague is one of the oldest identifiable diseases known to man Plague is spread from one rodent to another by fleas, and to humans either by the bite of infected fleas or when handling infected hosts Recent outbreaks have shown that plague may reappear in areas that have long been free of the disease Plague can be treated with antibiotics such as streptomycin and tetracycline Source: World ... ... ... 
\\ \hline
Gold & 
– Skeletons dug up in London last year are indeed the remains of people who died from the Black Plague—and who suffered a tough life before falling ill, the BBC reports. Forensic analysis shows that teeth taken from at least four of the 12 corpses discovered during excavation for a rail line contained trace amounts of plague DNA, indicating exposure. Early burials found at the site, from the late 1340s, are nice and orderly, with bodies wrapped in white shrouds, but skeletons from a second outbreak in the 1430s are tossed in with what appear to be upper-body injuries—evidence of "a period of lawlessness and social breakdown," Phys.org reports. Among other significant details: Several skeletons suffered from malnutrition and 16\% had rickets. Many had back damage, signalling stressful manual labor. Analysis of one victim is amazingly detailed: He was born outside of London, breastfed, had bad tooth decay as a boy, and worked in manual labor before dying as a young man of the Black Death. Archaeologists suspect that thousands more Black Death victims lie nearby, and a dig is planned for this summer. DNA experts are analyzing the plague genome in victims' teeth in case there's more to learn about the disease, which still kills 2,000 people per year (
\\ \hline
Model & 
– It's a "hugely important" find, one that could shed light on the Black Death, the plague that wiped out at least 60\% of London's population between 1347 and 1351, the BBC reports. Archaeologists digging for railway tracks in London's Crossrail station uncovered 25 skeletons, along with pottery dated to the mid-14th century, that have DNA from the bacterium Yerinia pestis dated to 13 to 50 years old. That's a long way from the 660-year-old record showing that thousands of Londoners died from the plague and their bodies were dumped in a mass grave outside the City of London. "We can start to answer questions like: where did they come from and what were their lives about?" says Crossrail archaeologist Jay Carver. "I'm amazed of the reasons for why the Black Death was so devastating." DNA from 12 of the skeletons has confirmed that they had contact with the plague prior to their death, and researchers plan more excavations to see if they're coming across a much bigger plague trench. "We can see that Londoners weren't living an easy life," Carver says, "and that's possibly one of the explanations for why the Black Death was so devastating." Antibiotic-resistant strains of the plague
\\ \hline
ROUGE1-F1 & 
34.58
\\ \hline
\end{tabular}
\caption{Generated summaries by $\text{PEGASUS}_{\text{LARGE}}$ (\gnews) on Multi-News sampled by ROUGE1-F1.}
\end{table*}

\begin{table*}[bth!]
\scriptsize
\centering
\renewcommand{\arraystretch}{1.3}
\begin{tabular}{c|p{0.85\textwidth}}
\hline
\multicolumn{2}{c}{Multi-News} \\ \hline
\makecell{Document\\(ID \#261)} & 
A murder suspect remains on the run following a weekend shooting in Montpelier. It left a former Vermont Frost Heaves basketball player dead. WCAX News is learning the victim and the shooter may have been friends. It appears to have been a night out at the bar gone wrong. Jayveon Caballero, 29, is wanted for second-degree murder. Nearly 48 hours after he allegedly shot a man to death in Montpelier, police still don't know where he is. "He may have gotten on a bus yesterday morning in White River Junction, a Greyhound bus, heading south," said Maj. Glenn Hall, Vermont State Police. Police say Caballero fatally shot Markus Austin, 33. A friend told investigators Austin was invited out with friends to join Caballero, who he called "my boy Jay." But at the end of the night a fight broke out. "We do know from the investigation that there was an altercation outside Gusto's bar in Barre," said Hall. The alleged shooter's girlfriend, Desiree Cary, 22, told police that she got mad and hit Austin as they left the bar. Austin allegedly hit her back, injuring her jaw bone. Court documents show Caballero appeared upset and then went to find Austin at a Barre Street apartment. "He walked over to the victim's car and shot him and walked back," said Jeff Sarpong, witness. Sarpong says he woke up to someone yelling outside and a gunshot, and then looked out his window. "I see him turn around, wave with a gun," said Sarpong. He said Caballero appeared to show sympathy just seconds after pulling the trigger. "He then just kept on asking, 'Are you OK?'" Sarpong said. "After he noticed what he had done and that he wasn't responsive, he just drove away," said Sarpong. In an attempt to locate Caballero, police ... ... ... 
\\ \hline
Gold & 
– Police in the country's smallest state capital are investigating something that hasn't happened there in living memory: a murder. Police in Vermont say they have issued an arrest warrant for 29-year-old Jayveon Caballero, who is accused of fatally shooting Markus Austin in Montpelier early Sunday after an altercation, the Times Argus reports. Witnesses told investigators that Austin, 33, assaulted Caballero’s girlfriend, Desiree Cary, during the altercation outside a bar in the nearby city of Barre early Sunday. Police say Caballero later confronted Austin outside his apartment and shot him dead. The body was found in the parking lot of Austin's apartment complex. A Vermont State Police spokesman says Caballero, who knew Austin before the shooting, fled the town and probably the state after the shooting, WCAX reports. The police spokesman says Caballero is believed to have boarded a Greyhound bus bound for New York City on Sunday morning, which made numerous stops in other states along the way. Later Sunday, Cary, 22, was arrested on apparently unrelated charges of selling heroin and crack cocaine. With around 7,500 residents, Montpelier has the smallest population of any state capital, the AP notes. Police Chief Anthony Facos says this is the city's first murder since a woman shot her husband in the 1920s.
\\ \hline
Model & 
– A man has been shot dead outside an apartment complex in what police say is the first murder in the nation's smallest capital city in almost a century. An arrest warrant was issued for Jayveon Caballero on a murder charge following the Sunday morning shooting in Montpelier, the AP reports. Police were seeking Caballero in the killing of Markus Austin, whose body was found in the apartment complex's parking lot. Caballero is believed to have boarded a bus on Sunday in White River Junction. Police said the bus was headed to New York City but had several intermediate stops where he could have gotten off. Police were seeking Caballero in the killing of Austin, whose body was found in the apartment complex's parking lot. Caballero and Austin had been involved in an altercation outside Gustos bar in earlier Sunday morning, police said. During the altercation Austin assaulted Caballero's girlfriend, Desiree Cary, witnesses said. The victim, 33-year-old Markus Austin, was shot outside his apartment at 191 Barre Street in Montpelier around 4:30am Sunday. Caballero later confronted Austin outside Austin's apartment and shot him, police said. Caballero's girlfriend, pleaded not guilty to drug charges and was released. Montpelier has about 7,500 residents, making it the nation's least populous capital.
\\ \hline
ROUGE2-F1 & 
25.81
\\ \hline
\makecell{Document\\(ID \#176)} & 
Sandra Chesterman, left, puts a braclet on the wrist of Christine Kirby during a birthday party for her son Jayden Kirby, 1, on Sunday, Feb. 9, 2014, in Fremont, Calif. Chesterman's daughter Kristina was killed by an alleged drunk driver last September while coming home from nursing school in Chico. Jayden was given months to live before he received part of Kristina Chesterman's liver. Jayden and Christina share the same birth stone and that gem is embeded into the bracelet. (Aric Crabb/Bay Area News Group) ( ARIC CRABB ) LIVERMORE -- When she was still in high school, Kristina Chesterman wrote out her bucket list. Flying a plane was on it; so was running through a poppy field and breaking up a fight between two boys over her affections. She also wanted to save a life. The aspiring nurse's ambitions came to a halt in September, when she was killed by a suspected drunken driver near Chico State, where she attended school. Though Chesterman, 21, didn't get to mark much off her list, she has saved more lives than she hoped -- and is profoundly affecting many others. Five Northern Californians have been saved through Kristina's choice to donate her organs. And her grateful friends and family are making the rest of her bucket list their own. A photograph of Kristina Chesterman is displayed during a birthday party for Jayden Kirby on Sunday, Feb. 9, 2014, in Fremont, Calif. Chesterman was killed by an alleged drunk driver last September while coming home from nursing school in Chico. Jayden, 1, was given months to live before he received part of Kristina Chesterman's liver. (Courtesy of the Chesterman Family) ( Chesterman Family ) Chesterman's mother, Sandra, of Livermore, said her daughter wanted to help people from an early age. She routinely gave ... ... ... 
\\ \hline
Gold & 
– Kristina Chesterman, 21, was studying to be a nurse when she was killed by a suspected drunk driver last year—but she managed to save lives anyway. A registered organ donor, Chesterman gave five people, including a baby, new life, and now the woman who received her heart wants to do something in return. Susan Vieira, 64, has vowed to check off everything on Chesterman's bucket list—written down on a piece of paper her mom only recently found. (One of the items? "Save someone's life." Another? "Be in four places at once.") Vieira had completed several of the tasks already, including learning to fly a plane and riding a camel, ABC News reports. And now, "together, we will finish her bucket list," Vieira says. Other points on the list include running through a poppy field and riding in a hot-air balloon, the San Jose Mercury News reported earlier this year. "I'd like to think all the things I continue to accomplish in my life, I’m taking Kristina with me," Vieira continues; Chesterman's mom adds that she "felt an instant connection" to the woman she just met. Chesterman's friends are also helping to complete the bucket list, including
\\ \hline
Model & 
– When 21-year-old Kristina Chesterman was killed by a suspected drunk driver in September, she left behind a bucket list of things she wanted to do before she died. Among them: fly a plane, break up a fight between two boys over her affections, and donate her organs. Now, five Northern Californians have been saved thanks to her decision, the San Jose Mercury News reports. "I know she would've been so proud," says Chesterman's mother, Sandra. "It hasn't been an easy process for us, but at the same time, it's brought us so much comfort." One of those saved was 1-year-old Jayden Kirby, who might have died as an infant had he not received part of Chesterman's liver. Jayden's mother, Christine, figured out the name of the mystery donor after hearing news reports, the Mercury News reports. "I wanted her to see the impact Kristina had," she says. "(The transplant) absolutely saved (Jayden's) life." Chesterman's aunt, Patricia Picard, texted the couple, asking if Zak Pappachan was a match. Within hours, he had his new kidney. Now he's enjoying being a dad to his young daughter, Ray
\\ \hline
ROUGE2-F1 & 
10.97
\\ \hline
\end{tabular}
\caption{Generated summaries by $\text{PEGASUS}_{\text{LARGE}}$ (\gnews) on Multi-News sampled by ROUGE2-F1.}
\end{table*}

\begin{table*}[bth!]
\scriptsize
\centering
\renewcommand{\arraystretch}{1.3}
\begin{tabular}{c|p{0.85\textwidth}}
\hline
\multicolumn{2}{c}{Multi-News} \\ \hline
\makecell{Document\\(ID \#263)} & 
This July 24, 2016 photo provided by Niels Alpert, Betsy Davis, third from left, has a laugh with her friends during a going away party in Ojai, Calif. In early July, Davis emailed her closest friends... (Associated Press) SAN DIEGO (AP) — In early July, Betsy Davis emailed her closest friends and relatives to invite them to a two-day party, telling them: "These circumstances are unlike any party you have attended before, requiring emotional stamina, centeredness and openness." And just one rule: No crying in front of her. The 41-year-old artist with ALS, or Lou Gehrig's disease, held the gathering to say goodbye before becoming one of the first Californians to take a lethal dose of drugs under the state's new doctor-assisted suicide law for the terminally ill. "For me and everyone who was invited, it was very challenging to consider, but there was no question that we would be there for her," said Niels Alpert, a cinematographer from New York City. "The idea to go and spend a beautiful weekend that culminates in their suicide — that is not a normal thing, not a normal, everyday occurrence. In the background of the lovely fun, smiles and laughter that we had that weekend was the knowledge of what was coming." Davis worked out a detailed schedule for the gathering on the weekend of July 23-24, including the precise hour she planned to slip into a coma, and shared her plans with her guests in the invitation. More than 30 people came to the party at a home with a wraparound porch in the picturesque Southern California mountain town of Ojai, flying in from New York, Chicago and across California. One woman brought a cello. A man played a harmonica. There were cocktails, pizza from her favorite local joint, and a ... ... ... 
\\ \hline
Gold & 
– Betsy Davis spent the last years of her life losing control of her body as her illness worsened—but she could still control how she died, and she "turned her departure into a work of art," friend and cinematographer Niels Alpert says of the California artist's death. The 41-year-old, who in 2013 was diagnosed with ALS, also known as Lou Gehrig's disease, threw a party for more than 30 friends and family that lasted through the weekend before her doctor-assisted suicide late last month, People reports. Davis, a painter and performance artist who invited guests from across the country to what she called a "rebirth," died under the state's physician-assisted suicide law, which took effect on June 9. "You're all very brave for sending me off on my journey," she wrote in her invitation, which warned guests that "emotional stamina" would be required, the AP reports. "There are no rules," she wrote. "Wear what you want, speak your mind, dance, hop, chant, sing, pray, but do not cry in front of me. OK, one rule." After a weekend that included music, pizza, cocktails, individual chats with every guest, and a screening of favorite movie The Dance of Reality,
\\ \hline
Model & 
– In early July, Betsy Davis emailed her closest friends and relatives to invite them to a two-day party, telling them: "These circumstances are unlike any party you have attended before, requiring emotional stamina, centeredness, and openness." And just one rule: No crying in front of her. The 41-year-old artist with ALS, or Lou Gehrig's disease, held the gathering to say goodbye before becoming one of the first Californians to take a lethal dose of drugs under the state's new doctor-assisted suicide law for the terminally ill, the AP reports. Davis worked out a detailed schedule for the gathering on the weekend of July 23-24, including the precise hour she planned to slip into a coma, and shared her plans with her guests in the invitation. More than 30 people came to the party at a home with a wraparound porch in the picturesque Southern California mountain town of Ojai, flying in from New York, Chicago, and across California. One woman brought a cello. A man played a harmonica. There were cocktails, pizza from her favorite local joint, and a screening in her room of one of her favorite movies, The Dance of Reality, based on the life of a Chilean film director. As the weekend drew to a close, her friends kissed
\\ \hline
ROUGEL-F1 & 
23.06
\\ \hline
\makecell{Document\\(ID \#87)} & 
The text on the video above was edited on June 1, 2018 to make it clear that it is impossible to know why the polar bear pictured was starving. An earlier version of the video went too far in suggesting that climate change was responsible ( read more ). This story was updated on January 19, 2018 to reflect the more specific location of where the photographs were taken. When photographer Paul Nicklen and filmmakers from conservation group Sea Legacy arrived on Somerset Island—near the larger Baffin Island—in the Canadian Arctic in late summer, they came across a heartbreaking sight: a starving polar bear on its deathbed. Nicklen is no stranger to bears. From the time he was a child growing up in Canada's far north the biologist turned wildlife photographer has seen over 3,000 bears in the wild. But the emaciated polar bear, featured in videos Nicklen published to social media on December 5, was one of the most gut-wrenching sights he's ever seen. "We stood there crying—filming with tears rolling down our cheeks," he said. Video shows the polar bear clinging to life, its white hair limply covering its thin, bony frame. One of the bear's back legs drags behind it as it walks, likely due to muscle atrophy. Looking for food, the polar bear slowly rummages through a nearby trashcan used seasonally by Inuit fishers. It finds nothing and resignedly collapses back down onto the ground. In the days since Nicklen posted the footage, he's been asked why he didn’t intervene. "Of course, that crossed my mind," said Nicklen. "But it's not like I walk around with a tranquilizer gun or 400 pounds of seal meat." And even if he did, said Nicklen, he only would have been prolonging the bear's misery. Plus, feeding wild polar bears ... ... ... 
\\ \hline
Gold & 
– An estimated 2.5 billion people saw the image: a starving polar bear struggling across an Arctic landscape. "The mission was a success, but there was a problem: We had lost control of the narrative," writes Cristina Mittermeier in National Geographic. Accompanied by a photographic team, she snapped shots of the dying bear last year while colleague Paul Nicklen shot video—all part of their "mission to capture images that communicate the urgency of climate change," she writes. "When Paul posted the video on Instagram, he wrote, 'This is what starvation looks like.'" He also wondered if all 25,000 polar bears would die like this and urged people to reduce their carbon footprint. But he didn't say climate change had killed this bear. That nuance vanished when National Geographic posted the video with the subtitles "This is what climate change looks like" and the Washington Post ran a dramatic headline about "gut-wrenching" images. People also responded dramatically, expressing gratitude for validation of climate science, anger the crew hadn't fed the bear, or a stubborn refusal to acknowledge global warming. "Perhaps we made a mistake in not telling the full story—that we were looking for a picture that foretold the future and that we didn’t know what had happened to this particular polar bear
\\ \hline
Model & 
– "We stood there crying—filming with tears rolling down our cheeks." That's Paul Nicklen's take on a video he posted in December showing a starving polar bear on an island in the Canadian Arctic, National Geographic reports. The video was shot by Nicklen, a biologist and contributor to National Geographic, during a trip with the conservation group Sea Legacy. It shows the bear on Somerset Island, near the larger Baffin Island, as it searches a garbage can for food. At one point, the bear shows signs of starvation. "When scientists say bears are going extinct, I want people to realize what it looks like," Nicklen tells National Geographic. "Bears are going to starve to death. This is what a starving bear looks like." The video has since been viewed more than 3.5 million times, and Nicklen says it struck a nerve. "If you have any conscience at all, this video of a starving polar bear will break your heart," Newsweek states.
\\ \hline
ROUGEL-F1 & 
17.23
\\ \hline
\end{tabular}
\caption{Generated summaries by $\text{PEGASUS}_{\text{LARGE}}$ (\gnews) on Multi-News sampled by ROUGEL-F1.}
\end{table*}

\begin{table*}[bth!]
\scriptsize
\centering
\renewcommand{\arraystretch}{1.3}
\begin{tabular}{c|p{0.85\textwidth}}
\hline
\multicolumn{2}{c}{Gigaword} \\ \hline
\makecell{Document\\(ID \#269)} & 
zairean rebels , led by laurent-desire kabila , on saturday rejected calls by the united nations for a ceasefire , saying it could only be called after talks with kinshasa .
\\ \hline
Gold & 
zairean rebels reject un call for ceasefire
\\ \hline
Model & 
zairean rebels reject un ceasefire call
\\ \hline
ROUGE1-F1 & 
92.31
\\ \hline
\makecell{Document\\(ID \#208)} & 
a zairean diplomat was among three african men arrested for smuggling \#\# kilograms -lrb- \#\#\# pounds -rrb- of marijuana into sweden , the swedish news agency tt reported friday .
\\ \hline
Gold & 
zairean diplomat among three arrested for pot
\\ \hline
Model & 
zairean diplomat arrested for drug smuggling
\\ \hline
ROUGE1-F1 & 
61.54
\\ \hline
\makecell{Document\\(ID \#198)} & 
prime minister benjamin netanyahu has proposed completing talks on the final status of palestinian territories in the next six months instead of the scheduled two years , according to israeli public television .
\\ \hline
Gold & 
UNK netanyahu wants to finish talks in six months
\\ \hline
Model & 
netanyahu proposes completing final status talks in six months
\\ \hline
ROUGE1-F1 & 
55.56
\\ \hline
\makecell{Document\\(ID \#92)} & 
reinsurer hannover re ag said thursday that its second-quarter net profit doubled as demand increased for its products , particularly life and health reinsurance .
\\ \hline
Gold & 
hannover re s \#nd-quarter profit doubles
\\ \hline
Model & 
hannover re says \#q net profit doubled
\\ \hline
ROUGE1-F1 & 
42.86
\\ \hline
\makecell{Document\\(ID \#281)} & 
the us energy department -lrb- doe -rrb- announced tuesday that an experimental project in canada to inject carbon dioxide into oil fields for its permanent storage in geologic formations proved to be successful .
\\ \hline
Gold & 
carbon dioxide permanent storage in oil field a success : us energy department
\\ \hline
Model & 
canadian project to store carbon dioxide in oil wells successful
\\ \hline
ROUGE1-F1 & 
36.36
\\ \hline
\makecell{Document\\(ID \#149)} & 
los angeles might have lost its bid for an expansion team to houston on wednesday , but fans who like to watch football and the tv networks were winners .
\\ \hline
Gold & 
no l.a. team good for tv viewers
\\ \hline
Model & 
los angeles loses bid for nfl team
\\ \hline
ROUGE1-F1 & 
26.67
\\ \hline
\makecell{Document\\(ID \#286)} & 
football fans in the bangladesh capital dhaka will be able to watch the world cup live on big screens at \#\# city locations , officials said friday .
\\ \hline
Gold & 
the afp world news summary
\\ \hline
Model & 
world cup on big screens in dhaka
\\ \hline
ROUGE1-F1 & 
16.67
\\ \hline
\makecell{Document\\(ID \#258)} & 
the suicide bomb attacks in saudi arabia were    a cowardly and disgraceful terrorist atrocity , '' prime minister tony blair said wednesday .
\\ \hline
Gold & 
two britons missing after saudi suicide blasts
\\ \hline
Model & 
blair calls saudi attacks cowardly and disgraceful terrorist atrocity
\\ \hline
ROUGE1-F1 & 
12.50
\\ \hline
\makecell{Document\\(ID \#223)} & 
leading the preakness field at the far turn , edgar prado sensed victory as he prepared peace rules for the final push to the finish line .
\\ \hline
Gold & 
prado watches potential victory fade in stretch
\\ \hline
Model & 
peace rules wins preakness
\\ \hline
ROUGE1-F1 & 
0.00
\\ \hline
\makecell{Document\\(ID \#39)} & 
chelsea owner roman abramovich and england captain david beckham top english soccer 's rich lists .
\\ \hline
Gold & 
second test scoreboard
\\ \hline
Model & 
abramovich beckham top rich list
\\ \hline
ROUGE1-F1 & 
0.00
\\ \hline
\end{tabular}
\caption{Generated summaries by $\text{PEGASUS}_{\text{LARGE}}$ (\gnews) on Gigaword sampled by ROUGE1-F1.}
\end{table*}

\begin{table*}[bth!]
\scriptsize
\centering
\renewcommand{\arraystretch}{1.3}
\begin{tabular}{c|p{0.85\textwidth}}
\hline
\multicolumn{2}{c}{Gigaword} \\ \hline
\makecell{Document\\(ID \#42)} & 
a national human rights museum under the planned ministry of culture is scheduled to open on jan. \# , \#\#\#\# , council for cultural affairs chairman emile sheng said friday .
\\ \hline
Gold & 
human rights museum to open in \#\#\#\#
\\ \hline
Model & 
national human rights museum to open in \#\#\#\#
\\ \hline
ROUGE2-F1 & 
90.91
\\ \hline
\makecell{Document\\(ID \#57)} & 
greece international central defender sotiris kyrgiakos says he intends to join liverpool after receiving an offer from the premier league club .
\\ \hline
Gold & 
kyrgiakos says intends to join liverpool
\\ \hline
Model & 
greece defender kyrgiakos to join liverpool
\\ \hline
ROUGE2-F1 & 
40.00
\\ \hline
\makecell{Document\\(ID \#201)} & 
world number three david nalbandian said wednesday that he was optimistic of being fit to compete in wimbledon after being forced to retire injured in the french open semi-finals last week .
\\ \hline
Gold & 
nalbandian optimistic for wimbledon fitness
\\ \hline
Model & 
nalbandian optimistic of wimbledon return
\\ \hline
ROUGE2-F1 & 
25.00
\\ \hline
\makecell{Document\\(ID \#13)} & 
the weather was cool during georgia tech 's football practice monday , but coach george o'leary was hot and sweaty after the workout .
\\ \hline
Gold & 
o'leary gets defensive after loss to bc
\\ \hline
Model & 
o'leary feels the heat
\\ \hline
ROUGE2-F1 & 
18.18
\\ \hline
\makecell{Document\\(ID \#36)} & 
japanese electronics maker sharp corp. announced plans friday to ramp up its output of flat screen televisions with new production lines in japan and mexico in response to flourishing demand .
\\ \hline
Gold & 
sharp to expand flat tv production in UNK mexico
\\ \hline
Model & 
sharp to ramp up flat screen tv output
\\ \hline
ROUGE2-F1 & 
13.33
\\ \hline
\makecell{Document\\(ID \#291)} & 
nearly a year ago , south carolina supreme court chief justice jean UNK said that fair , swift justice has taken a backseat to an assembly line process that diminishes the quality of hearings afforded south carolinians .
\\ \hline
Gold & 
fair swift justice demands diversity
\\ \hline
Model & 
s.c. supreme court chief justice backs off criticism
\\ \hline
ROUGE2-F1 & 
0.00
\\ \hline
\makecell{Document\\(ID \#210)} & 
charlie UNK baffled the cubs with his knuckleball , matt kemp and casey blake homered , and the los angeles dodgers defeated chicago \#-\# in the national league to win their third game in a row on saturday .
\\ \hline
Gold & 
UNK s UNK fools cubs as dodgers win \#-\#
\\ \hline
Model & 
dodgers beat cubs for third straight win
\\ \hline
ROUGE2-F1 & 
0.00
\\ \hline
\makecell{Document\\(ID \#173)} & 
marc rosset upset boris becker in straight sets to pace switzerland over germany in the opening round of the world team cup monday , while spain beat sweden \#-\# .
\\ \hline
Gold & 
switzerland spain win opening matches
\\ \hline
Model & 
rosset upsets becker in world team cup
\\ \hline
ROUGE2-F1 & 
0.00
\\ \hline
\makecell{Document\\(ID \#96)} & 
in chicago , frank thomas and albert belle hit consecutive homers for the seventh time this season , and chicago beat baltimore to stop a six-game losing streak .
\\ \hline
Gold & 
toronto : the sixth
\\ \hline
Model & 
thomas belle power cubs past orioles
\\ \hline
ROUGE2-F1 & 
0.00
\\ \hline
\makecell{Document\\(ID \#59)} & 
his leftist fringe party does n't even have \#\#\# members , but \#\#-year-old budiman UNK is accused of being at the heart of a communist plot to topple the government .
\\ \hline
Gold & 
UNK weekly sports calendar
\\ \hline
Model & 
fringe party leader accused of plotting to topple government
\\ \hline
ROUGE2-F1 & 
0.00
\\ \hline
\end{tabular}
\caption{Generated summaries by $\text{PEGASUS}_{\text{LARGE}}$ (\gnews) on Gigaword sampled by ROUGE2-F1.}
\end{table*}

\begin{table*}[bth!]
\scriptsize
\centering
\renewcommand{\arraystretch}{1.3}
\begin{tabular}{c|p{0.85\textwidth}}
\hline
\multicolumn{2}{c}{Gigaword} \\ \hline
\makecell{Document\\(ID \#246)} & 
a group of puerto rican businessman will visit chile to explore possible new markets , this u.s. commonwealth 's economic development bank announced .
\\ \hline
Gold & 
puerto rican trade mission to visit chile
\\ \hline
Model & 
puerto rican businessmen to visit chile
\\ \hline
ROUGEL-F1 & 
76.92
\\ \hline
\makecell{Document\\(ID \#248)} & 
wednesday 's evening rubber prices in singapore cents per kilo provided by the singapore commodity exchange : UNK
\\ \hline
Gold & 
evening rubber prices
\\ \hline
Model & 
wednesday 's evening rubber prices
\\ \hline
ROUGEL-F1 & 
75.00
\\ \hline
\makecell{Document\\(ID \#2)} & 
after spending \#\# years in prison for a killing he did n't commit , ken marsh finally is poised to get some payback .
\\ \hline
Gold & 
\$ \#\#\#,\#\#\# for \#\# years wrongly held in prison
\\ \hline
Model & 
man exonerated after \#\# years in prison
\\ \hline
ROUGEL-F1 & 
50.00
\\ \hline
\makecell{Document\\(ID \#15)} & 
prime minister of antigua and barbuda baldwin spencer left here monday for hong kong , winding up his four-day visit to shanghai .
\\ \hline
Gold & 
antigua and barbuda pm leaves shanghai for hong kong
\\ \hline
Model & 
antigua and barbuda prime minister ends shanghai visit
\\ \hline
ROUGEL-F1 & 
47.06
\\ \hline
\makecell{Document\\(ID \#188)} & 
cuban sports officials and coaches censored the possible elimination of boxers ' protective head gear in the amateur boxing competitions , cuba 's official press    granma '' said on friday .
\\ \hline
Gold & 
cuba sports UNK UNK ask to keep boxers protective gear
\\ \hline
Model & 
cuban sports officials censor boxers head gear
\\ \hline
ROUGEL-F1 & 
35.29
\\ \hline
\makecell{Document\\(ID \#164)} & 
russia 's defense industry has been badly hit by the global financial crisis , deputy prime minister sergei ivanov , a former defense minister , said tuesday .
\\ \hline
Gold & 
russia 's defense sector hit by financial crisis : govt official
\\ \hline
Model & 
financial crisis hits russian defense industry
\\ \hline
ROUGEL-F1 & 
25.00
\\ \hline
\makecell{Document\\(ID \#99)} & 
one day after lakers owner jerry buss left his interest in phil jackson subject to interpretation , new york knicks president isiah thomas did anything but , telling reporters thursday that he believed a second meeting with jackson soon would take place .
\\ \hline
Gold & 
jackson at top of knicks list
\\ \hline
Model & 
thomas says he 'll meet with jackson
\\ \hline
ROUGEL-F1 & 
15.38
\\ \hline
\makecell{Document\\(ID \#265)} & 
us defense secretary donald rumsfeld said thursday the killing of al-qaeda 's leader in iraq , abu musab UNK , was a significant victory in the battle against terrorism but not the end of the violence .
\\ \hline
Gold & 
rumsfeld calls zarqawi death significant victory
\\ \hline
Model & 
rumsfeld hails killing of al-qaeda leader in iraq
\\ \hline
ROUGEL-F1 & 
13.33
\\ \hline
\makecell{Document\\(ID \#242)} & 
rankings reflect sales for the week ending oct. \#\# , at almost \#,\#\#\# bookstores plus wholesalers serving \#\#,\#\#\# other retailers -lrb- gift shops , department stores , newsstands , supermarkets -rrb- , statistically weighted to represent all such outlets nationwide .
\\ \hline
Gold & 
best sellers : UNK books
\\ \hline
Model & 
top \#\# book sales
\\ \hline
ROUGEL-F1 & 
0.00
\\ \hline
\makecell{Document\\(ID \#51)} & 
the rope on the flagpole is still broken .
\\ \hline
Gold & 
british reopen their baghdad embassy but they ca n't call it that yet
\\ \hline
Model & 
UNK rope still broken
\\ \hline
ROUGEL-F1 & 
0.00
\\ \hline
\end{tabular}
\caption{Generated summaries by $\text{PEGASUS}_{\text{LARGE}}$ (\gnews) on Gigaword sampled by ROUGEL-F1.}
\end{table*}

\begin{table*}[bth!]
\scriptsize
\centering
\renewcommand{\arraystretch}{1.3}
\begin{tabular}{c|p{0.85\textwidth}}
\hline
\multicolumn{2}{c}{WikiHow} \\ \hline
\makecell{Document\\(ID \#295)} & 
Make it a priority to shower often in order to smell clean and avoid foul odors. Try to brush your teeth, comb your hair, and upkeep any other part of the body before leaving for school. Feeling fresh not only looks good, but you’ll feel rejuvenated and confident.Most boys begin to have a more distinct body odor in middle school. Combat this by wearing deodorant. Avoid using strong colognes such as Axe unless it is in small proportions. If the smell is overbearing, ladies won’t want to hang out too close. If you feel the need to use cologne, use it sparingly. Try choosing a deodorant that smells refreshing with a hint of musk.; , Image is important for first impressions. Don’t give her the impression that you have a shabby personality because of your clothing. Upgrade some of your t-shirts, jeans, and button down shirts for nicer versions. Swap out your jeans for corduroy, slacks, or khakis. Grab some polos instead of a band t-shirt. Try out a flannel instead of hoodie.Even if your middle school requires uniforms, you can still pull off a sharp look by caring for your clothes. Do not let your clothes get shabby or smelly. As a general rule, don’t wear a top more than one day. Pants should only be worn two to three times before a wash. While looks are not the only thing that matters, it’s often the first thing that a girl will notice about you. Confidence is something that girls can easily pick up on. Try setting small goals for yourself like learning the piano (middle school is the perfect age to begin an instrument). Exercise is another way to feel comfortable in your body and boost your self-esteem. You can reach a confident state of mind by dressing ... ... ... 
\\ \hline
Gold & 
Maintain your daily hygiene. Dress well. Feel good in your skin. Have good posture.
\\ \hline
Model & 
Practice good hygiene. Dress well. Be confident. Maintain a good posture. Be friendly.
\\ \hline
ROUGE1-F1 & 
51.85
\\ \hline
\makecell{Document\\(ID \#56)} & 
Lip-plumping products come in many forms: glosses, balms, sticks, gels, and pots. Applying these to your lips can temporarily make your lips look fuller, often by irritating them.The plumping effect will only last for a couple of hours, but you can renew it by reapplying the product to your lips. Note that the effects won’t be as drastic as those achieved through cosmetic surgery.; , Ingredients like cinnamon, ginger mint, wintergreen, and capsicum will boost blood flow to your lips, causing them to redden and swell — i.e., to look fuller., If you wish to combine your lip-plumping product with a lipstick or gloss, apply the plumper to your lips first for maximum effectiveness. Dermatologists recommend against overusing lip-plumping products, as these may cause your lips to become dry and scaly.Try to save the lip plumpers for special occasions. If you want more from your lip-plumping products, consider trying a treatment plumper. Manufacturers claim that treatment plumpers can stimulate your lips into producing more collagen and elastin, making your lips fuller for a longer period of time. Treatment plumpers are available for purchase online and in person at shops that carry beauty products.They may be more expensive than traditional plumpers. Common ingredients in treatment plumpers include peptides, marine collagen, and human growth factors.
\\ \hline
Gold & 
Invest in a lip-plumping product. Know which lip-plumping ingredients to look for. Use a lip plumper as a primer. Don’t overuse lip plumpers. Look into treatment plumpers.
\\ \hline
Model & 
Apply a lip-plumping product. Use natural lip-plumping ingredients. Combine lip-plumping products. Try a treatment plumper.
\\ \hline
ROUGE1-F1 & 
50.00
\\ \hline
\makecell{Document\\(ID \#95)} & 
To do this they should offer a variety of ways to collect customer card details and process payments. These are often defined by your own technical expertise. It could mean providing a simple means of redirecting customers from your site to their hosted payment pages. Or maybe you require a way for you to authorise payments remotely. Whatever your requirement, a dedicated PSP should have the solutions to meet it. One size does not fit all, so don’t let anyone persuade you otherwise.; , A fixed monthly fee starts at around £10, but there are some cheaper options available, starting as low as £0.05 per transaction. As a rule of thumb, the higher your transaction volume, the lower the unit cost. And remember to compare like with like. Ask if it’s a complete solution with inclusive fees or whether you’ll be paying the bank separately on top of that. Will the PSP with those ‘oh-so-attractive’ prices charge huge set-up fees for enabling services – such as deferred payments, a virtual terminal, or subscriptions – that come as standard with other PSPs?. Look beyond the headline figures and consider the overall long-term operational cost of partnering with each PSP. For example, does the PSP have limited support for fraud management? The cost of flimsy security could do much more damage to your bottom line than any fractional differences in rates. As an online merchant you need to make sure you can also manage them as efficiently as possible. Which is why an effective PSP should provide you with the right reporting and administration tools to manage your merchant services... Will you have access to real-time transaction statements? Can you export valuable customer data? Is the reporting sophisticated enough to allow you to answer questions that can better target future sales. For ... ... ... 
\\ \hline
Gold & 
Enquire about their technical solutions Technically your PSP must be capable of supporting all of your plans. Look for genuinely competitive rates Costs (set-up fees and cost per transaction) can vary greatly between competing PSPs, so you need to shop around for the best rates. Establish if there are any upfront costs The real costs associated with accepting online card payments do not start and end at the rate quote. Manage reporting and administration facilities Accepting online payments is only one part of the e-commerce equation. Enquire about their technical support A service is not a service unless it’s fully backed up. Find out if you would be protected against fraud When accepting card payments online, fraud is potentially the biggest threat to your profits. Make sure you can trust their services Online, the difference between a prospect and a customer is trust. Look out for a full service history Okay, it’s not a car, but the principle is the same.
\\ \hline
Model & 
Establish what your PSP needs to do to meet the needs of your online merchant business. Look at the pricing offered by the PSP. Consider the level of support you’ll get from the PSP. Look at the tools you’ll be using to manage your merchant services. Consider the need for fraud management systems (PMS).There’s no doubt that the quality of a PSP’s fraud solution will depend upon its support for merchant services solutions – in other words, its ability to meet the needs of your online merchant business. Check the reliability of the PSP’s customer service.
\\ \hline
ROUGE1-F1 & 
31.58
\\ \hline
\makecell{Document\\(ID \#3)} & 
If you hang up on a telemarketer right away, you’ll probably be placed on the “no answer” list, and will be called again eventually. You don’t want to engage a telemarketer in conversation either, especially if you have absolutely no interest in their product or service. The easiest way to handle a telemarketer is to say, “Please put me on your do not call list.” If the telemarketer keeps interrupting you or a robot calls you, you might just have to hang up. If the same number persistently calls, report it to the FCC by calling 1-888-CALL-FCC or going to https://www.fcc.gov/. If you live in the United States, you can add yourself to the do not call registry. It'll then become illegal for telemarketers to contact you: https://www.donotcall.gov/. If you do get contacted by a telemarketer after putting yourself on the list, you can report the number that called you to the FTC or FCC. If they tell you a mutual acquaintance or other contact, consider taking the call. If they don’t or can’t tell you, or if they got your contact information from some data pool or list, ask to be removed from their call list and hang up., Be clear about conveying your annoyance with phone calls, especially if you might want to do business with or purchase something from that company but don’t want annoying callers. Make sure you provide them with an email address specifically for spam or sales deals.
\\ \hline
Gold & 
Don’t just hang up right away. Add yourself to the national do not call registry. Ask where they got your contact info. Tell them “No calls, emails only.” If you’re running a business, in charge of purchasing, or otherwise need to field sales pitches from time to time, tell telemarketers that you only respond to emails.
\\ \hline
Model & 
Hang up immediately. Report persistent callers. Ask to be removed from the call list. Be clear about your preferences.
\\ \hline
ROUGE1-F1 & 
20.78
\\ \hline
\end{tabular}
\caption{Generated summaries by $\text{PEGASUS}_{\text{LARGE}}$ (\gnews) on WikiHow sampled by ROUGE1-F1.}
\end{table*}

\begin{table*}[bth!]
\scriptsize
\centering
\renewcommand{\arraystretch}{1.3}
\begin{tabular}{c|p{0.85\textwidth}}
\hline
\multicolumn{2}{c}{WikiHow} \\ \hline
\makecell{Document\\(ID \#29)} & 
Microwave frozen spinach for two minutes at 50\% power. Break the spinach apart into smaller pieces. Microwave it again, at 50\% power, checking and stirring it whenever a minute passes by, until it’s defrosted.Alternately, you can use one packed cup of fresh baby spinach, washed and finely chopped with the stems removed.; , Wrap the spinach in a clean kitchen towel. Squeeze and wring the towel to remove the liquid.Another option is to press the wet spinach into a sieve over the sink., Make sure your hands are clean and free of jewelry. Place the ground meat in the bowl first, and press it down in the middle to form a well. In the well, put the spinach, garlic, egg, milk, bread crumbs, Parmesan cheese, salt and pepper, and all but three tablespoons of the chopped onion.Add a small dash of hot sauce, if desired.Since you’re working with you’re hands, wash them well first, with soap and a nail brush. If you wear any rings, remove them and set them aside somewhere safe (away from the sink drain).You may want to crack the egg into a separate bowl before adding it, to ensure that no egg shell gets into the meat mixture. Preheat the oven to 400$^\circ$ F (204$^\circ$ C). Mix the meat mixture well. Make twelve large meatballs, approximately 11  2” each., Place the meatballs on a nonstick cookie sheet or shallow baking pan. Drizzle extra-virgin olive oil over the meatballs.Cook the meatballs for twenty to twenty-eight minutes, until they are no longer pink., Add butter and a drizzle of extra-virgin olive oil to a small sauce pot over medium heat. Allow the butter to melt.Pour in the mushrooms and/or chopped onion. Cook until the vegetables are tender, about two minutes., Whisk in the flour, and cook for one minute. ... ... ... 
\\ \hline
Gold & 
Defrost the spinach. Drain defrosted spinach. Combine the ingredients in a large mixing bowl. Form the meatballs with your hands. Cook the meatballs in the oven. Sautee mushrooms and/or onions. Add flour, milk and stock. Mix in cheese and seasonings. Combine meatballs with sauce. Serve meatballs with spaghetti squash. Char broccoli in the oven. Serve meatballs with starch.
\\ \hline
Model & 
Defrost the spinach. Drain the spinach. Combine the spinach and other ingredients in a large bowl. Make the meatballs. Cook the meatballs. Make the sauce. Make the sauce. Serve the meatballs and sauce. Make the spaghetti squash. Cook the spaghetti squash. Serve the meatballs.
\\ \hline
ROUGE2-F1 & 
25.74
\\ \hline
\makecell{Document\\(ID \#193)} & 
While such snacks are not necessarily healthy, they do tend to be lower calorie than other candy options. If you want to buy snacks in the theater, these are generally your best choice of candy. In addition to being lower calorie than other varieties, they are chewier. This means they're more difficult to eat. If you take longer to finish your candy, you're less likely to return to the concession stand for a second helping.Twizzlers have the lowest calorie count, containing only about 165 calories. Milk duds contain 370 calories, while Starbursts contain 480 calories. Keep in mind, if you're on a specific diet plan that restricts sugar, this is not the best route for you. While these snacks are lower calorie, the bulk of the calories come from sugar.; , Many movie theaters now sell protein bars at the counter. You can also sneak one in from a nearby gas station or drug store. Lower calorie protein bars may contain some ingredients you're craving, like nuts and chocolate, while still having some nutrients.Remember to read nutritional labels. Some protein bars may be misleading. In general, it's best to skip bars with a lot of added sugar. Swedish Fish are much lower in sugar than the average movie snack. They also tend to have a tougher texture, making chewing them difficult. If you slow down the pace of your eating, you will eat less overall.A serving of Swedish Fish is about 19 pieces, roughly a large handful, and contains only 140 calories and 29 grams of sugar. Movie theater containers tend to be smaller, so you may find a container that has a single serving. If you can't find a smaller container, try splitting one with a friend. As it can be tricky to find candy that's lower calorie at ... ... ... 
\\ \hline
Gold & 
Go for Twizzlers, Milk Duds, or Starbursts. Buy a protein bar. Look for Swedish Fish. Ask about the theater's food policy.
\\ \hline
Model & 
Buy chewier candy. Try a lower calorie protein bar. Go for Swedish Fish. Find a smaller container. Bring your own snacks.
\\ \hline
ROUGE2-F1 & 
19.51
\\ \hline
\makecell{Document\\(ID \#189)} & 
Never interrupt, shout over, or insult another person (even if you feel that they really, really deserve it). Religion and government are separate, and religion and abortion rights are separate. Condemning someone's entire religion is cruel and divisive. Not all religious people are anti-choice; for example, it is possible for some self identified Christians to be pro-choice. Even if the person in front of you is a flaming bigot, think of the nice religious people, and don't insult their religion. Pro-choice people have long pointed out how "baby" and "unborn child" are manipulative ways to refer to an unwanted fetus. But calling all fetuses "clumps of cells" is also an insensitive use of rhetoric, because there are some fetuses that are truly loved and wanted, and to those parents the fetus is far more special. Stick with the factual term fetus. Dismissing fetuses as mere "clumps of cells" can be deeply upsetting to people who miscarried a loved, wanted fetus.Don't accidentally minimize the hardship of losing a beloved fetus. Transgender men, nonbinary people, and some intersex people of any gender can also get pregnant. Be sure not to step on the toes of transgender people as you fight for the rights of everyone with a uterus. You are not obligated to change anyone's mind, nor do you have to keep talking to someone who says very rude or upsetting things. This may be personally upsetting to you. Say "I don't want to talk about this anymore," "I need a break," or "I need to get some air" if you can't keep calm. Any decent person will respect your wishes. If the person makes you feel unsafe, leave right away. Call police if you are being threatened or stalked.
\\ \hline
Gold & 
Be polite and courteous when defending your beliefs. Never insult someone else's religion. Don't manipulate rhetoric. Remember that not all pregnant people are women. Take a break if you need to.
\\ \hline
Model & 
Don't demonize the other person's religion. Don't call all fetuses "clumps of cells.", Stay calm.
\\ \hline
ROUGE2-F1 & 
12.24
\\ \hline
\makecell{Document\\(ID \#272)} & 
Instead of eating a ton of sugar, which is tempting to do when you're tired, try eating complex carbohydrates and protein. Sugar can give you a rush of energy, but then your energy will crash. Complex carbohydrates, on the other hand, offer more sustained energy, especially when you pair them with protein.For instance, try some natural peanut butter on whole-wheat bread or a piece of fruit with a slice of cheese. Dehydration can lead to fatigue. Therefore, staying hydrated will help keep your energy up. Try sipping on water throughout the day to make sure you get enough. The Institute of Medicine recommends that men drink 13 cups and women drink 9 cups of water every day.Other liquids can contribute to your water intake, such as juice and coffee. However, drinking too much juice can pack on extra calories and cause a sugar crash, so try to limit your intake. Also, though you can count caffeinated beverages, you shouldn't make them the majority of the liquid you drink in a day., Keeping something in your mouth, such as a piece of gum, can help increase your alertness. The best time for this type of trick is when you're in a meeting that you absolutely need to stay awake for.
\\ \hline
Gold & 
Skip the sugar. Make hydration a priority. Try a piece of gum.
\\ \hline
Model & 
Eat the right foods. Stay hydrated. Chew something.
\\ \hline
ROUGE2-F1 & 
0.00
\\ \hline
\end{tabular}
\caption{Generated summaries by $\text{PEGASUS}_{\text{LARGE}}$ (\gnews) on WikiHow sampled by ROUGE2-F1.}
\end{table*}

\begin{table*}[bth!]
\scriptsize
\centering
\renewcommand{\arraystretch}{1.3}
\begin{tabular}{c|p{0.85\textwidth}}
\hline
\multicolumn{2}{c}{WikiHow} \\ \hline
\makecell{Document\\(ID \#241)} & 
No matter what size the paint spill, carefully use a putty knife or any sort of flat tool to scoop up paint that is sitting on top of the carpet, not yet soaked down into the fibers.Scoop up as much excess paint as you can without spreading the paint around. Wipe off the collected paint with a paper towels, and throw the towels away in the garbage.; , Use a paper towel or clean rag to blot and absorb as much wet paint as you can.Lift your towel up and down and gently dab at the paint. Adjust the towel so you are always dabbing at the paint with a clean section. Don’t rub or try to scrub the paint out; that will merely spread the paint around and push it deeper into the carpet. Mix a cleaning solution of 1 cup of warm water with 1  4 - 1 teaspoon of mild dish soap.Use another clean towel to drip on a small amount of cleaning solution onto the paint. Let the cleaning solution sit on the carpet for a few of minutes., After the solution has had some time to soak on the paint, dip a clean paper towel or rag into the cleaning solution and continue to blot at the wet paint.Always adjust the cloth so you are blotting with a clean section. Keep blotting at the stain until your cloth no longer absorbs any of the paint.Blot from the outside of the paint, moving inward; this will prevent you from spreading out the stain even further.If the stain still remains, repeat the cleaning process with the soapy water, or move on to using vinegar to blot at the paint.Vinegar works well with removing water-based paints like latex paint., Once the paint stain is removed, use a ... ... ... 
\\ \hline
Gold & 
Collect excess paint. Blot the paint with a clean cloth. Mix a cleaning solution. Blot the paint with the cleaning solution. Dry the area. Call in a professional.
\\ \hline
Model & 
Scoop up excess paint. Blot at the paint with a paper towel. Apply a cleaning solution to the wet paint. Blot at the paint with a clean cloth. Dry the carpet. Call a professional carpet cleaner.
\\ \hline
ROUGEL-F1 & 
59.37
\\ \hline
\makecell{Document\\(ID \#150)} & 
This area does not need to be perfectly square or rectangular unless you are planning on using sod instead of simply replanting your grass. Drive wooden or plastic stakes around the area to be leveled.; , Attach strings to stretch between the stakes, a few inches above the ground. Attach a string level to the strings to determine the high point. This will usually be your starting point and the rest of the ground will be brought to here, but you can also lower some ground if that makes more sense for your project. Using a tape measure and your level, adjust the strings until you can see how much height needs to be added or taken away across the area you are adjusting. Keep in mind that you may want to adjust the grade of your land to help combat drainage problems. Your land should be graded 1" for every 4', leading away from your home. If you are leveling a small area and it is mostly level, you probably will not need to remove the grass. However, if you have a large area and a lot of leveling to do, removing the grass will be much easier. A simple shovel is all that's necessary for a reasonable amount of space. Depending on how much ground you have to cover and what will be going on the ground afterwards, you'll need to level your land with differing mixtures of soil, sand, and compost/manure fertilizer. If you want to grow grass in this area, the cover will need to be nutrient rich. If you are simply wanting to place a small shed or pool, soil and sand will do just fine. Use a garden rake to spread the material evenly, checking using your level and a measuring tape to make ... ... ... 
\\ \hline
Gold & 
Stake off your area to level. Use a string level. Adjust the strings. Adjust for grade. Remove the grass if necessary. Add your ground cover. Spread the topsoil. Tamp the soil. Let it settle. Spread your seeds. Lightly cover with more soil. Water lightly. Reseed as necessary. Alternatively, buy sod.
\\ \hline
Model & 
Determine the shape of the area to be leveled. Set a string level. Adjust the string level. Decide if you will need to remove the grass. Level the land. Compact the soil. Wait for the soil to settle. Purchase the grass. Plant the grass.
\\ \hline
ROUGEL-F1 & 
29.79
\\ \hline
\makecell{Document\\(ID \#266)} & 
For the best texture and taste, insist on meat that is organic and/or free of artificial hormones.Specify that you want a raw pork belly that hasn’t already been cured or sliced. For a fattier bacon, ask for meat that came from the hog’s belly and/or chest. For a meatier bacon, ask for meat that covered the hog’s spare ribs.Whole pork bellies typically weigh between 10 and 12 pounds, but if you are experimenting with cure recipes, ask for a smaller cut in case you end up not liking your concoction.; , If you purchased your pork belly from a commercial vendor other than a farm and plan on curing it as soon as you get home, skip this step, since the meat has already been chilled. But if you buy direct from a farm, ask the farmer how long ago the hog was slaughtered, because you need to chill your meat within 24 hours of that time.In either case, bring along a cooler stocked with ice to keep the pork belly relatively cool on the ride home if it takes you more than half an hour.Once home, place the pork belly in a sealable container to keep your fridge clean and prevent cross-contamination with other items. For a quicker chill, set the container inside the fridge wherever it is coldest. To further reduce the chance of cross-contamination, set it in the meat compartment, away from other products. Keep the refrigerator’s temperature at 40 degrees Fahrenheit and let the pork belly sit until its core temperature drops to 42 degrees, which will help kill bacteria. If you are chilling multiple bellies, set each one in its own container rather than stacking the meat directly on top of each other. Either ask your butcher to do it for you when you purchase ... ... ... 
\\ \hline
Gold & 
Buy a fresh pork belly. Chill your meat. Cut the skin off. Decide on a curing agent. Mix a basic cure. Try a saltier cure with sage and thyme. Cure your bacon with honey. Rub the cure into the meat. Refrigerate the pork belly. Rinse your meat. Store your meat safely. Finished.
\\ \hline
Model & 
Purchase a pork belly. Chill the pork belly. Slice the skin off the pork belly. Mix the curing salts. Soak the pork belly in the salts. Rub the pork belly with the curing salts. Let the pork belly rest. Slice the pork belly. Fry the pork belly. Slice the pork belly.
\\ \hline
ROUGEL-F1 & 
27.18
\\ \hline
\makecell{Document\\(ID \#38)} & 
Read books and learn all you can about dementia so you better understand what you are dealing with. There are a wide range of products available to help make both the dementia patient and the caregivers life easier.; , People with dementia are for the most part caring, smart, funny and always wanting to help others. They respond to kindness and they need a lot of patience. Sometimes people with dementia become anxious or angry. At these times diversion is most important. They like to be kept busy with games, crafts, singing, or just talking about the past. Their abilities will be different because the stage of their dementia is different. It will be hard for a mate not to grieve the loss of his/her mate of many years, or for the child not to grieve the loss of a loving parent. Attend support groups and seminars, and know you are not alone. For those who are caring for relatives suffering with dementia, there are a range of dementia products and dementia aids designed to make both carer and patients' lives easier. Many drug stores and medical supply shops may carry them and you can find sites on the web which carry a whole range of items specifically geared towards patients with dementia. These products cover a range of categories, such as memory aids for dementia, mobility support aids, products to improve bathing safety, assistive devices, nutritional items for specific needs, products to help deal with and manage incontinence, special clothing and aids for vision and hearing. Some of these products are designed to help daily life. Dementia products such as mobility aids and hearing devices provide essential day to day support and good nutritional products can provide valuable calories, nutrients and support a balanced diet, or act as a ... ... ... 
\\ \hline
Gold & 
Look for information. Be kind and patient. Use diversion. Find support. Consider buying dementia aids.
\\ \hline
Model & 
Educate yourself. Be patient and kind. Keep them occupied. Don't forget that people with dementia are different. Get support. Look for dementia products.
\\ \hline
ROUGEL-F1 & 
20.51
\\ \hline
\end{tabular}
\caption{Generated summaries by $\text{PEGASUS}_{\text{LARGE}}$ (\gnews) on WikiHow sampled by ROUGEL-F1.}
\end{table*}

\begin{table*}[bth!]
\scriptsize
\centering
\renewcommand{\arraystretch}{1.3}
\begin{tabular}{c|p{0.85\textwidth}}
\hline
\multicolumn{2}{c}{Reddit TIFU} \\ \hline
\makecell{Document\\(ID \#185)} & 
this happened 5/6 years ago so my whole family every xmas day goes around to my aunties for celebrations. my cousin (of course) was there and he asked if i wanted to play cops and robbers. i accepted of course. now, next to the side of my aunts house is a little area with a small fence, a covered water tank and super duper sharp stones. my cousin (who was the cop) was gaining on me. i (tried) to jump over the fence, aaand i failed the jump and went crashing onto the gravel, my leg hitting the sharpest bit and, then the next thing i knew it had a nasty gash.
\\ \hline
Gold & 
played cops and robbers with my cousin on xmas day, gashed my leg failing a jump
\\ \hline
Model & 
i tried to play cops and robbers and ended up with a nasty gash in my leg.
\\ \hline
ROUGE1-F1 & 
42.42
\\ \hline
\makecell{Document\\(ID \#295)} & 
this happened about 3 years ago at my best friend's apartment. i had come over to meet our friend's new boyfriend and go out for a fun night of bar hopping and meeting up with more friends. i wanted to do my makeup and hair since i had showered before coming over and didn't have time prior. now, my best friend keeps her bathroom extremely hot since she has a pet snake and it was the middle of winter...so i started to kinda sweat and remembered that i forgot deodorant and i was two seconds from smelling like a hot dude in a sauna with very poor hygiene. i start panicking and getting even more sweaty when i notice the spray bottle on a small shelf. the outside of it was black and purple, i wrongly assumed it was safe to put on the ol pits and go crazy spraying them nice and good. i felt better, panic attack relieved and i come out of the bathroom. fast forward maybe 30 seconds and i start to feel like my under arm skin is slowly and mercilessly being ripped off. i'm wincing in pain and the three of them are just like, what...the hell? i confess and tell what i did and my best friend just starts dying from laughter, then goes into a fit of concern because i need to get the crap off of my skin asap. i washed it off and had really really horrible chemical burns on my arm pits for about 10 days and couldn't put deodorant on for the duration while my skin healed. needless to say i'm sure everyone loved my natural smell for that period of time... probably one of the more painful (mostly painful) and a little embarrassing moments of my life, ... ... ... 
\\ \hline
Gold & 
i sprayed home fragrance on my pits, burned my skin and smelled gross for over a week/uncomfortable discussion with my family doctor
\\ \hline
Model & 
forgot deodorant in my best friend's bathroom and sprayed it on my arm pits.
\\ \hline
ROUGE1-F1 & 
31.58
\\ \hline
\makecell{Document\\(ID \#204)} & 
throwaway here for obvious reasons.. today my friends and i decided to go off-roading in nowhereland. we packed up all our stuff, made the roughly hour drive off to the mountains to make a fire, go fishing and just talk about life until we got too tired to stay any longer. we got everything packed up and brought along one of my friends' dog because she's awesome and loves the outdoors. the dog was flipping out in the suv on the way to the path because she knew was a kick-ass day she was about to have breaking out of her normally lame, domesticated dog life. my friends decided to drink during the off-roading adventure, which was fine because i volunteered to drive since i cannot drink alcohol (mouth is wired shut [long story but i can't drink alcohol for a while]) so we were playing it safe. the dog couldn't be any happier and was about to jump out of the truck (literally) when we got there so the dog's owner let her get out and run along side of us while we drove the dirt road up to the destination for the fire. as i was driving, the dog went in and out of vision, mostly biting the tires as most dogs do, playing around. the owner kept asking us (the two guys up front) if we could see her. we said yes, and kept driving. as i was driving at no more than 5-10mph along the dirt road, i could hear the dog biting at the tires playfully, but we just laughed it off bc we thought she was having fun. the horrible, seconds-long event that ensued was me feeling the dreaded 'double-thud' under the tires and heard the dog yelp in pain. i instantly stopped the ... ... ... 
\\ \hline
Gold & 
we were off roading, dog jumped in the way of the truck i was driving, i killed the dog by accident.
\\ \hline
Model & 
i ran over my friends' dog with an suv while off-roading.
\\ \hline
ROUGE1-F1 & 
24.24
\\ \hline
\makecell{Document\\(ID \#297)} & 
i and a friend i will call tony were meeting two other friends (steven and logan) for dinner before our upcoming gaming session. steven and logan arrive at smash burger about a minute before us. we beheld a sight that will forever be ingrained in my mind. i just wish i could have memorialized it for posterity. it had been snowing that day, and there were 5 inches already on the ground. steven and logan pulled into the parking lot in his truck. logan gets out of the truck but slips and falls on a patch of ice covered under the snow. steven walks around the truck to see where logan disappeared. he was greeted by the site of a baby seal flopping on the ice floor struggling to get up. when tony and i pulled in, steven was dragging logan across the parking lot to a drier patch of ground. he was regaining his dignity just as steven was telling us the story. i just wish i had the sense to capture the moment. but, my mind was still processing what i was seeing. dammit, this almost surpasses steven's own [embarrassing flabalanche of 2009](http://www.reddit.com/r/askreddit/comments/977hp/anyone\_with\_pictures\_or\_video\_of\_a\_400lb\_man/)
\\ \hline
Gold & 
went to meet friends for dinner. was met with the sight of a snow eel being dragged over a patch of ice. laughed too hard to take picture.
\\ \hline
Model & 
friend dragged a baby seal across a parking lot covered in snow.
\\ \hline
ROUGE1-F1 & 
20.00
\\ \hline
\makecell{Document\\(ID \#213)} & 
obligatory "not today" but last thursday night/friday morning in the wee hours. this is a \_loooong\_ one so bear with me. not throw-away but the embarrassment is real. i have an rv. for thanksgiving week, my wife, daughter, and i traveled to myrtle beach and stayed in an ocean-side spot. this year, my wife's best friend from childhood and her son flew down and stayed with us making it a bit more crowded than usual but we enjoyed it. wife and i slept in our bed in the back, kids slept in the bunks, wife's friend slept in the pull out in the living room. worked out just fine! flash to late thursday night. i woke up not feeling great. a bit nauseous, stomach doing backflips, kind of loopy. it got to the point that i almost woke my wife to get her to grab a garbage can because i didn't know if i'd make it to the bathroom to puke if it came to that. i eventually slipped back to sleep... only to be awoken by a strident, loud beeping. **alarm!** my sleepy brain first tried to convince me it was just an alarm clock but i popped up and reoriented myself and realized it was the combination carbon monoxide/propane detector going off. i quickly popped out of bed and got down on my hands and knees to press the silence button -- the alarm is mounted on side of the bed frame directly below me -- while my brain tried to figure out what was going on. my immediate thought was "false alarm". these things are (supposedly) notoriously buggy and tend to deteriorate over time. i pulled it out and groggily read everything on it but couldn't really tell what it was complaining about. as i continued to ... ... ... 
\\ \hline
Gold & 
propane gas alarm thought we were all going to die... from my own apparent flatulence.**
\\ \hline
Model & 
i almost killed my wife's best friend with carbon monoxide poisoning.
\\ \hline
ROUGE1-F1 & 
7.41
\\ \hline
\end{tabular}
\caption{Generated summaries by $\text{PEGASUS}_{\text{LARGE}}$ (\gnews) on Reddit TIFU sampled by ROUGE1-F1.}
\end{table*}

\begin{table*}[bth!]
\scriptsize
\centering
\renewcommand{\arraystretch}{1.3}
\begin{tabular}{c|p{0.85\textwidth}}
\hline
\multicolumn{2}{c}{Reddit TIFU} \\ \hline
\makecell{Document\\(ID \#293)} & 
so, context. i currently have summer job working in a local amusement park in a small town within northern ireland. one of the requirements for this job is to operate the ghost train carts as they go in one door and come out the other, while allowing customers on. using a operating desk, with three buttons on it, i am able to control when cars go in, when the ones in the middle can be moved up and when the ones that have just come out the bottom (with people in them) can be moved up once the customer has left the cart. so...when the customer comes out the bottom door of the ghost train, they are about a 20ft gap away from due to the space for the other carts to take up. because of this i try to make them leave their cart asap to prevent another coming out and crashing into them, and some times i'll press the button to control that cart just a tad to move it so they get the idea to move out quickly. one day a woman comes out the bottom of the ghost train....and just sits there. it's a busy day, im trying to deal with a lot of customers, and i have 6 other people breathing down my neck trying to get on. i tell them i need to get this woman off before they can go on. so, i hit the bottom and shunt her car while she's only starting to try and sit up, she looks shocked at me. i do it again, she looks even more shocked. i start staring at her, implying "please hurry up. you're taking your time" and she slowly moves her way out of the car, grabbing the bars around her and slowly ... ... ... 
\\ \hline
Gold & 
i tried to get a woman out of her ghost train cart quickly by shunting it, i forgot she was disabled.
\\ \hline
Model & 
i shunted a woman with cerebral palsy out of a ghost train.
\\ \hline
ROUGE2-F1 & 
19.35
\\ \hline
\makecell{Document\\(ID \#259)} & 
yesterday, i tried to cook in a bit of a rush so i could get to work on time. i set the heat to high to heat up my noodles quickly, and in my haste i didn’t notice that some grease had spilled on the eye earlier in the day. i left the noodles to cook, and went to the bathroom. i come back, and the pot is engulfed in flames. i panicked, moved the pot off of the stove, and put out the fire. after that, i opened up the doors in the house to let it air out so the fire alarm wouldn’t go off. now, i had two of my family’s dogs locked up so they wouldn’t run out. but, i couldn’t find the third. i had assumed that he was in my sister’s room, and resumed trying to save the food. well, my sister came downstairs after a few minutes and was looking for her dog. turns out, the dog had escaped. so, i of course had to stop cooking so i could look for the dog. it took about twenty minutes to find him, and i ended up being late for work and getting chewed out by my manager.
\\ \hline
Gold & 
tried to cook in a rush, almost burned my house down, had a dog escape, and ended up being late to work.
\\ \hline
Model & 
tried to cook noodles, ended up setting the house on fire, and lost my sister’s dog in the process.
\\ \hline
ROUGE2-F1 & 
15.00
\\ \hline
\makecell{Document\\(ID \#251)} & 
so a little background first: i'm on my school's student council, and today was our junior prom assembly. we had a lot going on, much of it relying on a fast internet connection to work (live cameras streamed over appletv, etc.) we realized that as everyone settled into the auditorium, we'd have a large portion of the student body connecting to the school wifi, and we were worried about slowdown. our live camera was basically useless without a good internet connection, so i had the bright idea to hook up the appletv to my personal hotspot. it connected automatically, i ran the feed for a few seconds, worked like a charm, so we moved on. assembly starts, about 15 minutes in and everything is going great. i'm backstage when i see our principal run up to microphone and tell everyone to exit the school immediately, as there is a potential terror threat in the area (i live in an area where things like this never happen; we've had one lockdown in my 10 years of schooling here) obviously, everyone's more than a little worried. everyone files out, myself included. as i'm standing outside, a friend walks up and tells us what he found out from a teacher. one of the students, while trying to connect to the wifi, had "discovered" a network... my network: "taliban secure communications." needless to say, the student went to a teacher, things escalated, and the school ordered a full evacuation. i talked to the principal (a very, very awkward conversation) and got everything sorted out. everyone's parents are panicking, checking their kids out of school now. my personal hotspot is now called "i love school."
\\ \hline
Gold & 
turned on personal hotspot with the name "taliban secure communications" for an assembly, school got evacuated because of a supposed terror threat.
\\ \hline
Model & 
i connected my school's live camera to my personal hotspot, one of the students found out and evacuated the school.
\\ \hline
ROUGE2-F1 & 
4.88
\\ \hline
\makecell{Document\\(ID \#228)} & 
unlike the majority of these, this occurred about 20 minutes ago i am a pc gamer, which means i have a desk with tons of junk on it. sometimes i will walk in and grab a snack to eat before i start anything, which this time turned out to be a nice bag of chipsticks. i sit down and was so excited to eat them that i teared open the bag. apparently these have a very low tolerance to force, so as i rip them open the i watch each individual yellow stick of glory fly everywhere. these not only landed in all the small keyboard gaps but in every little gap imaginable. obviously before i moved i ate each individual stick to try and minimise the situation. so im now sitting here typing this up with a tiny dog vacuum trying clean up this mess.
\\ \hline
Gold & 
back of the packet doesn't state the clock inside
\\ \hline
Model & 
i ripped open a bag of chipsticks and they flew everywhere
\\ \hline
ROUGE2-F1 & 
0.00
\\ \hline
\makecell{Document\\(ID \#29)} & 
this started two days ago and it followed up today. so we had a new sales guy start at the office. i handle most of the i.t. helpdesk/network admin stuff as well as work in sales (its a private company). i was training him on our systems on his first day and it got to around noon and a little bit past so we went out to lunch. chipotle! we're talking about stuff getting to know each other. people always tend to ask questions about college, family blah blah blah. he asks me if i had any siblings, now this question is always tricky. well, my sister died 7 years ago, and i'm a pretty chill guy so i don't like to bring it up and be the debby downer so i usually do a little white lie. if i have no connection whatsoever back to my real life i say 1, if not i say 2. so i said i have 2, i **was** the middle child growing up, (haven't lied yet). hbu? he goes on saying he has an older sister.. nice dodged a bullet. so i make sure it went off on his topic so i didn't have to lie through my teeth. i hate people feeling sorry for me, compassion isn't really a trait that i have (before or after). so fast forward to today when he asks me to go to lunch again today. i try to keep my distance with people who i just meet just so i don't seem like a stage 5 clinger. so i tell him the good place around the area and we go. this is when it gets bad.. he brings back up, "so what do your sisters do" well i say my little sister is going to school ... ... ... 
\\ \hline
Gold & 
i lied to a new coworker about dead sister being alive, dads the boss.
\\ \hline
Model & 
don't lie to your boss.
\\ \hline
ROUGE2-F1 & 
0.00
\\ \hline
\end{tabular}
\caption{Generated summaries by $\text{PEGASUS}_{\text{LARGE}}$ (\gnews) on Reddit TIFU sampled by ROUGE2-F1.}
\end{table*}

\begin{table*}[bth!]
\scriptsize
\centering
\renewcommand{\arraystretch}{1.3}
\begin{tabular}{c|p{0.85\textwidth}}
\hline
\multicolumn{2}{c}{Reddit TIFU} \\ \hline
\makecell{Document\\(ID \#238)} & 
first time posting in this sub. so this happened a few minutes ago, i'm an identical twin and i was sat with the clone in the college canteen waiting for the bus to take us home when a lunch/dinner lady who was replacing the bin liners came up to us and asked if we're twins (duh). note: she doesn't know/understand english very well. her: are you twins? me: yes her: are you from saudi? me: no, africa (my mothers indian however my fathers half tanzanian half arab so to make things easy i just say i'm african since its more plausible as i have afroish hair) her: ah, you come here to study? me: yes her: i see you around college and yeah me: yeah her: are you going to university? me: yes her: how long have you been here? me: i'm in my second year her: so you stay at your cousins? (this is where i thought huh i thought she asked my ethnicity and the situation was getting really awkward) me: yes then she left and i realised i made her think i've come over to england to study and am staying at my cousins when in fact i was born in england and have stayed here my whole life... throughout the ordeal my twin was sat quiet in his phone -.- edit: sorry for the formatting, i'm posting on my phone.
\\ \hline
Gold & 
made a lunch lady believe i've immigrated to study and stay at my cousins because of how awkward the situation was when in fact i've lived in england my whole life
\\ \hline
Model & 
made a lunch lady think i was born in england and have stayed here my whole life.
\\ \hline
ROUGEL-F1 & 
44.00
\\ \hline
\makecell{Document\\(ID \#224)} & 
so this happened last night and proceeded into today. went out with this girl for drinks, we started fooling around. went back to my car and she went ape-shit on my neck. didn't notice anything at the time, woke up in the morning with a nice bruise on my neck. went to work in a very corporate setting. thought that my dress shirt buttoned all the way up would be enough to conceal it but nope. boss inquires about the bruise, i say "yeah i don't know where i got that, they keep popping up all over my body". i then proceeded to show him bruises on my arms (i was trying to climb trees whilst drunk over the weekend) as a cover up. boss became concerned and said i need to see a doctor right away as it could be something more serious like leukemia. he has given me next monday off to see the doctor...
\\ \hline
Gold & 
girl attacked my neck, boss sent me to the doctor.
\\ \hline
Model & 
went out with a girl, she ape-shit on my neck, woke up with a bruise, tried to cover it up by showing boss bruises on my arms.
\\ \hline
ROUGEL-F1 & 
21.05
\\ \hline
\makecell{Document\\(ID \#128)} & 
a few days ago me and my younger cousin (male) were playing with nerf guns shooting eachother around the house. being older and more experienced with weapons i got him good quite a few times and we were having a great time. his family had bought some knock-off nerf darts that seemed like a great deal, until they arrived and had solid tips. so instead of feeling like a mild flick these feel like a very strong flick. now so far these darts had caused no real injuries, just a bit more pain than your average nerf dart. then i foolishly decided to bring out the big guns, namely a pipe i used as a blowgun that was the perfect diameter for nerf darts. i have used blowguns for a good few years and am pretty good with them in both accuracy and power, so i knew to regulate my power when shooting at people. it was still more powerful, so in an attempt to draw my fire and catch me ammoless my cousin held a cowboy hat out from behind a doorway for me to shoot exactly like in a western. so seeing as i'm shooting an innanimate object i took a deep breath and lined up my shot. i could see my other even younger cousin (female) in the background watching me but due to my blowgun experience i assumed she'd be fine. so i let this breath go and instead of the thump of it hitting the hat my cousin crumples and runs outside crying. i follow her and she is crying and saying i shot her and her eye is blurry and its not getting better. she was being very brave given the context and i tried to console and reassure her it'd be fine so ... ... ... 
\\ \hline
Gold & 
while playing nerf with one cousin i accidentally shot another in the eye causing her iris to bleed (which if jerked too much could cause blindness) with a fake nerf dart shot powerfully from a blowgun.
\\ \hline
Model & 
shot my cousin in the eye with a blowgun, ruined their holiday, stopped them enjoying my mothers birthday party and ruined my friendship with my cousin all due to a single unlucky shot with knock-off nerf darts.
\\ \hline
ROUGEL-F1 & 
18.92
\\ \hline
\makecell{Document\\(ID \#175)} & 
regret and guilt are two of the worst things that one can feel. unlike most tifu's this one actually occurred today. me and my little brother who is 10 years old were playing football. sometimes my little brother is really annoying and this was a prime case. he decided to kick my shin, now i'm notorious for being quick to anger and can be violent, i know not one of my better traits. so i pushed him and to be honest i pushed him pretty hard. he fell and presumably stuck out his left arm to break his fall. however he instead landed on his wrist. at first it looked like he was ok but then he cried out in pain and screamed and shouted very loudly. at first i though he was bluffing because he often was melodramatic and did things like this. however after 5 minutes it became apparent that he was completely serious. i lifted him and sat him down on a soft chair. since i have done first aid training i applied ice to his wrist. then i called my mum and told her everything including that it was all my fault. i decided that lying wasn't my best option and just spilt the beans and told her everything. she was really pissed off and told us that she would be home in ten minutes. when i heard the knock on the door i knew it was not going to be good. she was really angry and shouted at me before looking at my brother. she saw his wrist and turned pale. it was black or horrible looking like something you could find on google images but it was definitely swollen. she decided to take him to a\&e, on the way out she grilled me about ... ... ... 
\\ \hline
Gold & 
i managed to accidentally break my brothers arm. my aims to buy a car before uni have faced a major setback and my hopeful future in a medical career has faced a major setback and it's all my fault.
\\ \hline
Model & 
i pushed my brother and he broke his wrist.
\\ \hline
ROUGEL-F1 & 
12.24
\\ \hline
\end{tabular}
\caption{Generated summaries by $\text{PEGASUS}_{\text{LARGE}}$ (\gnews) on Reddit TIFU sampled by ROUGEL-F1.}
\end{table*}

\begin{table*}[bth!]
\scriptsize
\centering
\renewcommand{\arraystretch}{1.3}
\begin{tabular}{c|p{0.85\textwidth}}
\hline
\multicolumn{2}{c}{BIGPATENT} \\ \hline
\makecell{Document\\(ID \#235)} & 
the left portion of fig1 and 2 shows part of a processing device in the form of a punching mechanism or installation head 2 that is used for pressing nuts m into a piece of sheet metal ( not shown ). the nuts m are fed by a sorting unit ( not shown ) through the feed device 4 by means of a flexible or solid feed line 26 designed , for example , as a hose . the punching mechanism 2 features a two - part 6a , b guide piece 6 that is flexibly braced against an upper press part ( not shown ) that moves up and down ; this press is also provided with a feed channel 8 . a sliding plunger 10 in the feed channel 8 is used to drive a nut m through the feed channel 8 during a downward movement of the upper press part and to press it into the sheet metal ( not shown ). during the downward movement of the plunger 10 , the nut m is held in position by two retaining catches 12 indicated in fig2 . furthermore , fig1 shows a sensor device 14 which determines the presence of a nut m . the remaining design of the punching device 2 is irrelevant for understanding the present application and will not be described further . the feed device 4 consists of a nozzle 20 in the form of a plate - like component which is equipped with a nozzle channel 22 going through it for the nuts m . the nozzle channel 22 features a region 24 of expanded cross section which holds one end section of the feed line 26 coming from the sorting unit . in the portion of the nozzle 20 including region ... ... ... 
\\ \hline
Gold & 
the invention describes a feed device that feeds assembly parts such as nuts , bolts , and rivets , from a sorting unit to the feed channel of a processing device for pressing or mounting of assembly parts to sheet metal . the feed device consists of a nozzle with a nozzle channel through it connected on one end to a feed line coming from the sorting unit . a protrusion of the nozzle can deploy into a recess opening of a guide piece of the processing device so that the nozzle channel will open laterally into the feed channel of the device . to connect the nozzle to the guide piece , two levers linked to the nozzle are used ; they are pretensioned by springs in a closed position surrounding the guide piece and can be moved manually , possibly with the assistance of a tool , into an open position . in this manner the feed device can be easily connected to the processing device and then detached from it .
\\ \hline
Model & 
a feed device for feeding a processing device with a feed line , particularly for feeding nuts to a punching mechanism , features a nozzle equipped with a nozzle channel that is connected to the processing device . the nozzle is equipped with a clamping plate that holds the feed line securely in the nozzle . the nozzle is releasably connected to a guide piece of the processing device by means of two levers that are pivotably mounted on the nozzle . the levers are tensioned in the closed position by springs and can be moved into a connected position by the application of pressure . the connection between the nozzle and the guide piece is released by a release mechanism when the levers are moved into the connected position .
\\ \hline
ROUGE1-F1 & 
55.94
\\ \hline
\makecell{Document\\(ID \#298)} & 
it will be readily apparent to one skilled in the art that various substitutions and modifications may be made in the invention disclosed herein without departing from the scope and spirit of the invention . the term “ a ” or “ an ” as used herein in the specification may mean one or more . as used herein in the claim ( s ), when used in conjunction with the word “ comprising ”, the words “ a ” or “ an ” may mean one or more than one . as used herein “ another ” may mean at least a second or more . the term “ animal ” as used herein refers to any species of the animal kingdom . in preferred embodiments it refers more specifically to humans , animals in their wild state , animals used as pets ( birds , dogs , cats , horses ), animals used for work ( horses , cows , dogs ) and animals which produce food ( chickens , cows , fish ), farm animals ( pigs , horses , cows , sheep , chickens ) or are themselves food ( frogs , chickens , fish , crabs , lobsters , shrimp , mussels , scallops , goats , boars , cows , lambs , pigs , ostrich , emu , eel ) and other animals well known to the art . the term “ effective amount ” as used herein is defined as the amount of the composition required to produce an effect in a host which can be monitored using several endpoints known to those skilled in the art . in a specific embodiment , these endpoints are surrogate markers . the term “ feed conversion efficiency ” as used herein is defined as the ... ... ... 
\\ \hline
Gold & 
the intramuscular electroporated injection of a protease - resistant growth hormone - releasing hormone cdna into rat dams at 16 days of gestation resulted in the enhanced long - term growth of the f1 offspring . the offspring were significantly heavier by one week of age and the difference was sustained to 10 weeks of age . consistent with their augmented growth , plasma igf - i concentration of the f1 progeny was increased significantly . the pituitary gland of the offspring was significantly heavier , and contained an increased number of somatotropes and lactotrophs , and is indicative of an alteration in cell lineages . these unique findings demonstrate that enhanced ghrh expression in pregnant dams can result in intergenerational growth promotion , by altering development of the pituitary gland in the offspring .
\\ \hline
Model & 
the present invention relates to the use of growth hormone releasing hormone and growth hormone secretagogue receptors in the manufacture of medicaments for the treatment of growth deficiencies in newborn animals .
\\ \hline
ROUGE1-F1 & 
22.22
\\ \hline
\end{tabular}
\caption{Generated summaries by $\text{PEGASUS}_{\text{LARGE}}$ (\gnews) on BIGPATENT sampled by ROUGE1-F1.}
\end{table*}
\begin{table*}[bth!]
\scriptsize
\centering
\renewcommand{\arraystretch}{1.3}
\begin{tabular}{c|p{0.85\textwidth}}
\hline
\multicolumn{2}{c}{BIGPATENT} \\ \hline
\makecell{Document\\(ID \#127)} & 
in fig1 , a container in the form of a tube 100 is provided , the tube 100 comprising a chamber 102 having a wall 104 , an upper portion 106 , and a lower portion 108 , the lower portion 106 holding a freeze - dried material comprising a biological sample ; the freeze - dried material is hereinafter referred to as a cake 110 . typically , the cake 110 is in a disc - shaped form . the tube 100 comprises a physical structure 112 to inhibit movement of the freeze - dried biological sample from the lower portion 108 of the chamber 102 to the upper portion 106 of the chamber 102 . the tube 100 comprises a first end 114 , which may initially be open in order to enable insertion of the material prior to freeze - drying , and subsequently closed with a seal ( not shown ). the external dimensions of the tube 100 may be fixed at an industry standard for ease of handling , for example , automated handling by existing laboratory equipment . typically , the tube 100 is 8 mm in diameter and 18 . 3 mm long . typically , the internal dimensions of the tube are large enough in both diameter and depth to accommodate standard dispensing tools , for example pipetting needles in automated dispensing systems ; this has the advantage that there is no need for readjustment or calibration of existing tools . the biological sample may comprise an enzyme , such as a polymerase , reverse transcriptase or any other enzyme , blood , tissue , serum or any other biological substance . in fig2 a and 2 b , a tube 100 a is provided in which a physical structure 112 in ... ... ... 
\\ \hline
Gold & 
embodiments of the invention relate to a container holding , and a method of storing , freeze - dried biological samples . in particular , there is provided a container holding a freeze - dried material comprising a biological sample , the container comprising a chamber having an upper portion and a lower portion , the chamber comprising a wall and the lower portion being fluidly connected to the upper portion such that , when liquid is received at the upper portion , the received liquid can pass to and accumulate in the lower portion . further , the freeze - dried material is located in the lower portion , and the container comprises a physical structure in the form of a stop protruding inwards from the wall , the physical structure being for inhibiting the freeze - dried material from moving from the lower portion of the chamber to the upper portion of the chamber . this ensures that the biological sample can be kept in a lower portion of the tube , ensuring that , when the biological sample is subsequently reconstituted by inserting water , substantially all of the biological sample is dissolved .
\\ \hline
Model & 
a container for containing a biological sample , the container comprising : a chamber having a wall , an upper portion , and a lower portion , the lower portion holding a freeze - dried material comprising a biological sample ; and a physical structure to inhibit movement of the freeze - dried biological sample from the lower portion of the chamber to the upper portion of the chamber . the invention also relates to methods of preparing a biological sample , and to systems and methods for dispensing a biological sample .
\\ \hline
ROUGE2-F1 & 
38.28
\\ \hline
\makecell{Document\\(ID \#155)} & 
the automated edge finishing apparatus and method of the present invention comprises hardware components and associated software , providing the ability to learn a trajectory of a workpiece by \&\# 34 ; guarded moves ,\&\# 34 ; machining the workpiece by moving the workpiece along the trajectory and controlling the trajectory in real time . similar workpieces can be machined from the learned trajectory without expensive fixtures by the sensors and software locating the tool and adjusting the trajectory . the invention is useful for performing edge finishing on intricately designed workpieces or workpieces with close edge finishing tolerances . the configuration of the preferred system is shown in fig1 and 2 . the preferred invention comprises an xy table 10 , a force sensor 12 , cutting tool 14 , xy table position control hardware 32 , and a control system as shown in fig2 . the preferred xy table 10 comprises a daedal series 200000 , or the like , powered by tlm120 linear motors from trilogy systems corporation , or the like . the linear actuators comprise three - phase brushless linear motors , with a stationary magnet track and a moving coil assembly . the linear motors are driven by dc amplifiers , such as pwm amplifiers . both table axes are equipped with incremental encoders with a resolution of 12 , 700 counts / inch . the innermost level of table control is position control , by a controller 32 such as a dmc - 520 motion controller from galil motion control , as shown in fig2 . this board controls position of both table axes using proportional plus derivative ( pd ) control action . the resulting position - controlled table is the starting point for the force control system discussed below . a ... ... ... 
\\ \hline
Gold & 
the disclosure is directed to an apparatus and method for automated edge finishing using hybrid position / force control of an xy table . the disclosure is particularly directed to learning the trajectory of the edge of a workpiece by \&\# 34 ; guarded moves \&\# 34 ;. machining is done by controllably moving the xy table , with the workpiece mounted thereon , along the learned trajectory with feedback from a force sensor . other similar workpieces can be mounted , without a fixture on the xy table , located and the learned trajectory adjusted
\\ \hline
Model & 
an apparatus and method for edge finishing comprising a position - controlled linear table , a cutting tool , a force sensor , and a control system . the position - controlled table allows the workpiece to be moved past the cutting tool . the table is equipped with two linear motors and associated encoders for position control . the control system can be programmed to allow the workpiece to be cut from a learned trajectory in real time . the cutting tool can be programmed to cut the workpiece along the trajectory , or it can be programmed to cut the workpiece on a normal path . the system combines the two major elements of edge finishing : position control and force control . the position control allows the workpiece to be moved past the cutting tool , while the force control allows the workpiece to be moved in the normal direction . the system components can be placed in modular form , allowing rapid upgrade and replacement of existing systems .
\\ \hline
ROUGE2-F1 & 
11.72
\\ \hline
\end{tabular}
\caption{Generated summaries by $\text{PEGASUS}_{\text{LARGE}}$ (\gnews) on BIGPATENT sampled by ROUGE2-F1.}
\end{table*}

\begin{table*}[bth!]
\scriptsize
\centering
\renewcommand{\arraystretch}{1.3}
\begin{tabular}{c|p{0.85\textwidth}}
\hline
\multicolumn{2}{c}{BIGPATENT} \\ \hline
\makecell{Document\\(ID \#185)} & 
referring to fig3 , one aspect of the disclosed system for connecting a sensor to a controller , generally designated 100 , may include a sensor 102 , an electronic controller 104 and a power source 106 , such as a battery ( e . g ., a 12 v automotive battery ). the system 100 may be connected to ground 108 , such as a vehicle chassis . the wiring inductance l w , wiring resistance r w and current noise source n of the system 100 may represent ground noise created by transient currents in the ground path of the controller 104 . in one aspect , sensor 102 may be a pedal feel emulator ( not shown ) that indicates a driver \&\# 39 ; s brake request and the controller 104 may be associated with a front right electric caliper ( not shown ) and may generate and communicate a braking signal to the caliper based upon signals received from the pedal feel emulator . the controller 104 may include resistors r 10 , r 11 , r 12 and capacitors c 6 , c 7 . the input to the controller 104 from the sensor 102 may be in the form of a single wire 110 that supplies a current . for example , a single pin connector may be used to connect the sensor 102 to the controller 104 . the use of a single wire connection between the sensor 102 and the controller 104 may provide several advantages , including reduced costs and manufacturing time . the current supplied by the wire 110 may be converted to a signal voltage by resistor r 11 , which may be filtered by a low pass filter 112 created by resistors r 10 , r 12 ... ... ... 
\\ \hline
Gold & 
a sensor to controller connection system including a power source , a controller in communication with the power source , and a sensor in communication with the power source and the controller , the sensor including sensor electronics and a current source , the current source having a control input and an output , the control input being applied by the sensor electronics and the output being applied to the controller , wherein the current source controls an electric signal communicated to the controller from the sensor based upon the control input .
\\ \hline
Model & 
a system including a sensor having a first terminal for receiving a first voltage and a second terminal for receiving a second voltage , the second voltage being a voltage greater than the first voltage ; a controller having a first terminal for receiving the first voltage and a second terminal for receiving the second voltage ; and a current source connected between the second terminal of the sensor and the first terminal of the controller .
\\ \hline
ROUGEL-F1 & 
28.93
\\ \hline
\makecell{Document\\(ID \#237)} & 
referring now to fig2 - 4 a preferred embodiment of the control system 10 of the present invention is shown mounted upon the underside of gun 11 and comprised of an emd 12 , electronic circuitry denoted schematically by box 13 , a battery 14 , and electrical switch trigger 15 . the illustrated gun is comprised of a forestock 22 which supports barrel 16 , receiver portion 17 located at the rear extremity of said barrel , magazine 18 , conventional trigger 19 with associated pistol grip 20 , and shoulder stock 21 which contains a compressed propellant gas . the emd is intended to produce reciprocating linear movement of a push rod 23 . the emd may be a solenoid , either of an in - line type or clapper or rotary type . alternatively , the emd may be a servo type device using an arm , lever or gear system to activate rod 23 . all such devices are characterized in that a pulse of electrical energy produces a controlled mechanical force , and the discontinuation of said pulse either produces a reverse force or permits interaction of a reverse force such as may be produced by a spring - biased conventional trigger . the emd may be secured to forestock 22 by brackets or removable fastening means . a push rod 23 , extending from said emd is slideably positioned by guide 25 mounted by bracket 35 beneath receiver portion 17 . in those embodiments wherein the emd is a solenoid , guide 25 may not be required . the length of rod 23 and its positioning by guide 25 is such as to cause the distal extremity 26 of rod 23 to contact trigger 19 . said distal extremity may be equipped with coupling ... ... ... 
\\ \hline
Gold & 
a system for controlling the firing of a paint ball gun includes an electromechanical device that produces linear motion interactive with the conventional trigger of the gun . electronic circuitry produces pulses that activate the emd . the pulses are based upon control signals produced by either the timing of paint balls entering the firing chamber of the gun or the detection of the presence of a single paint ball within the chamber . a manually operated electrical switch trigger activates the electronic circuitry . an adjustably predetermined number of paint balls will fire based upon each depression of the electrical switch trigger while minimizing the chopping of paint balls in the firing chamber .
\\ \hline
Model & 
a paint ball gun control system which permits selective firing of paint balls in response to the depression of the trigger . the system includes an electronic circuit which converts a dc current into a regulated pulse current , and an electromagnetic device which is actuated by the regulated pulse current to produce a reciprocating linear motion . a push rod attached to the emd is adapted to contact the trigger of the gun .
\\ \hline
ROUGEL-F1 & 
22.10
\\ \hline
\end{tabular}
\caption{Generated summaries by $\text{PEGASUS}_{\text{LARGE}}$ (\gnews) on BIGPATENT sampled by ROUGEL-F1.}
\end{table*}

\begin{table*}[bth!]
\scriptsize
\centering
\renewcommand{\arraystretch}{1.3}
\begin{tabular}{c|p{0.85\textwidth}}
\hline
\multicolumn{2}{c}{arXiv} \\ \hline
\makecell{Document\\(ID \#34)} & 
consider a set of objects which should be ranked on the basis of information about their bilateral relationships . similar problems arise , among others , in social choice theory \@xcite , sports \@xcite , psychology \@xcite , internet search \@xcite , and bibliometrics \@xcite . we discuss a universal version of the problem involving arbitrary preference intensities as well as incomplete and multiple comparisons . the main contribution of this paper is the presentation of an impossibility theorem : consistency requiring that if an object is ranked at least as high as another in two independent problems , then it is ranked as high as the other in the unified problem , too and self - consistency a less known but intuitive property , introduced in \@xcite , which prohibits to assign a lower rank for an object with a better or equivalent performance than another can not be met simultaneously by any ranking method on the set of all problems . domain restrictions and weakening of the properties are also investigated in order to get some positive results . since self - consistency is an axiom difficult to debate , we can not demand consistency from a ranking method . at the first glance , it is a somewhat surprising and controversial result making the aggregation of ( incomplete ) preferences impossible . one may argue that the whole concept of paired comparisons - based ranking is flawed . however , loss of consistency may be regarded as a necessary sacrifice for dimension reduction since the outcome should be an inherently transitive ranking even from intransitive data . this interpretation is reinforced by the connection between consistency and independence of irrelevant matches ( which requires the relative ranking of two objects to be independent of the outcome of ... ... ... 
\\ \hline
Gold & 
assume a set of objects is given with information about their bilateral relationships , allowing for incomplete and multiple comparisons as well as different preference intensities . an axiomatic approach is applied for the problem of ranking the objects . consistency requires the preservation of relative ranking if two sets of such preferences are aggregated . self - consistency assigns the same rank for objects with the same performance , furthermore , an object should be ranked strictly higher if it shows an obviously better performance than another . it is revealed that these two properties can not be satisfied simultaneously . the impossibility holds under various restrictions on the domain . however , a positive result emerges if only the aggregation of ranking problems with the same comparison structure is allowed , for example , we have two round - robin tournaments .
\\ \hline
Model & 
consider a set of objects which should be ranked on the basis of information about their bilateral relationships . similar problems arise , among others , in social choice theory , sports , psychology , internet search , and bibliometrics . we discuss a universal version of the problem involving arbitrary preference intensities as well as incomplete and multiple comparisons . the main contribution of this paper is the presentation of an impossibility theorem : consistency requiring that if an object is ranked at least as high as another in two independent problems , then it is ranked as high as the other in the unified problem , too and self - consistency a less known but intuitive property , introduced in \@xcite , which prohibits to assign a lower rank for an object with a better or equivalent performance than another can not be met simultaneously by any ranking method on the set of all problems . domain restrictions and weakening of the properties are also investigated in order to get some positive results .
\\ \hline
ROUGE1-F1 & 
48.61
\\ \hline
\makecell{Document\\(ID \#289)} & 
machine learning methods are used widely within high energy physics ( hep ) . one promising approach , used extensively outside of hep for applications such as handwriting recognition , is that of support vector machines ( svms ) , a supervised learning model used with associated learning algorithms for multivariate analysis ( mva ) . developed originally in the 1960s , with the current standard version proposed in 1995 \@xcite , svms aim to classify data points using a maximal margin hyperplane mapped from a linear classification problem to a possibly infinite dimensional hyperspace . however this means svms , like other mva classifiers , have a number of free parameters which need to be tuned on a case by case basis . this motivates a number methods for ensuring the classifier is sufficiently generalised such that when used on an unseen dataset the performance can be accurately predicted . in this paper a brief overview of svms is given in section [ sec : svm ] , with an example using svms shown in section [ sec : checker ] . generalisation is discussed in section [ sec : gen ] with an illustrative example of how this can improve performance given in section [ sec : check2 ] . consider the problem of linear classification with the svm where the training set , \@xmath0 , is linearly separable . we define a separating hyperplane given by \@xmath1 , where \@xmath2 , the weight vector , is perpendicular to the hyperplane , and \@xmath3 , the bias , determines the distance of the hyperplane from the origin ( fig . ) . a decision function defined by \@xmath4 is used to positively and negatively classify \@xmath5 , the points in the training set . without further constraint the ... ... ... 
\\ \hline
Gold & 
we review the concept of support vector machines ( svms ) and discuss examples of their use . one of the benefits of svm algorithms , compared with neural networks and decision trees is that they can be less susceptible to over fitting than those other algorithms are to over training . this issue is related to the generalisation of a multivariate algorithm ( mva ) ; a problem that has often been overlooked in particle physics . we discuss cross validation and how this can be used to improve the generalisation of a mva in the context of high energy physics analyses . the examples presented use the toolkit for multivariate analysis ( tmva ) based on root and describe our improvements to the svm functionality and new tools introduced for cross validation within this framework .
\\ \hline
Model & 
support vector machines ( svms ) are one of the most promising methods for machine learning in high energy physics . svms aim to classify data points using a maximal margin hyperplane mapped from a linear classification problem to a possibly infinite dimensional hyperspace . however this means svms , like other mva classifiers , have a number of free parameters which need to be tuned on a case by case basis . this motivates a number methods for ensuring the classifier is sufficiently generalised such that when used on an unseen dataset the performance can be accurately predicted . in this paper a brief overview of svms is given , with an example using svms shown in section [ sec : checker ] . generalisation is discussed with an illustrative example of how this can improve performance given in section [ sec : gen ] .
\\ \hline
ROUGE1-F1 & 
35.29
\\ \hline
\end{tabular}
\caption{Generated summaries by $\text{PEGASUS}_{\text{LARGE}}$ (\gnews) on arXiv sampled by ROUGE1-F1.}
\end{table*}

\begin{table*}[bth!]
\scriptsize
\centering
\renewcommand{\arraystretch}{1.3}
\begin{tabular}{c|p{0.85\textwidth}}
\hline
\multicolumn{2}{c}{arXiv} \\ \hline
\makecell{Document\\(ID \#294)} & 
the pure quantum states are traditionally associated with the wave function \@xcite or a vector in the hilbert space \@xcite . the mixed quantum states are described by the density matrix \@xcite or the density operator \@xcite . there exist several representations of quantum states in terms of the quasidistribution functions like the wigner function \@xcite and the husimi kano function \@xcite . the diagonal representation of quantum states was suggested in \@xcite ( see also \@xcite ) . it was studied and applied in \@xcite . in this representation , a quantum state is represented in terms of weighted sum of coherent - state \@xmath0 projectors . the properties of all the quantum - state representations considered are associated with the properties of the density operator which is hermitian , trace - class nonnegative operator . this means , in particular , that all the eigenvalues of the density operators must be nonnegative . in the quantum domain , the multipartite systems have a specific property connected with strong correlations of the quantum subsystems . this property provides the entanglement phenomenon \@xcite . in the diagonal representation of the density states , the weight function \@xmath1 is an analog of the probability - distribution function in the phase space . for some class of states , this function is identical to the probability - distribution function like in classical statistical mechanics . in \@xcite , the tomographic - probability representation of quantum states , where the quantum state is associated with the so - called symplectic tomogram , was introduced . the tomogram is a fair probability distribution containing the same information on quantum state that the density operator does ( or such its characteristics as the wigner or husimi kano functions ) . the aim of this work ... ... ... 
\\ \hline
Gold & 
the quasidistributions corresponding to the diagonal representation of quantum states are discussed within the framework of operator - symbol construction . the tomographic - probability distribution describing the quantum state in the probability representation of quantum mechanics is reviewed . the connection of the diagonal and probability representations is discussed . the superposition rule is considered in terms of the density - operator symbols . the separability and entanglement properties of multipartite quantum systems are formulated as the properties of the density - operator symbols of the system states .
\\ \hline
Model & 
the diagonal and tomographic representations of quantum states are considered . the quantum states are represented in terms of the weighted sum of coherent - state \@xmath0 projectors . in the tomographic - probability representation , the quantum state is associated with the so - called symplectic tomogram . the tomographic - probability distribution is a fair probability distribution containing the same information on quantum state that the density operator does ( or such its characteristics as the wigner or husimi kano functions ) . the aim of this work is to find the explicit formulae realizing the connection of the diagonal and tomographic probability representations . the functions like the wigner function , husimi kano function and tomographic - probability - distribution function are considered as symbols of the density operators of a corresponding star - product scheme . the superposition rule given in terms of tomograms and in terms of weight functions of the diagonal representation where explicit kernels of the corresponding star - products are employed to obtain the addition rules for the tomograms and weight functions are considered . we discuss also the formulation of the separability and entanglement properties of composed system in the tomographic probability and diagonal representations .
\\ \hline
ROUGE2-F1 & 
26.52
\\ \hline
\makecell{Document\\(ID \#32)} & 
deep neural networks ( dnns ) have been receiving ubiquitous success in wide applications , ranging from computer vision \@xcite , to speech recognition \@xcite , natural language processing \@xcite , and domain adaptation \@xcite . as the sizes of data mount up , people usually have to increase the number of parameters in dnns so as to absorb the vast volume of supervision . high performance computing techniques are investigated to speed up dnn training , concerning optimization algorithms , parallel synchronisations on clusters w / o gpus , and stochastic binarization / ternarization , etc \@xcite . on the other hand the memory and energy consumption is usually , if not always , constrained in industrial applications \@xcite . for instance , for commercial search engines ( e.g. , google and baidu ) and recommendation systems ( e.g. , netflix and youtube ) , the ratio between the increased model size and the improved performance should be considered given limited online resources . compressing the model size becomes more important for applications on mobile and embedded devices \@xcite . having dnns running on mobile apps owns many great features such as better privacy , less network bandwidth and real time processing . however , the energy consumption of battery - constrained mobile devices is usually dominated by memory access , which would be greatly saved if a dnn model can fit in on - chip storage rather than dram storage ( c.f . \@xcite for details ) . a recent trend of studies are thus motivated to focus on compressing the size of dnns while mostly keeping their predictive performance \@xcite . with different intuitions , there are mainly two types of dnn compression methods , which could be used in conjunction for better parameter savings . the ... ... ... 
\\ \hline
Gold & 
as the complexity of deep neural networks ( dnns ) trend to grow to absorb the increasing sizes of data , memory and energy consumption has been receiving more and more attentions for industrial applications , especially on mobile devices . this paper presents a novel structure based on functional hashing to compress dnns , namely funhashnn . for each entry in a deep net , funhashnn uses multiple low - cost hash functions to fetch values in the compression space , and then employs a small reconstruction network to recover that entry . the reconstruction network is plugged into the whole network and trained jointly . funhashnn includes the recently proposed hashednets \@xcite as a degenerated case , and benefits from larger value capacity and less reconstruction loss . we further discuss extensions with dual space hashing and multi - hops . on several benchmark datasets , funhashnn demonstrates high compression ratios with little loss on prediction accuracy .
\\ \hline
Model & 
deep neural networks ( dnns ) have been receiving ubiquitous success in wide applications , ranging from computer vision to speech recognition . as the sizes of data mount up , people usually have to increase the number of parameters in dnns so as to absorb the vast volume of supervision . compressing the model size becomes more important for applications on mobile and embedded devices . with different intuitions , there are mainly two types of dnn compression methods , which could be used in conjunction for better parameter savings . the first type tries to revise the training target into more informative supervision using dark knowledge . the second type observes the redundancy existence in network weights , and exploits techniques to constrain or reduce the number of free - parameters in dnns during learning . in applications , we observe hashednets compresses model sizes greatly at marginal loss of accuracy for some situations , whereas also significantly loses accuracy for others . after revisiting its mechanism , we conjecture this instability comes from at least three factors . first , hashing and training are disjoint in a two - phase manner , i.e. , once inappropriate collisions exist , there may be no much optimization room left for training . second , one single hash
\\ \hline
ROUGE2-F1 & 
7.21
\\ \hline
\end{tabular}
\caption{Generated summaries by $\text{PEGASUS}_{\text{LARGE}}$ (\gnews) on arXiv sampled by ROUGE2-F1.}
\end{table*}

\begin{table*}[bth!]
\scriptsize
\centering
\renewcommand{\arraystretch}{1.3}
\begin{tabular}{c|p{0.85\textwidth}}
\hline
\multicolumn{2}{c}{arXiv} \\ \hline
\makecell{Document\\(ID \#248)} & 
stripped supernovae ( sne ) and long - duration gamma - ray bursts ( long grbs ) are nature s most powerful explosions from massive stars . they energize and enrich the interstellar medium , and , like beacons , they are visible over large cosmological distances . however , the mass and metallicity range of their progenitors is not known , nor the detailed physics of the explosion ( see reviews by \@xcite and \@xcite ) . stripped - envelope sne ( i.e , sne of types iib , ib , and ic , e.g. , \@xcite ) are core - collapse events whose massive progenitors have been stripped of progressively larger amounts of their outermost h and he envelopes ( fig . [ fig1 ] ) . in particular , broad - lined sne ic ( sne ic - bl ) are sne ic whose line widths approach 20,000\@xmath030,000 \@xmath1 around maximum light ( see below ) and whose optical spectra show no trace of h and he . for the last 15 years , the exciting connection between long grbs and sne ic - bl , the only type of sne observed accompanying long grbs ( for reviews , see \@xcite ) , and the existence of many more sne ic - bl \_ without \_ grbs raises the question of what distinguishes sn - grb progenitors from those of ordinary sne ic - bl without grbs . viewing angle effects are probably not the reason why those sne ic - bl did not show an accompanied grbs \@xcite and based the same radio upper - limits , only \@xmath2 1\% of sne ib / c appear to be accompanied by grbs \@xcite . one promising line of attack is to investigate what sets apart sne ib ... ... ... 
\\ \hline
Gold & 
while the connection between long gamma - ray bursts ( grbs ) and type ib / c supernovae ( sne ib / c ) from stripped stars has been well - established , one key outstanding question is what conditions and factors lead to each kind of explosion in massive stripped stars . one promising line of attack is to investigate what sets apart sne ib / c * with * grbs from those * without * grbs . here , i briefly present two observational studies that probe the sn properties and the environmental metallicities of sne ib / c ( specifically broad - lined sne ic ) with and without grbs . i present an analysis of expansion velocities based on published spectra and on the homogeneous spectroscopic cfa data set of over 70 sne of types iib , ib , ic and ic - bl , which triples the world supply of well - observed stripped sne . moreover , i demonstrate that a meta - analysis of the three published sn ib / c metallicity data sets , when including only values at the sn positions to probe natal oxygen abundances , indicates at very high significance that indeed sne ic erupt
\\ \hline
Model & 
stripped - envelope supernovae ( sne ) and long - duration gamma - ray bursts ( long grbs ) are nature s most powerful explosions from massive stars . however , the mass and metallicity range of their progenitors is not known , nor the physics of the explosion . in particular , broad - lined sne ic ( sne ic - bl ) are sne ic whose line widths approach 20,000\@xmath030,000 \@xmath1 around maximum light and whose optical spectra show no trace of h and he . for the last 15 years , the exciting connection between long grbs and sne ic - bl , the only type of sne observed accompanying long grbs , raises the question of what distinguishes sn - grb progenitors from those of ordinary sne ic without grbs . one promising line of attack is to investigate what sets apart sne ib / c * with * grbs from those * without * grbs to elucidate the conditions and progenitors of these two types of explosions .
\\ \hline
ROUGEL-F1 & 
22.50
\\ \hline
\makecell{Document\\(ID \#270)} & 
the alice experiment is dedicated to the study of the properties of qcd matter created in nucleus - nucleus collisions at lhc energies \@xcite . the inner tracking system in the alice apparatus is made of position sensitive detectors which have to handle several thousands tracks per unit of rapidity . the two innermost layers at 3.9 \@xmath10 and 7.6 \@xmath10 radii , respectively , constitute the silicon pixel detector ( spd ) . the spatial precision and hit efficiency of the spd are key parameters since they determine the alice capability of detecting particles with open heavy - flavour \@xcite . + the basic detector unit of the alice spd is the ladder , a two - dimensional silicon matrix of p\@xmath11n reverse biased diodes of dimensions 50 x 425 \@xmath12 , flip - chip bonded to five read - out chips . each diode is connected to a cell of the front - end read - out asic via a pb - sn solder bump of 25 \@xmath13 diameter . the detector contains nearly 10\@xmath14 active cells in total . the read - out is binary . to reduce the material budget , the sensor thickness is limited to 200 \@xmath13 and the read - out chip wafers are thinned down to 150 \@xmath13 . further details can be found in \@xcite . + early prototypes of the alice spd elements , in the form of single - chip assemblies , were tested in high energy proton / pion beams at the cern sps in 2002 and 2003 . these assemblies were made with sensors of 200 \@xmath13 and 300 \@xmath13 thicknesses , while the read - out chips ( unthinned ) were 725 \@xmath13 thick . those beam tests were primarily aimed at evaluating the performance of ... ... ... 
\\ \hline
Gold & 
the two innermost layers of the alice inner tracking system are instrumented with silicon pixel detectors . single chip assembly prototypes of the alice pixels have been tested in high energy particle beams at the cern sps . detection efficiency and spatial precision have been studied as a function of the threshold and the track incidence angle . the experimental method , data analysis and main results are presented . d. elia\@xmath0 , g. anelli\@xmath1 , f. antinori\@xmath2 , a. badal\@xmath3 , g.e . bruno\@xmath4 , m. burns\@xmath1 , i.a . cali\@xmath5 , m. campbell\@xmath1 , m. caselle\@xmath4 , s. ceresa\@xmath1 , p. chochula\@xmath1 , m. cinausero\@xmath6 , j. conrad\@xmath1 , r. dima\@xmath2 , d. fabris\@xmath2 , r.a . fini\@xmath4 , e. fioretto\@x
\\ \hline
Model & 
the spatial precision and hit efficiency of the alice silicon pixel detector ( spd ) are key parameters for the detection of heavy - flavour in nucleus - nucleus collisions at lhc energies . the performance of the detector under test , in the form of single - chip assemblies , together with a detailed cluster analysis of the hits , can yield a good determination of the intrinsic spatial precision and detection efficiency of the pixel plane under test . results from the 2002 beam test at the cern sps , where the sensor thickness was the same as the one used in ladder production , are presented and compared with the main results for the thicker sensor . , , , , , , , ,
\\ \hline
ROUGEL-F1 & 
17.70
\\ \hline
\end{tabular}
\caption{Generated summaries by $\text{PEGASUS}_{\text{LARGE}}$ (\gnews) on arXiv sampled by ROUGEL-F1.}
\end{table*}

\begin{table*}[bth!]
\scriptsize
\centering
\renewcommand{\arraystretch}{1.3}
\begin{tabular}{c|p{0.85\textwidth}}
\hline
\multicolumn{2}{c}{PubMed} \\ \hline
\makecell{Document\\(ID \#80)} & 
aging is a gradual process , proportional to time , that causes structural and functional changes due to internal degeneration . aging can be divided into intrinsic aging , which is natural with the progression of time , and environmental aging caused by the external environment . uv - induced photoaging is usually observed on skin that is exposed to the sun . its clinical characteristics include earlier observation than intrinsic aging , deep and wide wrinkles , inconsistent pigmentation and solar lentigo on the exposed skin surface2 . along with recent socio - financial improvements , improvements in medical technology , hygiene and nutrition have increased life expectancy . as the elderly population increases , many treatment methods to reduce skin aging and improve wrinkles and pores are currently being researched4 , 5 . however , ablative laser treatment causes a financial burden and side effects , such as erythema , edema , and hyperpigmentation . in addition , the damage on the epidermis caused by ablative laser treatment delays the time between returning to daily activities6,7,8 . recently , non - ablative skin rejuvenation , which improves skin aging and photoaging without damaging the epidermis , has been the preferred treatment9 . previous studies suggest that radiofrequency ( rf ) effectively reduces abdominal obesity by transferring heat to the dermis , increases blood volume in the capillaries and hence increases blood flow , induces collagen formation , and increases the thickness of the epidermis10,11,12 . electroacupunture ( ea ) has been reported to reduce inflammation , induce healing of damaged tissues , reduce pain , and improve facial rejuvenation13,14,15 . low - level laser therapy ( lllt ) has been used in various research studies and has been reported to have anti - inflammatory effects , reduce pain ... ... ... 
\\ \hline
Gold & 
[ purpose ] the purpose of this study was to investigate the effects of radiofrequency ( rf ) , electroacupuncture ( ea ) , and low - level laser therapy ( lllt ) , which are used in physical therapies , on facial wrinkles and moisture . [ subjects and methods ] a total of 30 female participants aged 3055 years participated in this study and the results will be used as a reference for further development of skin physical therapy . thirty adult females were assigned to an lllt ( n=10 ) , ea ( n=10 ) , or rf group ( n=10 ) . the intervention was performed in two 15-minute sessions per week for six weeks . subjects skin tone and pigmentation were observed before and after the intervention . [ results ] treatment of the under eye area showed that wrinkles were significantly decreased on both sides after rf , ea , and lllt . treatment of the eye rims indicated that wrinkles significantly decreased on the right side after rf , ea , and lllt . [ conclusion ] the application of lllt , ea , and rf had positive effects on wrinkle and moisture content of adult women s faces .
\\ \hline
Model & 
[ purpose ] the purpose of this study was to investigate the effects of radiofrequency ( rf ) , electroacupuncture ( ea ) , and low - level laser therapy ( lllt ) on wrinkles and moisture content . [ subjects and methods ] the subjects of this study were 30 adult women aged between 30 and 55 years who presented at m dermatology hospital in yongin , gyeonggi - do , south korea and who voluntarily agreed to participate after listening to an explanation of the purpose of the study . the subjects were assigned to rf , ( n=10 ) , ea ( n=10 ) , or lllt ( n=10 ) groups . the subjects received rf , ea , and lllt treatments for six weeks . [ results ] the rf , ea , and lllt groups showed significant differences in wrinkles and moisture content . [ conclusion ] the results of this study suggest that rf , ea , and lllt may be effective for facial rejuvenation .
\\ \hline
ROUGE1-F1 & 
62.50
\\ \hline
\makecell{Document\\(ID \#124)} & 
acute appendicitis is considered as the most frequent cause of acute abdomen in young patients with a slight predominance in males . although its clinical diagnosis is simple , 20 - 30\% of appendicitis cases are reported to be gangrenous or perforated because of delayed diagnosis , which can lead to increased morbidity and mortality . additionally , up to 10\% of patients especially older ones , immunocompromised patients , and pregnant women are diagnosed with atypical presentations such as appendiceal mass which requires long hospital admissions , intravenous antibiotics , and often needs percutaneous drainage . using colonoscopy for the diagnosis and treatment of asymptomatic acute appendicitis and appendiceal abscess are rare . in unusual appendicitis cases , especially in the presence of abdominal comorbidities , colonoscopy is used ; although colonoscopy is not considered as the first line intervention because of complications risk such as bleeding or perforation . furthermore non - invasive tests such as ultrasonography and computed tomography with more sensitivity can be used . we report an atypical case of appendiceal abscess , which was diagnosed during colonoscopy . a 30-year - old man referred for evaluation to our center with epigastric pain from three days prior to his evaluation . the pain was nonspecific and colicky distributed in periumblical region , did not radiate to any site and was unrelated to bowel movements . he had experienced also nausea , anorexia and intermittent vomiting . there was a history of marijuana and alcohol usage from several years ago . upon physical examination , he was ill with restlessness . other laboratory values such as hemoglobin , platelets , serum amylase , erythrocyte sedimentation rate , aspartate aminotransferase , alanine aminotransferase , alkaline phosphatase , and electrolytes were normal . his workup on admission including ... ... ... 
\\ \hline
Gold & 
although complicated acute appendicitis is occasionally observed in general practice , the diagnosis of appendiceal abscess by colonoscopy is unusual . we report a case of appendiceal abscess with spontaneous drainage into the lumen of colon diagnosed during colonoscopy and resolved by the time of surgery .
\\ \hline
Model & 
acute appendicitis is considered as the most frequent cause of acute abdomen in young patients with a slight predominance in males . although its clinical diagnosis is simple , 20 - 30\% of appendicitis cases are reported to be gangrenous or perforated because of delayed diagnosis , which can lead to increased morbidity and mortality . additionally , up to 10\% of patients especially older ones , immunocompromised patients , and pregnant women are diagnosed with atypical presentations such as appendiceal mass which requires long hospital admissions , intravenous antibiotics , and often needs percutaneous drainage . using colonoscopy for the diagnosis and treatment of asymptomatic acute appendicitis and appendiceal abscess are rare . in unusual appendicitis cases , especially in the presence of abdominal comorbidities , colonoscopy is used ; although colonoscopy is not considered as the first line intervention because of complications risk such as bleeding or perforation . we report an atypical case of appendiceal abscess , which was diagnosed during colonoscopy .
\\ \hline
ROUGE1-F1 & 
31.25
\\ \hline
\end{tabular}
\caption{Generated summaries by $\text{PEGASUS}_{\text{LARGE}}$ (\gnews) on PubMed sampled by ROUGE1-F1.}
\end{table*}

\begin{table*}[bth!]
\scriptsize
\centering
\renewcommand{\arraystretch}{1.3}
\begin{tabular}{c|p{0.85\textwidth}}
\hline
\multicolumn{2}{c}{PubMed} \\ \hline
\makecell{Document\\(ID \#299)} & 
a positive family history of prostate cancer ( pca ) is an established risk factor for pca . first - degree relatives of affected men have a 23 fold increased risk of pca . when 3 or more first - degree relatives are affected ( or at least 2 first - degree relatives before the age of 55 years ) , the family is considered a hereditary prostate cancer ( hpc ) family according to the so - called johns hopkins or carter criteria . yet , only a few very rare high - penetrance gene mutations have been identified to cause hpc . in recent years , genome - wide association studies have added approximately 40 low - penetrance genetic polymorphisms that are associated with an increased risk of pca [ 4 , 5 ] . several polymorphisms have also been identified that are associated with serum prostate - specific antigen ( psa ) , the most commonly used marker for early detection of pca [ 6 , 7 ] . an ongoing matter of debate is whether psa testing should be used for population - wide screening . in a population - based setting , the european randomized study of screening for prostate cancer ( erspc ) showed a decrease in pca mortality of 31\% in the screening arm after correction for non - attendance and contamination . by contrast , the prostate , lung , colorectal and ovarian cancer screening trial ( plco ) , found no effect of psa screening on mortality , but suffered from methodological problems which severely hamper interpretation of the results . previous studies into the effectiveness of psa sceening in men with an increased risk due to family history have yielded largely inconsistent results . these range from a marked benefit ... ... ... 
\\ \hline
Gold & 
a family history of prostate cancer ( pca ) is an established risk factor for pca . in case of a positive family history , the balance between positive and adverse effects of prostate - specific antigen ( psa ) testing might be different from the general population , for which the european randomized study of screening for prostate cancer ( erspc ) showed a beneficial effect on mortality . this , however , went at the cost of considerable overtreatment . this study assessed dutch physicians knowledge of heredity and pca and their post - erspc attitude towards pca testing , including consideration of family history . in january 2010 , all dutch urologists and clinical geneticists ( cgs ) and 300 general practitioners ( gps ) were invited by email to complete an anonymous online survey , which contained questions about hereditary pca and their attitudes towards pca case - finding and screening . 109 urologists ( 31\% ) , 69 gps ( 23\% ) and 46 cgs ( 31\% ) completed the survey . cgs had the most accurate knowledge of hereditary pca . all but 1 cg mentioned at least one inherited trait with pca , compared to only 25\% of urologists and 9\% of gps
\\ \hline
Model & 
purposea positive family history of prostate cancer ( pca ) is an established risk factor for pca . to guide the public and physicians in translating the results of the european randomized study of screening for prostate cancer ( erspc ) and the prostate , lung , colorectal and ovarian cancer screening trial ( plco ) into clinical practice , the dutch association of urology ( nvu ) and the dutch college of general practitioners ( nhg ) concurrently released a policy statement in march 2009 . to date , in absence of official internationally accepted guidelines , the advice is to attempt to distinguish genetic predisposition - based families with multiple pca cases from ascertainment - based multiple - case families and offer pca testing only to the former . this study assessed the knowledge of dutch urologists , general practitioners ( gps ) and clinical geneticists ( cgs ) about hpc and pca as a phenotype in hereditary syndromes and their post - erspc attitude towards pca testing and the role of family history in clinical decision - making about pca testing.methodsonline surveys were sent to all dutch urologists ( n = 351 ) , gps ( n = 69 ) and cgs ( n = 50 
\\ \hline
ROUGE2-F1 & 
24.86
\\ \hline
\makecell{Document\\(ID \#72)} & 
this retrospective study was approved by the institutional review board of our hospital ( 2014 - 09 - 037 ) . between november 2011 and october 2014 , 245 patients in the icu underwent bedside pdt with a ciaglia blue rhino percutaneous tracheostomy set ( cook medical , bloomington , in , usa ) due to prolonged endotracheal tube insertion . pdt was not performed in patients aged under 18 years , those with any pulsation palpated over the tracheostomy site , patients with a history of surgery or radiotherapy in the cervical region , and those with coagulopathy ( increased prothrombin time , inr > 2 ) . the first four procedures were performed by two physicians and one nurse using the standard technique , with one physician operating the bronchoscope while the other performed pdt . the next 55 procedures were performed by one physician and one nurse , with the physician using the bronchoscope only to confirm and evaluate the proper depth for the endotracheal tube before performing pdt using the simplified method described below . the remaining 186 procedures were performed using the simplified technique after repositioning the endotracheal tube at the predetermined depth ( 17 cm for males and 15 cm for females at the incisor ) . all pdts were performed by a single physician and one nurse , both of whom fully understood the entire procedure . the nurse administered sedatives and analgesics , as well as repositioning the endotracheal tube . except for patients with limited neck extension , a small pillow was placed under the patient 's shoulders to slightly extend the neck . transverse markings were made at the cricothyroid membrane and 2.0 cm toward the sternal notch ( fig . was also made in the midline of the trachea ... ... ... 
\\ \hline
Gold & 
when first introduced , percutaneous dilatational tracheostomy ( pdt ) was performed using a bronchoscope . the bronchoscope itself , however , has several disadvantages , including interruption of ventilation . modifications of the standard pdt technique have suggested that pdt can be safely performed without a bronchoscope . however , the safety of these modifications is unclear and many modifications have yet to be widely adopted . this study retrospectively evaluated the safety of a simplified pdt technique using the cook ciaglia blue rhino in 186 patients . after confirming the underlying structures by palpation , pdt was performed without a bronchoscope or blunt dissection , 2.0 cm below the cricothyroid membrane . ultrasonography was used only when palpation was difficult . except for bleeding in three patients with coagulopathy , none of the patients experienced any early or late complications , and there were no deaths related to pdt . these findings confirm that pdt can be safely performed using a simplified technique , which may be useful in situations with limited equipment and medical personnel .
\\ \hline
Model & 
backgroundpercutaneous tracheostomy ( pdt ) is one of the most common tracheostomy procedures performed in the intensive care unit ( icu ) . although the use of a fiberoptic bronchoscope ( fob ) during pdt has been recommended for precise positioning and to avoid complications , bronchoscopy itself can cause complications . this study assessed the safety of a simplified pdt technique using the ciaglia blue rhino , by repositioning the existing endotracheal tube at a pre - determined depth and determining proper pdt site by superficial palpation.methodsbetween november 2011 and october 2014 , 245 patients in the icu underwent bedside pdt with a ciaglia blue rhino percutaneous tracheostomy set due to prolonged endotracheal tube insertion . the first four procedures were performed by two physicians and one nurse using the standard technique , with one physician operating the bronchoscope while the other performed pdt . the next 55 procedures were performed by one physician and one nurse , with the physician using the bronchoscope only to confirm and evaluate the proper depth for the endotracheal tube before performing pdt using the simplified method described below . the remaining 186 procedures were performed using the simplified technique after repositioning the endo
\\ \hline
ROUGE2-F1 & 
10.53
\\ \hline
\end{tabular}
\caption{Generated summaries by $\text{PEGASUS}_{\text{LARGE}}$ (\gnews) on PubMed sampled by ROUGE2-F1.}
\end{table*}

\begin{table*}[bth!]
\scriptsize
\centering
\renewcommand{\arraystretch}{1.3}
\begin{tabular}{c|p{0.85\textwidth}}
\hline
\multicolumn{2}{c}{PubMed} \\ \hline
\makecell{Document\\(ID \#240)} & 
stainless steel alloys have remained the material of choice despite the emergence of the more recent titanium , composite and polycarbonate orthodontic brackets . stainless steel alloy contains 8\%-12\% nickel , 17\%-22\% chromium and other elements such as copper , iron molybdenum , manganese , silicon and sulfur[3 - 5 ] in the oral environment , orthodontic brackets are subjected to mechanical and chemical damaging which results in susceptibility to corrosion . corrosion leads to loss of substance from the material , change in its structural characteristics , or loss of structural integrity . due to the electrolytic capabilities of saliva various types of brackets are commercially available and each demonstrates a unique pattern of corrosion . in soldered brackets , this corrosion is due to the presence of dissimilar metals ( i.e. the silver solder and the stainless steel ) , a phenomenon termed galvanic corrosion . metal injection molding ( mim ) brackets are manufactured as a single unit and therefore do not demonstrate galvanic corrosion . corrosion can have detrimental effects on the surface of stainless steel brackets due to the continuous loss of metal ions . corrosion can increase the surface roughness of the bracket which leads to elevated friction forces between the bracket and the archwire . this increase in friction results in unfavorable distribution of forces and reduces the effectiveness of archwire guided orthodontic tooth movement.[7 - 8 ] moreover , by means of increased stress , the friction would further accelerate the corrosion process . the release of metal ions following the corrosion of brackets has concerned clinicians and has instigated research in this field . among these metal ions , furthermore , direct and prolonged contact of orthodontic appliances and the resulting corrosion products have been shown to cause local pain and ... ... ... 
\\ \hline
Gold & 
statement of the problem : stainless steel brackets release metallic ions following the process of corrosion in the oral environment . these released ions have potential adverse effects on health , friction between wire and bracket , staining , strength of brackets . choosing a bracket with favorable corrosive properties ; therefore , should be a goal of every practitioner . purpose : the goal of this study is to compare the amount of corrosion among five different brands of brackets using inductively coupled plasma ( icp ) mass spectrometry . materials and method : five different brands of brackets ( dentaurum , 3 m , ortho organizer , cobas and o.r.g ) were chosen and ten brackets were selected from each brand . a piece of stainless steel wire was ligated to each bracket . the bracket - archwire complex was then immersed in artificial saliva . subsequently , the samples were analyzed using an icp device and the levels of iron , chromium , nickel , and manganese ions were measured . results : the findings of this study demonstrated that iron was released the most from the tested brackets , followed by nickel . we also found that the cobas bracket had the most ion release among the tested brackets ( p < 0.05 ) 
\\ \hline
Model & 
abstractobjectivesthe release of metal ions following the corrosion of brackets has concerned clinicians and has instigated research in this field . the present study aimed to investigate five different brands of stainless steel brackets and compare their tendency towards corrosion by measuring ion release in an in - vitro setting.material and methodsten central incisor brackets were selected from each brand . in order to simulate conventional orthodontic treatment , 8 mm of 0.016 stainless steel archwire was tied in each bracket using 10 mm of 0.25 mm ligature wire ( dentaurum , germany ) . once the brackets were prepared , they were placed in poly - ethylene capped vials containing 10 ml of artificial saliva at a ph of 7.2 . the vials were incubated at 37c for 6 weeks and then they were subjected to thermocycling with 500 temperature cycles from 5c to 55c to simulate the effect of temperature changes in the oral cavity . the brackets were immersed in each bath for 30 seconds with 2 seconds at air temperature in - between the immersions . the details of the brackets selected for the study after thermal - cycling the solutions from the vials were analyzed to determine the amount of nickel , chromium , manganese , and iron using an inductively coupled plasma spectrometer ( icp )
\\ \hline
ROUGEL-F1 & 
23.41
\\ \hline
\makecell{Document\\(ID \#114)} & 
chronic pain affects up to 20\% of the population in developed nations.14 this represents a profound impact on individuals and their families alongside the sizeable burden on employers , health care systems , and society in general.3 when chronic pain occurs , it has the potential to become disease itself , and subsequently , chronic pain has emerged as a distinct phenomenon.5 management of chronic pain varies greatly between nations and even within nations . literature supports a multidisciplinary approach as the standard of care , although various health care systems may not always support this concept consistently.2 the current standard of care for chronic , noncancer pain typically includes many disciplines with the clinician developing an individualized treatment plan with the options of utilizing surgical interventions , pharmacology , and psychological and physical therapies . opioid analgesics are often prescribed , despite the lack of clinical evidence supporting their long - term use in the management of chronic pain.6 however , for many patients , this multidisciplinary approach is inadequate or ineffectual or is accompanied by the burden of side effects that are unacceptable and debilitating . only at this late stage , the field of neuromodulation for the treatment of pain has developed rapidly since the seminal paper on the electrical inhibition of pain by the stimulation of the dorsal column almost 50 years ago.7 the original term of dorsal column stimulation has evolved to become known as spinal cord stimulation ( scs).8 scs has been particularly effective as an adjunct in treating mixed neuropathic / nociceptive and neuropathic / radicular pain conditions such as failed back surgery syndrome ( fbss ) and complex regional pain syndrome ( crps ) . neuromodulation therapies offer a treatment option that has minimal side effects and that is relatively safe and ... ... ... 
\\ \hline
Gold & 
spinal cord stimulation ( scs ) applications and technologies are fast advancing . new scs technologies are being used increasingly in the clinical environment , but often there is a lag period between the clinical application and the publishing of high - quality evidence on safety and efficacy . recent developments will undoubtedly expand the applicability of scs , allowing more effective and individualized treatment for patients , and may have the potential to salvage patients who have previously failed neuromodulation . already , high - level evidence exists for the safety , efficacy , and cost - effectiveness ( level i ii ) of traditional scs therapies in the treatment of chronic refractory low back with predominant limb pain ( regardless of surgical history ) . more than half of all patients with chronic painful conditions experience sustained and significant levels of pain reduction following scs treatment . although only limited evidence exists for burst stimulation , there is now level i evidence for both dorsal root ganglion scs and high - frequency scs that demonstrates compelling results compared with traditional therapies . the body of evidence built on traditional scs research may be redundant , with newer iterations of scs therapies such as dorsal root ganglion scs , high - frequency scs , and burst scs . a number of variables
\\ \hline
Model & 
chronic pain affects up to 20\% of the population in developed nations . the current standard of care for chronic , noncancer pain typically includes many disciplines with the clinician developing an individualized treatment plan with the options of utilizing surgical interventions , pharmacology , and psychological and physical therapies . however , for many patients , this multidisciplinary approach is inadequate or ineffectual or is accompanied by the burden of side effects that are unacceptable and debilitating . only at this late stage , the field of neuromodulation for the treatment of pain has developed rapidly since the seminal paper on the electrical inhibition of pain by the stimulation of the dorsal column almost 50 years ago . spinal cord stimulation ( scs ) has been particularly effective as an adjunct in treating mixed neuropathic / nociceptive and neuropathic / radicular pain conditions such as failed back surgery syndrome ( fbss ) and complex regional pain syndrome ( crps ) . in traditional scs therapies , the objective has been to replace the pain sensation with paresthesia that requires mapping of stimulation to the region of pain . in recent years , the next phase in the evolution of neuromodulation has become available with the development of dorsal root ganglion scs and the emerging use of two novel advances in stimulation frequencies
\\ \hline
ROUGEL-F1 & 
17.68
\\ \hline
\end{tabular}
\caption{Generated summaries by $\text{PEGASUS}_{\text{LARGE}}$ (\gnews) on PubMed sampled by ROUGEL-F1.}
\end{table*}

\begin{table*}[bth!]
\scriptsize
\centering
\renewcommand{\arraystretch}{1.3}
\begin{tabular}{c|p{0.85\textwidth}}
\hline
\multicolumn{2}{c}{AESLC} \\ \hline
\makecell{Document\\(ID \#260)} & 
Subject: Access Card Information Dear Colleagues, We know you have many questions about your new benefits plan, payroll issues and other changes that will occur over the next few weeks. Human Resources and Communications are preparing a thorough Q\&A, which we will send to you tomorrow and will regularly update as more information is available. Meanwhile, please read the announcement below from Security about how you can obtain a new UBS Warburg access card that will allow you to enter our new office space after Friday. I am confident that you share my enthusiasm about our new opportunity. We look forward to working along side each of you in the coming days. David Oxley Human Resources
\\ \hline
Gold & 
Access Card Information
\\ \hline
Model & 
Access Card Information
\\ \hline
ROUGE1-F1 & 
100.00
\\ \hline
\makecell{Document\\(ID \#33)} & 
The funerals of Andy, Darren and Paul, who tragically lost their lives in last Wednesday's explosion at Teesside Power Station, will be held this week at the following times: Andy Sherwood 2.00pm, Wednesday, August 15 St James Church, Hartlepool Darren Higgins 9.15 am, Thursday, August 16 St. Andrews Church, Teesville, Middlesbrough Paul Surtees 1.30 pm, Thursday, August 16 Easington Church of the Ascension, Easington Colliery We know that everyone at Enron will want to remember the colleagues we have lost at this difficult time, and our thoughts and prayers are with their families and friends. We are considering how best to commemorate the lives of Andy, Darren and Paul, and will be discussing our ideas with their families, in due course. Graeme White is recovering in Middlesbrough General Hospital where his condition remains stable and we are continuing to provide as much support as possible for Graeme and his family. The plant remains off-line and Enron continues to co-operate fully with the Health \& Safety Executive's investigation into the cause of the explosion. John Sherriff and Michael Brown
\\ \hline
Gold & 
TEESSIDE POWER STATION -- UPDATE
\\ \hline
Model & 
Teesside Update
\\ \hline
ROUGE1-F1 & 
66.67
\\ \hline
\makecell{Document\\(ID \#230)} & 
Wayne, I have modified the above contract as we discussed. The "Customer's Agent" language is contained in Appendix "1". Please have the contract signed by JER Partners, a J.E. Robert Company and fax it back to me. I will then have it signed by Houston Pipe Line Company. I have left the attachment at the end as originally drafted until I am able to talk to Gary Lamphier. If it is truly correct that there are no minimum quantity obligations under the deal, then Gary may have to reprice the \$0.09 discount for the 1000/d. I will leave that up to Gary and you to resolve.
\\ \hline
Gold & 
JER / BRE Austin Tech, L.P.
\\ \hline
Model & 
JER Partners, L.L.C.
\\ \hline
ROUGE1-F1 & 
36.36
\\ \hline
\makecell{Document\\(ID \#231)} & 
Steve: I just spoke with Bill Bradford and I understand that Bill and Mark Tarkington discussed the use of "Specified Entity" and agreed, subject to DB committee approval, to eliminate the use of Specified Entity in the ENA and affiliated masters with DB (i.e., "Specified Entity: none"). Once you have confirmed this with Mark, please prepare a final blacklined draft for our review and then execution copies of the ISDA Master Agreement. In connection with the final draft, Tana Jones will fax to you minor format comments. [She will also fax to you the electronic trading agreements and password applications for London and U.S. based trading.] The following ENA affiliates will also need to execute ISDA Master Agreements with DB: (1) ECT Investments, Inc. (Delaware corporation) (2) Enron Capital \& Trade Resources International Corp. (Delaware corporation) (3) EnronCredit.com Limited (U.K. corporation)
\\ \hline
Gold & 
Enron North America Corp. ("ENA")/Deutsche Bank ("DB") ISDA Master Agreement
\\ \hline
Model & 
Deutsche Bank
\\ \hline
ROUGE1-F1 & 
30.77
\\ \hline
\makecell{Document\\(ID \#162)} & 
Ms. Peppers, I am very sorry, but I must once again reschedule the appointment that you have with Mr. Lay. He will be out of town the entire week of September 26. The first available time to reschedule is October 19, 2000. Would an afternoon appointment work for you - about 2:00 p.m.? I look forward to hearing from you. Rosalee
\\ \hline
Gold & 
Reschedule appointment
\\ \hline
Model & 
Mr. Lay's schedule
\\ \hline
ROUGE1-F1 & 
0.00
\\ \hline
\makecell{Document\\(ID \#108)} & 
Due to a variety of circumstances we did not migrate employees to the UBSWE environment tonight (Monday, February 4, 2002). We will communicate further details of the migration plan on Tuesday, February 5, 2002. Thank you for your cooperation in advance, please contact the Resolution Center at x 3-1411 should you have any questions. UBSW Energy Global Infrastructure
\\ \hline
Gold & 
DRAFT: INFORMATION, Please Read: UBSWE Migration postponed
\\ \hline
Model & 
migrated employees
\\ \hline
ROUGE1-F1 & 
0.00
\\ \hline
\end{tabular}
\caption{Generated summaries by $\text{PEGASUS}_{\text{LARGE}}$ (\gnews) on AESLC sampled by ROUGE1-F1.}
\end{table*}
\begin{table*}[bth!]
\scriptsize
\centering
\renewcommand{\arraystretch}{1.3}
\begin{tabular}{c|p{0.85\textwidth}}
\hline
\multicolumn{2}{c}{AESLC} \\ \hline
\makecell{Document\\(ID \#292)} & 
Just a reminder - We have added a new pub code to EnPower. The code is for the Dow Jones COB Sunday Index. Please advise that the Portland West Desk has one deal in the system at this time against that index, and more may follow. Thanks for your help with entering these prices. Kate
\\ \hline
Gold & 
New Pub Code
\\ \hline
Model & 
New pub code for EnPower
\\ \hline
ROUGE2-F1 & 
66.67
\\ \hline
\makecell{Document\\(ID \#100)} & 
El Paso recently filed their proposal to reallocate receipt point capacity. Comments are due on Thursday, May 17, 2001. We would like to meet with you to make sure our comments address the commercial concerns regarding the reallocation plan. In that regard, please plan to attend a meeting on Thursday, May 3 at 3:45 in 3269. Becky will provide for your review a summary of El Paso's proposal. Please let me know if you have any questions. Thanks
\\ \hline
Gold & 
El Paso's Proposal to Address the Allocation of Receipt Point Capacity
\\ \hline
Model & 
Comments on El Paso's Reallocation Proposal
\\ \hline
ROUGE2-F1 & 
23.53
\\ \hline
\makecell{Document\\(ID \#241)} & 
Media reports out of France indicate that explosions have just taken place at two chemical plants in the southern city of Tououse. The main explosion reportedly took place at a plant that manufactures fuel for Arianespace, the European Space Agency's commercial arm. Explosions have also been reported at a plant owned by Groupe SNPE, which manufactures propellants and explosives. The incidents are reported to have involved deaths at the scene. We have yet to acquire independent confirmation of the explosions, or whether terrorism is suspected. We are in touch with military and intelligence sources, and will provide updates.
\\ \hline
Gold & 
**** Unconfirmed Explosions at French Chemical Plants
\\ \hline
Model & 
Recent explosions in France
\\ \hline
ROUGE2-F1 & 
0.00
\\ \hline
\makecell{Document\\(ID \#189)} & 
Nothing from Ben yet. If I don't hear from him in the next couple of hours, then I think we should be ready with an extension letter, giving them a couple of days extra for free. Ben indicated to me last week that he would give them some time once they asked. MK called with questions on the assignment - nothing serious. Kay
\\ \hline
Gold & 
NW
\\ \hline
Model & 
SMUD
\\ \hline
ROUGE2-F1 & 
0.00
\\ \hline
\makecell{Document\\(ID \#104)} & 
Hey Chris, I was wondering if there was a fax number at the base so that my apartment complex can send you some info. regarding getting your name taken off of our lease. If there is, just shoot me off an email with the number and I'll pass it along to them. Hope things are going well and that you're not too homesick with Thanksgiving and all. I have a feeling I will be talking to you guys tomorrow along with both of our parents and extended family members. Have a great holiday and root on the Horns ( I think this is going to be a tough one). Love,
\\ \hline
Gold & 
Apartment Lease
\\ \hline
Model & 
Hello!
\\ \hline
ROUGE2-F1 & 
0.00
\\ \hline
\makecell{Document\\(ID \#37)} & 
Kal/Susan - Attached is an updated version of the disclosure consent agreement for EOL information. Please take a look at the attached and let me know your thoughts. Mark
\\ \hline
Gold & 
EOL Disclosure Consent
\\ \hline
Model & 
EOL Consent
\\ \hline
ROUGE2-F1 & 
0.00
\\ \hline
\end{tabular}
\caption{Generated summaries by $\text{PEGASUS}_{\text{LARGE}}$ (\gnews) on AESLC sampled by ROUGE2-F1.}
\end{table*}

\begin{table*}[bth!]
\scriptsize
\centering
\renewcommand{\arraystretch}{1.3}
\begin{tabular}{c|p{0.85\textwidth}}
\hline
\multicolumn{2}{c}{AESLC} \\ \hline
\makecell{Document\\(ID \#192)} & 
I need a copy of your insurance before the flight on 11/25. Our records show your insurance expired on 10/4/2001 I have to have your insurance on file before your flight. thank you
\\ \hline
Gold & 
insurance
\\ \hline
Model & 
Insurance
\\ \hline
ROUGEL-F1 & 
100.00
\\ \hline
\makecell{Document\\(ID \#106)} & 
As discussed during our recent demonstration of our new portfolio tracking system (PortRAC), we have completed a security system that will provide a "firewall" controlling access to each business unit's data. Our intent is to open up the system to designated members of each business unit or others as approved by the COO or Managing Director of that group. In order to complete the initial setup, we need to receive a listing of approved users for your business unit and their login Id's. If you prefer to delegate security approval authority to someone else on your staff, please let us know and we will establish procedures with them. If additional demonstrations of PortRAC are necessary for your group, please let us know. Thank you for your assistance. Rick C.
\\ \hline
Gold & 
PortRAC Security
\\ \hline
Model & 
PortRAC Security System
\\ \hline
ROUGEL-F1 & 
80.00
\\ \hline
\makecell{Document\\(ID \#254)} & 
Mike, Here's a revised version of the option agreement. I changed the shipping dates, quantity (10), Outside Exercise Date, and the amount of the option payment. I did not change the prices, since I had not discussed this with Ben, and couldn't get the numbers to work. Have there been some changes to the data sheets? It appears to me that there might be a version which reflects the inclusion of the low sound option. In the last draft we did not include the low sound option because the impact on the load losses had not been agreed, so the prices did not reflect the option. Please call or email me if you have any questions or comments. Kay
\\ \hline
Gold & 
Option
\\ \hline
Model & 
Rev option agreement
\\ \hline
ROUGEL-F1 & 
50.00
\\ \hline
\makecell{Document\\(ID \#267)} & 
Susan: I'm not sure that what I told you to do in the CSA Annex with respect to Custodians and their qualifications work so for now let's keep it out and go back to our old language. Sara and Mark: Shari had a counterparty point out to her that in our CSA Annex language where we talk about the circumstances under which a party can "hold" collateral, all of our triggers are based on something happening to our counterparty and nothing is said about what would happen if a Custodian holds the colateral and fails to be creditworthy. Do you have any thoughts on this? At first I thought that we should require the Custodian to be a Qualified institution but I'm not sure that that is the right approach. Let me know what you think. Carol
\\ \hline
Gold & 
Custodian Language
\\ \hline
Model & 
Custodians
\\ \hline
ROUGEL-F1 & 
0.00
\\ \hline
\makecell{Document\\(ID \#162)} & 
Ms. Peppers, I am very sorry, but I must once again reschedule the appointment that you have with Mr. Lay. He will be out of town the entire week of September 26. The first available time to reschedule is October 19, 2000. Would an afternoon appointment work for you - about 2:00 p.m.? I look forward to hearing from you. Rosalee
\\ \hline
Gold & 
Reschedule appointment
\\ \hline
Model & 
Mr. Lay's schedule
\\ \hline
ROUGEL-F1 & 
0.00
\\ \hline
\makecell{Document\\(ID \#22)} & 
the tax issue that we discussed at the rmt meeting has been cleared. now we need to see if the \$250 million is somehow at the rmt level or would be trapped in some way in a rmt filing. i keep thinking that i will have time to track down with weil and has not happened. can you track this down and then let the group know. add stuart to this list. can you do? ?
\\ \hline
Gold & 
rmt
\\ \hline
Model & 
ISDA
\\ \hline
ROUGEL-F1 & 
0.00
\\ \hline
\end{tabular}
\caption{Generated summaries by $\text{PEGASUS}_{\text{LARGE}}$ (\gnews) on AESLC sampled by ROUGEL-F1.}
\end{table*}

\begin{table*}[bth!]
\scriptsize
\centering
\renewcommand{\arraystretch}{1.3}
\begin{tabular}{c|p{0.85\textwidth}}
\hline
\multicolumn{2}{c}{BillSum} \\ \hline
\makecell{Document\\(ID \#193)} & 
SECTION 1. SHORT TITLE. This Act may be cited as the   Veterans Homebuyer Accessibility Act of 2015''. SEC. 2. VETERAN FIRST-TIME HOMEBUYER TAX CREDIT. (a) In General.--Section 36(a) of the Internal Revenue Code of 1986 is amended to read as follows:   (a) Allowance of Credit.--In the case of an eligible veteran who purchases a principal residence in the United States during the taxable year, there shall be allowed as a credit against the tax imposed by this subtitle for such taxable year an amount equal to 10 percent of the purchase price of the residence.''. (b) Additional Limitation for Adaptive Housing Improvements.-- Section 36(b)(1) of such Code is amended by redesignating subparagraphs (B), (C), and (D) as subparagraphs (C), (D), and (E), respectively, and by inserting after subparagraph (A) the following new subparagraph:   (B) Special rule for adaptive housing improvements.--In the case of a principal residence with special fixtures or movable facilities made necessary by the nature of the disability of the veteran, if such fixtures and facilities are--   (i) provided to the veteran pursuant to specially adapted housing assistance under chapter 17 or 21 of title 38, United States Code, or   (ii) similar to such fixtures and facilities that would be provided to the veteran if the veteran received such assistance, then subparagraph (A) shall be increased by the lesser of \$8,000 or the portion of the purchase price of the principal residence attributable such fixtures or movable facilities.''. (c) Eligible Veteran.-- (1) In general.--Section 36(c)(1) of such Code is amended by striking   First-time homebuyer.--The term   first time homebuyer' means any individual'' and inserting   Eligible veteran.--The term   eligible veteran' means any individual who is a veteran (as defined in section 101(2) ... ... ... 
\\ \hline
Gold & 
Veterans Homebuyer Accessibility Act of 2015 This bill amends the Internal Revenue Code, with respect to the tax credit for first-time homebuyers, to: (1)\&nbsp;allow a veteran of the Armed Forces a tax credit for 10\% of the purchase price of a principal residence purchased prior to January 1, 2018; (2)\&nbsp;allow an additional credit for the cost, not exceeding \$8,000 in a taxable year, of installing special fixtures or movable facilities in a residence to accommodate a disability of the veteran; and (3) require a recapture of credit amounts if the veteran sells such residence within 36 months after purchasing it.
\\ \hline
Model & 
Veterans Homebuyer Accessibility Act of 2015 Amends the Internal Revenue Code, with respect to the tax credit for first-time homebuyers, to allow veterans of the Armed Forces a tax credit for 10\% of the purchase price of a principal residence purchased prior to January 1, 2017. Allows an additional credit for the cost of installing special fixtures or movable facilities in a residence to accommodate a disability of the veteran. Requires a recapture of credit amounts if the veterans sells such residence within 36 months after purchasing it.
\\ \hline
ROUGE1-F1 & 
86.15
\\ \hline
\makecell{Document\\(ID \#122)} & 
SECTION 1. SHORT TITLE. This Act may be cited as the   Patients' Formulary Rights Act of 1999''. SEC. 2. PATIENT PROTECTIONS AGAINST ABUSE OF FORMULARIES FOR PRESCRIPTION DRUGS. (a) Group Health Plans.-- (1) Public health service act amendments.--(A) Subpart 2 of part A of title XXVII of the Public Health Service Act is amended by adding at the end the following new section:   SEC. 2707. STANDARDS RELATING TO USE OF FORMULARIES AND THERAPEUTIC SUBSTITUTION.   (a) Requirements on Use of Formularies.--   (1) In general.--A group health plan, and a health insurance issuer offering group health insurance coverage, shall not use a formulary unless the plan or issuer--   (A) notifies participants, beneficiaries, and enrollees, prior to initial enrollment or coverage, and makes available at any time to health care professionals who prescribe prescription drugs under the plan or coverage of the information described in paragraph (2);   (B) notifies participants, beneficiaries, enrollees, and health care providers who prescribe covered prescription drugs under the plan or coverage on a routine and annual basis of any changes in (including deletions from) the formulary; and   (C) in the case of a participant, beneficiary, or enrollee who is provided coverage for a prescription drug at the time the drug is removed from the formulary, to permit the participant, beneficiary, or enrollee to continue to have the drug prescribed for treatment of the same condition for which it was previously prescribed.   (2) Information to be disclosed.--The information described in this paragraph is as follows (with respect to prescription drug coverage under a group health plan or health insurance coverage):   (A) Extent of therapeutic substitution.--What constitutes the practice or therapeutic substitution that may be effected under the plan or coverage.  ... ... ... 
\\ \hline
Gold & 
Includes within required formulary information: (1) the extent of therapeutic substitution; (2) a complete list of all drugs included in the formulary; (3) information on access to nonformulary drugs; (4) whether or not, and to what extent, coverage will exist for nonformulary drugs; (5) copayments or other cost-sharing for formulary drugs; and (6) limits on coverage payments for outpatient formulary and nonformulary drugs. Amends title XVIII (Medicare) of the Social Security Act to require its authorized health insurance issuers and any Medicare+Choice organization to comply with such notification requirements in the same manner that such requirements apply to health insurance coverage or issuers in connection with a group plan. Requires the coordination of notification and formulary requirements through an interagency memorandum of understanding among the Secretaries of Labor, the Treasury, and Health and Human Services.
\\ \hline
Model & 
Patients' Formulary Rights Act of 1999 - Amends the Public Health Service Act, the Employee Retirement Income Security Act of 1974, and the Internal Revenue Code to prohibit a group health plan and a health insurance issuer offering group coverage from using a formulary unless the plan or issuer: (1) notifies participants, beneficiaries, and enrollees, prior to initial enrollment or coverage, and makes available at any time to health care professionals who prescribe prescription drugs; (2) notifies participants, beneficiaries, enrollees, and health care providers who prescribe covered prescription drugs on a routine and annual basis of any changes in the formulary; and (3) in the case of a participant, beneficiary, or enrollee who is provided coverage for a prescription drug at the time the drug is removed from the formulary, permits the participant, beneficiary, or enrollee to continue to have the drug prescribed for treatment of the same condition for which it was previously prescribed. Requires such information to be disclosed to participants, beneficiaries, enrollees, and health care providers. Prohibits a plan or issuer from using a formulary unless the plan or issuer: (1) notifies participants, beneficiaries, and enrollees, prior to initial enrollment or coverage, and makes available at any time to health care professionals who prescribe prescription drugs; (2) makes available at any time to health care professionals
\\ \hline
ROUGE1-F1 & 
35.49
\\ \hline
\end{tabular}
\caption{Generated summaries by $\text{PEGASUS}_{\text{LARGE}}$ (\gnews) on BillSum sampled by ROUGE1-F1.}
\end{table*}

\begin{table*}[bth!]
\scriptsize
\centering
\renewcommand{\arraystretch}{1.3}
\begin{tabular}{c|p{0.85\textwidth}}
\hline
\multicolumn{2}{c}{BillSum} \\ \hline
\makecell{Document\\(ID \#153)} & 
SECTION 1. SHORT TITLE. This Act may be cited as the   Statewide PTV Access Act of 2009''. SEC. 2. SECONDARY TRANSMISSION OF NONCOMMERCIAL STATIONS BY SATELLITE CARRIERS OUTSIDE LOCAL MARKETS. (a) Secondary Transmissions.--Section 119(a)(2)(C) of title 17, United States Code, is amended-- (1) by redesignating clauses (iii), (iv), and (v) as clauses (iv), (v), and (vi), respectively; and (2) by inserting after clause (ii) the following:   (iii) Noncommercial educational broadcast stations.--   (I) Secondary transmissions within state-wide network.--In the case of a State-wide network of noncommercial educational broadcast stations, the statutory license provided for in subparagraph (A) shall apply to the secondary transmission of any noncommercial educational broadcast station in that State-wide network to any subscriber in any county within that State that is located outside that station's local market and is not served by a noncommercial educational television broadcast station that is located within that State.   (II) State-wide network defined.--In this clause, the term   State-wide network of noncommercial educational broadcast stations' is a network of three or more noncommercial educational broadcast stations that are licensed to a single State, political, educational, or special purpose subdivision of a State, or a single public agency.''. (b) Royalty Fee.--Section 119(b)(1)(B) of title 17, United States Code, is amended by inserting   or paragraph (2)(C)(iii)'' after   subscriber under paragraph (3)''. SEC. 3. AMENDMENTS TO THE COMMUNICATIONS ACT OF 1934. (a) In General.--Section 325(b)(2) of the Communications Act of 1934 (47 U.S.C. 325 (b)(2)) is amended-- (1) by striking   or'' at the end of subparagraph (D); (2) by striking the period at the end of subparagraph (E) and inserting   ; or''; and (3) by inserting after subparagraph (E) the following new subparagraph:   (F) to retransmission of ... ... ... 
\\ \hline
Gold & 
Statewide PTV Access Act of 2009 - Permits a satellite carrier to retransmit the signals of a state-wide network of noncommercial educational broadcast stations to any subscriber in any county within such state that is: (1) outside that station's local market; and (2) not served by an in-state noncommercial educational television broadcast station. Amends the Communications Act of 1934 to permit a satellite carrier to retransmit the signals of a television broadcast station outside the station's local market directly to its subscribers if: (1) such station is a part of a network of three or more noncommercial educational broadcast stations that are licensed to a single state, political, educational, or special purpose subdivision of a state, or to a single public agency; and (2) the carrier retransmits such signal only to subscribers in a county within that state that is outside the station's local market and is not served by another in-state noncommercial educational broadcast station.
\\ \hline
Model & 
Statewide PTV Access Act of 2009 - Amends federal copyright law to provide that the statutory license provided for the secondary transmission of any noncommercial educational broadcast station in a state-wide network of such stations shall apply to the secondary transmission of any such station in any county within that state that is located outside that station's local market and is not served by a noncommercial educational television broadcast station located within that state. Defines "state-wide network of noncommercial educational broadcast stations" as a network of three or more noncommercial educational broadcast stations that are licensed to a single state, political, educational, or special purpose subdivision of a state, or a single public agency. Requires a satellite carrier that retransmits the signal of a television broadcast station outside the station's local market by a satellite carrier directly to its subscribers, if: (1) the station is part of a network of three or more noncommercial educational broadcast stations that are licensed to a single state, political, educational, or special purpose subdivision of a state, or a single public agency; and (2) the satellite carrier retransmits the secondary signal of such eligible station only to its subscribers located in any county within that state that is located outside that station's local market and not served by another noncommercial educational broadcast station located
\\ \hline
ROUGE2-F1 & 
56.40
\\ \hline
\makecell{Document\\(ID \#210)} & 
SECTION 1. SHORT TITLE. This Act may be cited as the   Medicare Common Access Card Act of 2017''. SEC. 2. MEDICARE SMART CARD PILOT PROGRAM. Part E of title XVIII of the Social Security Act is amended by inserting after section 1866E the following new section:   SEC. 1866F. SMART CARD PILOT PROGRAM.   (a) Implementation.--   (1) In general.--Not later than 36 months after the date of the enactment of this section, the Secretary shall establish a pilot program (in this section referred to as the   pilot program') to demonstrate the feasibility of using smart card technology under this title.   (2) Smart card technology defined.--In this section, the term   smart card technology' means the following:   (A) Beneficiary smart card.--A machine readable, fraud- and tamper-resistant card (in this section referred to as a   smart card') that includes an embedded integrated circuit chip with a secure micro- controller that enables the verification and secure, electronic authentication of the identity of a Medicare beneficiary at the point of service through a combination of the smart card and a personal identification number known by or associated with such beneficiary.   (B) Card reader technology.--Information technology that enables a supplier and provider to authenticate the identity of a Medicare beneficiary through presentation of such a smart card and such components, with such authentication to be reflected through the use of a modifier or in another appropriate manner, as determined by the Secretary, in the claims adjudication process.   (3) Program design elements.--The pilot program shall be conducted for a period of 3 years consistent with the following:   (A) Selection of area.--In consultation with the Inspector General of the Department of Health and Human Services, the ... ... ... 
\\ \hline
Gold & 
Medicare Common Access Card Act of 2017 This bill requires the Centers for Medicare \&amp; Medicaid Services (CMS) to establish a three-year pilot program, in at least three\&nbsp;geographic areas,\&nbsp;to demonstrate the feasibility of using smart-card technology to authenticate the identity of a Medicare beneficiary at points of service. The CMS shall select supplier and provider types that will be required to participate in the pilot program as a condition of Medicare payment, but\&nbsp;must exempt from participation a supplier or provider that: (1) does not have access to card-reader technology, (2) does not have sufficient internet access, or (3) has a low volume of Medicare claims. The CMS shall select a private contractor to implement and operate the pilot program. The CMS must submit specified reports to Congress on program design, implementation, and performance. For purposes of conducting the pilot program, the CMS shall\&nbsp;provide for the\&nbsp;transfer of \$150 million from the Supplemental Medical Insurance Trust Fund to the CMS Program Management Account.
\\ \hline
Model & 
Medicare Common Access Card Act of 2017 This bill requires the Centers for Medicare \&amp; Medicaid Services (CMS) to establish a pilot program that demonstrates the feasibility of using smart card technology to address Medicare fraud. Under the program, smart card technology must be issued free-of-charge to selected Medicare beneficiaries, suppliers, and providers; such technology must support the secure, electronic authentication of beneficiary identity at points of service. In selecting program participants, the CMS must consider the risk of fraud, waste, or abuse among categories of suppliers and providers.
\\ \hline
ROUGE2-F1 & 
25.10
\\ \hline
\end{tabular}
\caption{Generated summaries by $\text{PEGASUS}_{\text{LARGE}}$ (\gnews) on BillSum sampled by ROUGE2-F1.}
\end{table*}

\begin{table*}[bth!]
\scriptsize
\centering
\renewcommand{\arraystretch}{1.3}
\begin{tabular}{c|p{0.85\textwidth}}
\hline
\multicolumn{2}{c}{BillSum} \\ \hline
\makecell{Document\\(ID \#94)} & 
SECTION 1. SHORT TITLE. This Act may be cited as the   Sensible Estate Tax Act of 2008''. SEC. 2. RESTORATION OF ESTATE TAX; REPEAL OF CARRYOVER BASIS. (a) In General.--The following provisions of the Economic Growth and Tax Relief Reconciliation Act of 2001, and the amendments made by such provisions, are hereby repealed: (1) Subtitles A and E of title V. (2) Subsection (d), and so much of subsection (f)(3) as relates to subsection (d), of section 511. (3) Paragraph (2) of subsection (b), and paragraph (2) of subsection (e), of section 521. The Internal Revenue Code of 1986 shall be applied as if such provisions and amendments had never been enacted. (b) Sunset Not To Apply.-- (1) Subsection (a) of section 901 of the Economic Growth and Tax Relief Reconciliation Act of 2001 is amended by striking   this Act'' and all that follows and inserting   this Act (other than title V) shall not apply to taxable, plan, or limitation years beginning after December 31, 2010.''. (2) Subsection (b) of such section 901 is amended by striking   , estates, gifts, and transfers''. SEC. 3. UNIFIED CREDIT AGAINST THE ESTATE TAX. (a) In General.--Subsection (c) of section 2010 of the Internal Revenue Code of 1986 (relating to applicable credit amount) is amended by striking all that follows   the applicable exclusion amount'' and inserting   . For purposes of the preceding sentence, the applicable exclusion amount is \$2,000,000.''. (b) Inflation Adjustment.--Subsection (c) of section 2010 of such Code, as amended by subsection (a), is amended-- (1) by striking   For purposes of this section,'' and inserting the following:   (1) In general.--For purposes of this section,'', and (2) by adding at the end the following new paragraph:   ... ... ... 
\\ \hline
Gold & 
Sensible Estate Tax Act of 2008 - Repeals provisions of the Economic Growth and Tax Reconciliation Act of 2001 relating to the estate and gift tax. Amends the Internal Revenue Code to: (1) allow an estate tax exclusion of \$2 million adjusted for inflation in calendar years after 2008; (2) revise the estate tax rates for larger estates; (3) restore the estate tax credit for state estate, inheritance, legacy, or succession taxes; (4) restore the unified credit against the gift tax; and (5) allow a surviving spouse an increase in the unified estate tax credit by the amount of any unused credit of a deceased spouse.
\\ \hline
Model & 
Sensible Estate Tax Act of 2008 - Repeals provisions of the Economic Growth and Tax Relief Reconciliation Act of 2001 (EGTRRA) eliminating the tax on estates and generation-skipping transfers and the step-up in basis provisions for property acquired from a decedent for estates of decedents dying after 2008. Declares that the sunset provision (general terminating date of December 10, 2010) of EGTRRA shall not apply to title V of such Act (Estate, Gift, and Generation-Skipping Transfer Tax Provisions). Amends the Internal Revenue Code to: (1) restore the unified credit against the estate tax; (2) provide for an inflation adjustment to the estate tax exclusion amount after 2008; (3) establish maximum estate tax rates of over \$5 million; and (4) restore the credit for state estate, inheritance, legacy, or succession taxes after 2008.
\\ \hline
ROUGEL-F1 & 
41.84
\\ \hline
\makecell{Document\\(ID \#216)} & 
SECTION 1. SHORT TITLE; TABLE OF CONTENTS. (a) Short Title.--This Act may be cited as the   Meeting Our Responsibility to Medicare Beneficiaries Act of 2005''. (b) Table of Contents.--The table of contents of this Act is as follows: Sec. 1. Short title; table of contents. TITLE I--ELIMINATING SPECIAL INTEREST PREFERENCES Sec. 101. Negotiating fair prices for medicare prescription drugs. Sec. 102. Elimination of MA Regional Plan Stabilization Fund (Slush Fund). Sec. 103. Application of risk adjustment reflecting characteristics for the entire medicare population in payments to Medicare Advantage organizations. TITLE II--IMPROVING THE MEDICARE PROGRAM FOR BENEFICIARIES Sec. 201. Eliminating coverage gap. Sec. 202. Requiring two prescription drug plans to avoid Federal fallback. Sec. 203. Waiver of part D late enrollment penalty for transition period. Sec. 204. Improving the transition of full-benefit dual eligible individuals to coverage under the medicare drug benefit. Sec. 205. Part B premium reduction. Sec. 206. Study and report on providing incentives to preserve retiree coverage. Sec. 207. Promoting transparency in employer subsidy payments. TITLE I--ELIMINATING SPECIAL INTEREST PREFERENCES SEC. 101. NEGOTIATING FAIR PRICES FOR MEDICARE PRESCRIPTION DRUGS. (a) In General.--Section 1860D-11 of the Social Security Act (42 U.S.C. 1395w-111) is amended by striking subsection (i) (relating to noninterference) and by inserting the following new subsection:   (i) Authority To Negotiate Prices With Manufacturers.--   (1) In general.--The Secretary shall have authority similar to that of other Federal entities that purchase prescription drugs in bulk to negotiate contracts with manufacturers of covered part D drugs, consistent with the requirements and in furtherance of the goals of providing quality care and containing costs under this part.   (2) Required use of authority.--   (A) Fallback plans.--The Secretary shall exercise the authority described in paragraph (1) with respect to covered part D drugs ... ... ... 
\\ \hline
Gold & 
Meeting Our Responsibility to Medicare Beneficiaries Act of 2005 - Amends title XVIII (Medicare) of the Social Security Act (SSA) with respect to: (1) negotiating fair prices for Medicare prescription drugs; (2) elimination of the MA Regional Plan Stabilization Fund; (3) application of risk adjustment reflecting characteristics for the entire Medicare population in payments to Medicare advantage organizations; (4) modification of the annual out-of-pocket threshold with respect to prescription drug benefits; (5) requiring two prescription drug plans to avoid Federal fallback; (6) waiver of the part D (Voluntary Prescription Drug Benefit Program); (7) transition of full-benefit dual eligible individuals to coverage under the Medicare drug benefit; and (8) Medicare part B (Supplementary Medical Insurance) premium reduction. Directs the Secretary to study and report to Congress on providing incentives to preserve retiree coverage. Amends SSA title XVIII to direct the Secretary to make certain information regarding the sponsor of a qualified prescription drug plan receiving a subsidy under the prescription drug program available to the public through the Internet website of the Centers for Medicare \& Medicaid Services.
\\ \hline
Model & 
Meeting Our Responsibility to Medicare Beneficiaries Act of 2005 - Amends title XVIII (Medicare) of the Social Security Act to repeal the prohibition against interference by the Secretary of Health and Human Services with the negotiations between drug manufacturers and pharmacies and prescription drug plan sponsors. Grants the Secretary authority to negotiate contracts with manufacturers of covered Medicare part D (Voluntary Prescription Drug Benefit Program) drugs, consistent with requirements and in furtherance of the goals of providing quality care and containing costs under Medicare part D. Requires the Secretary to exercise such authority with respect to covered part D drugs offered under fallback prescription drug plans, if the Secretary determines that the negotiated prices are not fair and affordable prices compared to prices obtained by other Federal programs for such drugs. Eliminates the MA Regional Plan Stabilization Fund (Slush Fund). Directs the Secretary to ensure that payments to Medicare Advantage organizations are adjusted based on factors to ensure that the health status of the enrollee is reflected in such adjusted payments, including adjusting for the difference between the health status of the enrollee and individuals enrolled under the original Medicare fee-for-service program under parts A and B of title XVIII (Medicare) of the Social Security Act. Amends SSA title XVIII to provide for a waiver of the part D late enrollment penalty for the transition period
\\ \hline
ROUGEL-F1 & 
28.15
\\ \hline
\end{tabular}
\caption{Generated summaries by $\text{PEGASUS}_{\text{LARGE}}$ (\gnews) on BillSum sampled by ROUGEL-F1.}
\end{table*}

\end{document}